%% file: main.tex
\documentclass[journal]{IEEEtran}
\usepackage[T1]{fontenc}
\usepackage[utf8]{inputenc}

\usepackage[a4paper]{geometry}

\usepackage[sectionbib]{natbib}
\usepackage{chapterbib}

\begin{document}





\include{sample-manuscript}

\include{supp}
\end{document}

%% file: sample-manuscript.tex
\title{Continuous Human Action Recognition for Human-Machine Interaction: A Review}



\author{Harshala~Gammulle,
        David Ahmedt-Aristizabal,
        Simon~Denman,
        Lachlan Tychsen-Smith, \\
        Lars Petersson,
        Clinton~Fookes
\thanks{H. Gammulle, S. Denman, and C. Fookes  are with the Signal Processing, Artificial Intelligence and Vision Technologies (SAIVT) Lab, Queensland University of Technology, Brisbane, Australia.
E-mail: pranali.gammule@qut.edu.au
}
\thanks{D. Ahmedt-Aristizabal, L. Tychsen-Smith and L. Petersson are with the Imaging and Computer Vision group, CSIRO Data61, Canberra,  Australia.}
}









\maketitle

\begin{abstract}
With advances in data-driven machine learning research, a wide variety of prediction models have been proposed to capture spatio-temporal features for the analysis of video streams. Recognising actions and detecting action transitions within an input video are challenging but necessary tasks for applications that require real-time human-machine interaction.
By reviewing a large body of recent related work in the literature, we thoroughly analyse, explain and compare action segmentation methods and provide details on the feature extraction and learning strategies that are used on most state-of-the-art methods. We cover the impact of the performance of object detection and tracking techniques on human action segmentation methodologies. We investigate the application of such models to real-world scenarios and discuss several limitations and key research directions towards improving interpretability, generalisation, optimisation and deployment.
\vspace{-6pt}
\end{abstract}





\section{Introduction}
\label{sec:intro}

\IEEEPARstart{A}{s}
humans, we perform countless daily actions. The intent behind an action may be to achieve a day-to-day task, to convey an idea, or to communicate non-verbally as part of human-to-human interactions. The performance of these actions plays a crucial part in our lives, and occurs (at least in part) automatically. With advancements in robotics and machine learning, there is an increasing interest in developing robots that can work alongside humans. Such a service robot should understand those surrounding actions performed by their human co-workers. Related applications exist within intelligent surveillance, where it is desirable to recognise human behaviour and identify abnormal patterns. To address such scenarios, an action recognition method that is capable of understanding varying complex human actions plays a major role.

Human action recognition (HAR) aims to automatically detect and understand actions of interest (e.g. running, walking, etc) performed by a human subject, using information captured from a camera or other sensors. HAR is a widely investigated problem in the field of computer vision and machine learning. In the past few decades, action recognition has been investigated using both images or videos. Compared to image-based methods which learn actions through only spatial (appearance) features, video-based approaches learn temporal patterns in addition to spatial patterns included in video frames/images. Even though in some instances the actions can be recognised only through spatial information only (\textit{e.g.} ``riding bicycle'' vs ``lifting weight''), many day-to-day actions include similar behavioural patterns (\textit{e.g.} ``walking'' and ``running'') and can only be separated by exploiting temporal information. 


\begin{figure*}[htbp]
    \centering
    \subfigure[][Discrete Action Recognition]{\includegraphics[width=.44
    \linewidth]{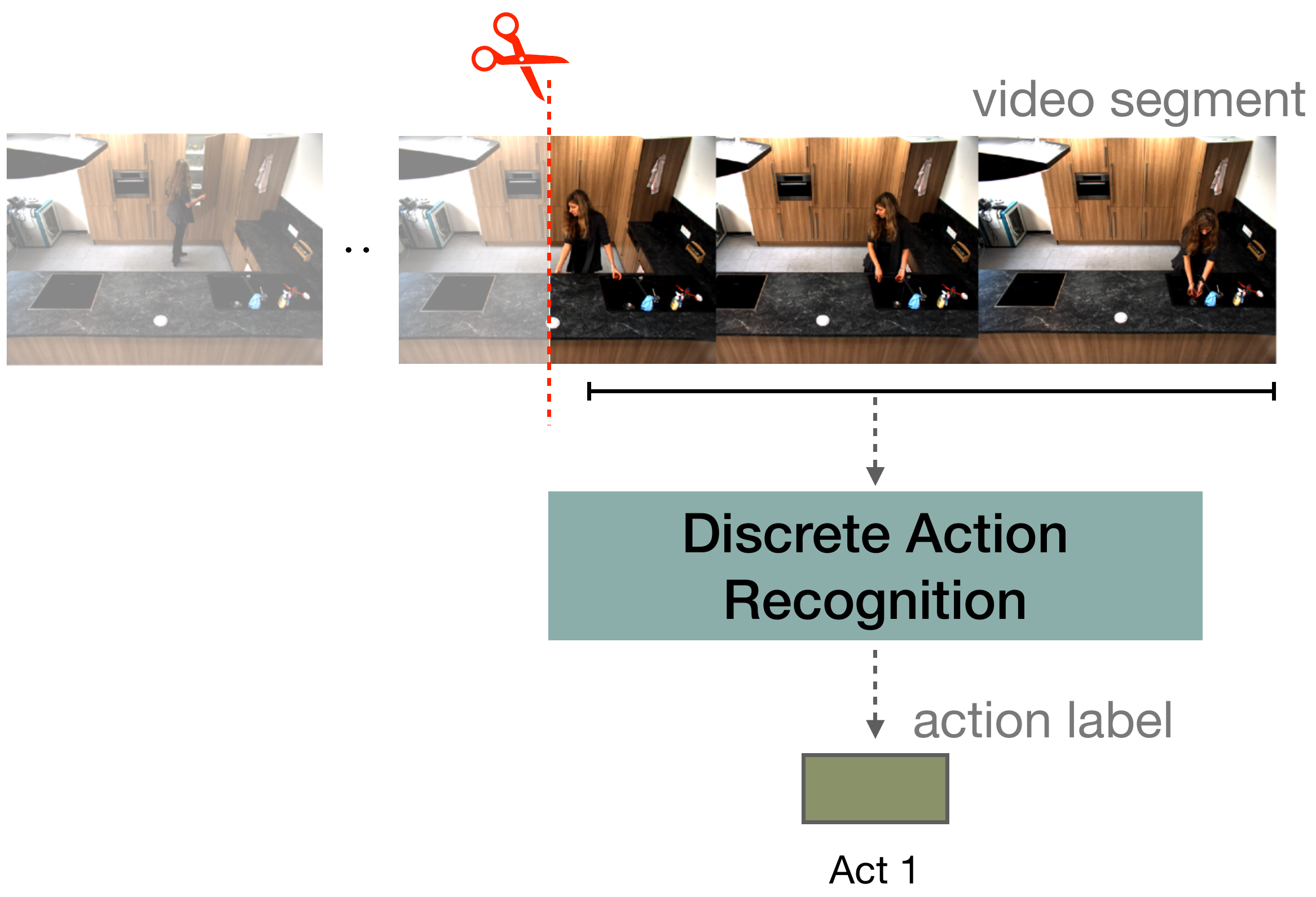}} 
    \subfigure[][Continuous Action Recognition]{\includegraphics[width=.45\linewidth]{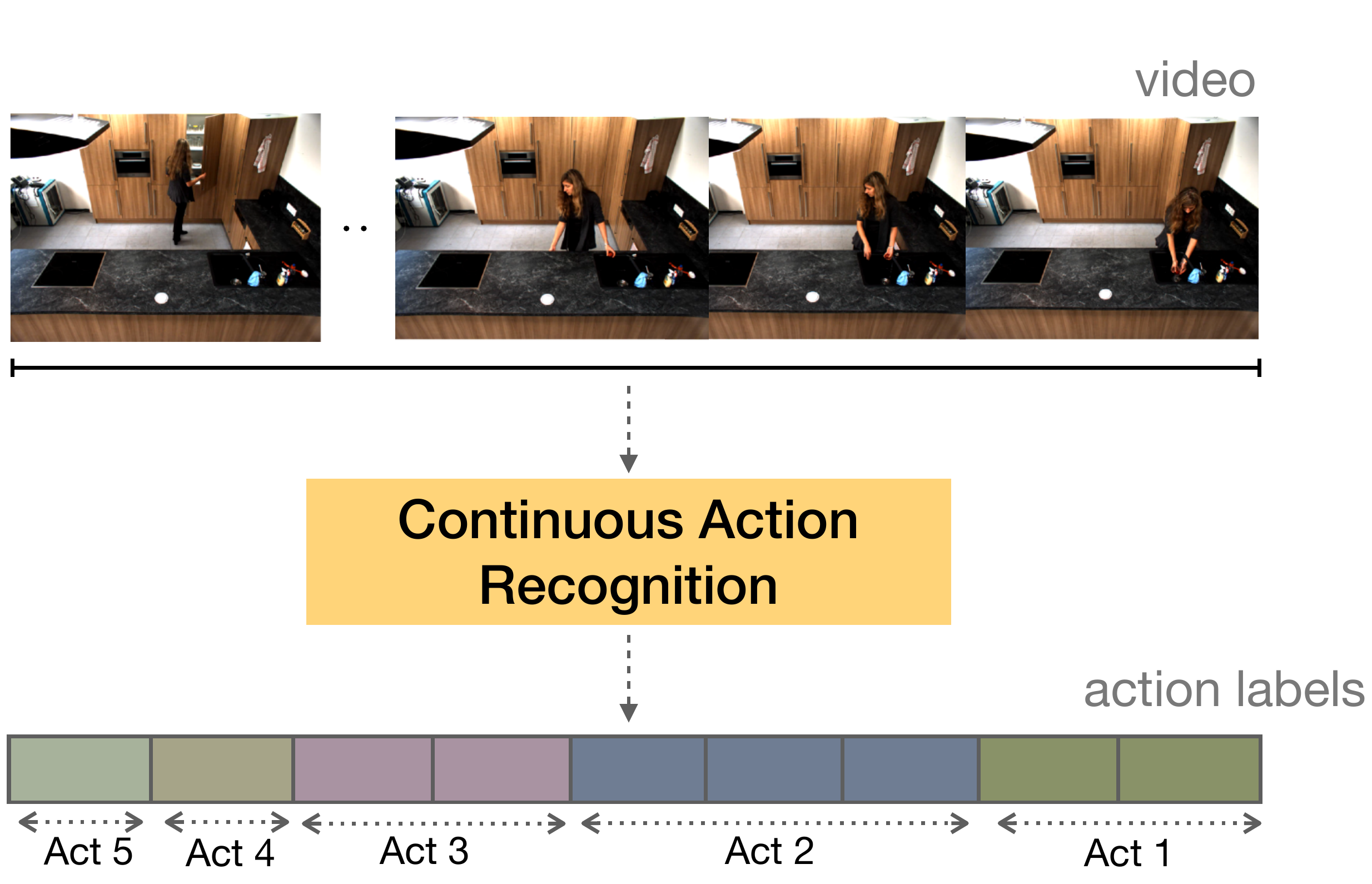}}
    \caption{Discrete and Continuous Action Recognition Comparison.}
    \label{fig:ablation_arc}
\end{figure*}

Video-based action recognition research can be broadly grouped into two types: discrete, and continuous action recognition (see Fig. \ref{fig:ablation_arc}) \cite{kong2018human}. Discrete action recognition uses segmented videos where only a single action is included within the video input. In contrast, continuous action recognition operates over unsegmented videos that contain multiple actions per video. Unlike in discrete action recognition, recognising continuous actions involves not only recognising the actions but also detecting action transitions. Therefore, continuous action recognition methods are well aligned with real-world scenarios where actions are continuous and are related to their surrounding actions. Despite the complexity and challenging nature of continuous action recognition, a considerable amount of research has been conducted into the continuous action recognition task. In this review, our main focus is evaluating recent advancements in continuous action recognition methods, and evaluating them in real-world settings. 

Continuous action recognition methods may also often be referred to as action segmentation and action detection. Both address the same problem scenario, yet the final output takes different forms. As stated in \cite{lea2017temporal}, action segmentation aims to predict the actions that occur at every frame of the video input while action detection aims to determine a sparse set of action segments where each segment includes the start time, end time and action label. In some research, \cite{rohrbach2012database,singh2016multi} action segmentation/ detection is refereed to as fine-grained action recognition, to highlight low variation that exists among classes that are to be detected \cite{rohrbach2012database}. For example, in the MPII cooking dataset, actions such as ``grating'', ``chopping'' and ``peeling'' exhibit fine-grained characteristics.

\begin{figure*}[!t]
    \centering
    \includegraphics[width=0.99\linewidth]{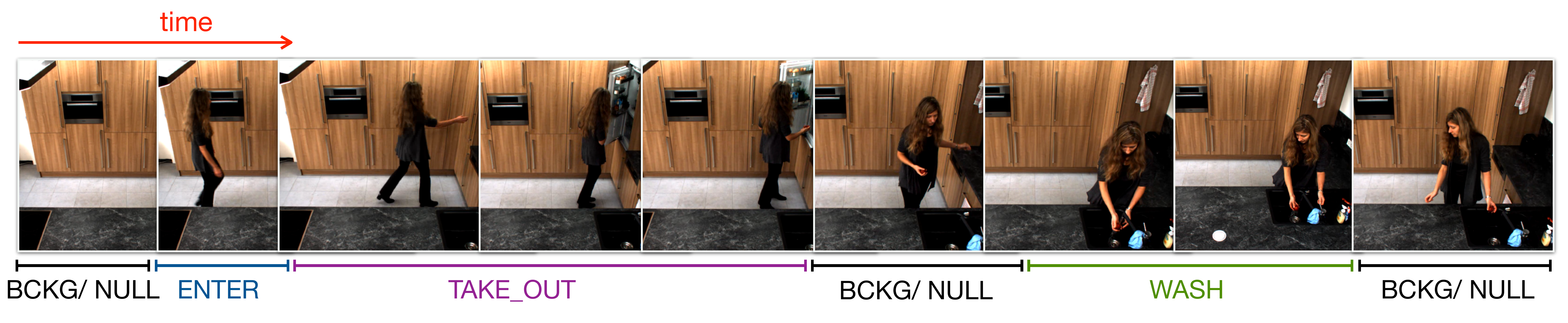}
    \vspace{-6pt}
	\caption{An Example of a Continuous Action Video Sequence: The frames that include action transitions are often labelled as ``background''(BCKG) or ``Null''.}
	\label{fig:act_backg}
    \vspace{-12pt}
\end{figure*}

In addition to the actions of interest, continuous action recognition videos also include frames that contain action transitions. These frames are typically annotated as ``background'' or ``null'', and must also be detected. This is further illustrated through an example action sequence in Fig. \ref{fig:act_backg}. As such, if a dataset contains $N$ actual actions, $N+1$ classes should be detected by the model. However, the background/null examples often have similar characteristics to their surrounding action frames, which makes precise detection of the boundary a challenging task. Regardless of these challenges, to date methods have shown considerable performance, yet there is capacity for methods to be further improved and  utilised in real-world settings.

Earlier work \cite{laptev2007retrieving,patron2010high} on video-based continuous action recognition has focused on detecting actions/interactions in movies; while in \cite{rohrbach2012database} the authors utilised holistic video features based on dense trajectories \cite{denseTraj}, histograms of oriented gradients(HOG)/ histograms of optical flow (HOF) \cite{hoghof}, motion boundary histograms (MBH) \cite{dalal2006human}, and articulated pose tracks motivated from \cite{zinnen2009analysis}. 
Work has also been completed that incorporates object information \cite{ni2014multiple,rohrbach2016recognizing} or utilises grammar-based approaches \cite{pirsiavash2014parsing, kuehne2014language} to model human actions. However, these methods are based on hand-crafted features, which are highly dependent on the knowledge of the human expert and can lack the ability to generalise. 

Therefore, recent efforts have shifted towards deep learning as such models are able to learn action-specific features automatically. While the data driven nature of deep learning methods requires large annotated datasets to train the models, they have nonetheless achieved state-of-the-art performance. For instance, in \cite{wang2015action} the authors improved upon existing hand-crafted approaches by replacing the hand-crafted feature-based model with a CNN model. However, as previously mentioned, irrespective of the technique used it is essential to capture spatio-temporal features when analysing video data. Multiple video-based discrete action recognition methods \cite{simonyan2014two,gkioxari2015finding,gammulle2017two} have used two-stream models to capture spatial and temporal information. However, with continuous action recognition, input sequences are much lengthier in comparison to discrete action methods, and thus require the capture of long-range relationships. 

Early works on action detection \cite{rohrbach2016recognizing,ni2016progressively} proposed sliding window approaches, yet these were unable to capture long-range temporal patterns. A two-stream architecture based on bi-directional LSTMs (Bi-LSTM) that operates over short video chunks was introduced in \cite{singh2016multi}. However, this model is claimed to be time-consuming due to the sequential predictions \cite{farha2019ms}. Considering the limitations of these early methods, a class of time-series models, called Temporal Convolutional Networks (TCN), was introduced for action segmentation and detection in \cite{lea2017temporal}. This model is able to overcome the shortcomings of previous methods and capture long-range temporal patterns through a hierarchy of temporal convolutional filters. The method in \cite{lea2017temporal} employs an encoder-decoder architecture where the encoder is built using temporal convolutions and pooling, while the decoder is composed of upsampling followed by convolutions. The utilisation of temporal pooling enables the capture of long-range temporal dependencies, although the approach may result in the loss of fine-grained details required for action segmentation \cite{farha2019ms}. This model is extended in \cite{lei2018temporal} by using deformable convolutions and adding residual connections in the encoder-decoder model. Motivated by these works, \cite{farha2019ms} proposed a multi-stage TCN architecture which operates over the full temporal resolution, and utilises dilated convolutions to capture long-range temporal patterns. Following these works, a large number of novel models have been introduced for the action segmentation task, and we provide further details in Sec.~\ref{sec:selected_models} and  investigate model capabilities.

\subsection{Our Contributions}

Although there exist numerous recent survey articles \cite{zhang2019comprehensive,liu2019rgb,kong2018human,jegham2020vision} concerning human action recognition and interaction detection, there is no systematic review of human action segmentation methodologies. In particular, to the best of our knowledge, none of the existing literature has investigated the application of state-of-the-art human action segmentation methodologies to real-world scenarios and the extensions that are required to enable this. Such an in-depth analysis would allow readers to compare and contrast the strengths and weaknesses of different deep learning action segmentation models in a real-world context, and select methods accordingly. 
In this paper, we address the limitations of existing surveys (which are summarised in Table~\ref{tab:survey_contributions}) and go beyond a simple feature and neural architecture level comparison of existing human action segmentation methods. We empirically compare the performance of four state-of-the-art deep learning architectures with respect to the utilised feature extraction models. In particular, we evaluate their performance with respect to the size of the extracted feature vector, and assess the possibility of using a compressed feature representation to achieve real-time throughput. Furthermore, we analyse different model training strategies that can be used to augment the feature extraction process, and demonstrate the utility of fine-tuning feature extractors on a sub-set of training data, rather than merely using pre-trained feature extractors.  

Moreover, this paper provides a comprehensive overview of the impact on performance of the hyper-parameters (including the observed sequence length and the feature extraction window size) of human action segmentation methodologies. These hyper-parameters are critical factors when applying such systems in multi-person environments and time-critical applications. 

Lastly, our paper details the limitations of existing human action segmentation methods and lists key research directions to encourage more generalisable, fast, and interpretable methodologies.

\begin{table}[!t]
\centering
\caption{Comparison of Our Review to Other Related Studies.}
\resizebox{0.46\textwidth}{!}{%
\begin{tabular}{|c|ccclc|}
\hline
\multirow{2}{*}{Topic}                               & \multicolumn{5}{c|}{Reference}                                                                                                       \\ \cline{2-6}
                                                     & \multicolumn{1}{c|}{\cite{zhang2019comprehensive}} & \multicolumn{1}{c|}{\cite{liu2019rgb}} & \multicolumn{1}{c|}{\cite{kong2018human}} & \multicolumn{1}{l|}{\cite{jegham2020vision}} & Ours \\ \hline
Feature Extractors                                   & \multicolumn{1}{c|}{\checkmark} & \multicolumn{1}{c|}{\checkmark} & \multicolumn{1}{c|}{\checkmark}        & \multicolumn{1}{l|}{\checkmark}        &    \checkmark      \\ \hline
Action Segmentation                                  & \multicolumn{1}{c|}{}          & \multicolumn{1}{c|}{}            & \multicolumn{1}{c|}{}        & \multicolumn{1}{l|}{}          & \checkmark  \\ \hline
Object detection and tracking for action segmentation & \multicolumn{1}{c|}{}         & \multicolumn{1}{c|}{}            & \multicolumn{1}{c|}{}        & \multicolumn{1}{l|}{}          & \checkmark  \\ \hline
Discussion of Challenges with real world Application  & \multicolumn{1}{c|}{}         & \multicolumn{1}{c|}{\checkmark}  & \multicolumn{1}{c|}{}        & \multicolumn{1}{l|}{\checkmark} & \checkmark \\ \hline
Analysis of extensions for real-time response         & \multicolumn{1}{c|}{}         & \multicolumn{1}{c|}{}            & \multicolumn{1}{c|}{}        & \multicolumn{1}{l|}{}          & \checkmark  \\ \hline
Methods for handling multiple-person data            & \multicolumn{1}{c|}{}         & \multicolumn{1}{c|}{}            & \multicolumn{1}{c|}{}        & \multicolumn{1}{l|}{}          & \checkmark  \\ \hline
\end{tabular}}
\label{tab:survey_contributions}
\vspace{-6pt}
\end{table}

\subsection{Organisation}

In Sec. \ref{sec:act_seg} we outline the traditional action segmentation pipeline that most state-of-the-art action segmentation methods are built upon, while providing a detailed review of each step of the pipeline. 
Specifically, in Sec. \ref{sec:feature_extraction} we discuss in detail different backbone models that are used for the feature extraction step. 
In Sec. \ref{sec:network_learning} we provide an overview of different networks and learning strategies that are used in the existing action segmentation methods, while providing a summary of each technique utilised. 
Sec. \ref{sec:selected_models} provides a detailed description of the action segmentation models that we selected for further experiments.
Sec.~\ref{sec:dettrack_actionrec} introduces the benefits of incorporating detection and tracking techniques to aid action segmentation. Recent advances in object detection and multi-object tracking are systematically discussed in Sec. \ref{sec:object_det} and Sec. \ref{sec:multi-track}, respectively. 
%
%
In Sec. \ref{sec:adapting_real}, we investigate the application of the action segmentation, detection and tracking methods to real-world applications.
Limitations and future directions for research conclude this paper, and are presented in Sec. \ref{sec:limitations}.

\section{Human Action Segmentation}
\label{sec:act_seg}

\begin{figure*}[ht!]
    \centering
    \includegraphics[width=0.65\linewidth]{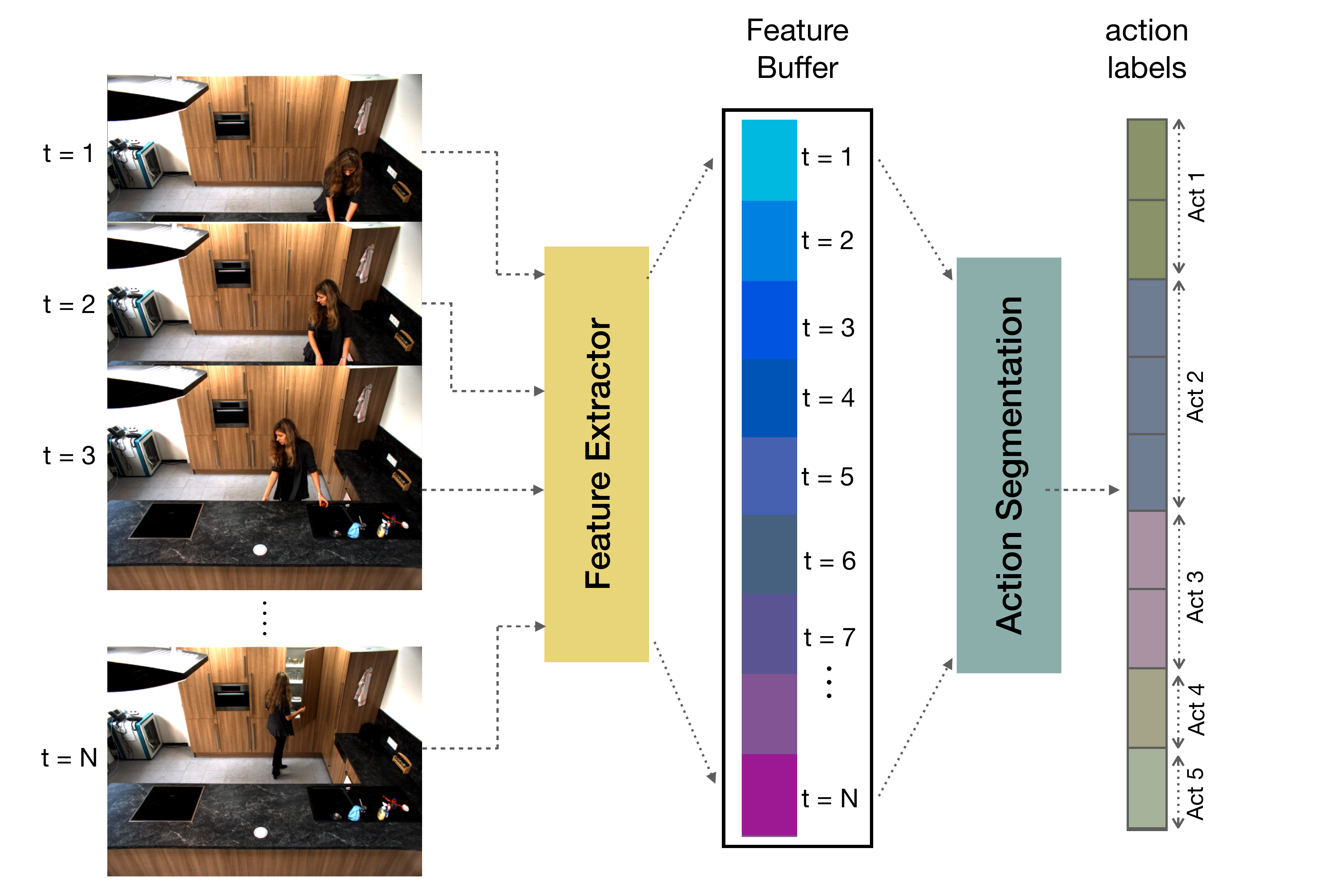}
	\caption{Action Segmentation Pipeline.}
	\label{fig:gen_framework}
\vspace{-8pt}
\end{figure*}

Continuous action recognition methods aim to recognise and localise actions in an unsegmented video stream. Fig. \ref{fig:gen_framework} illustrates the general pipeline of the state-of-the-art action segmentation models. Generally, the first step is to extract salient semantic features from each frame of the video input. These features are saved in a feature buffer and are passed through the action segmentation model. Finally, the action segmentation model outputs a sequence of action predictions, one per frame of the input video stream.

In the following sections, we provide a more detailed explanation of these steps in the action segmentation pipeline. 

\subsection{Feature Extraction}
\label{sec:feature_extraction}

The feature extraction step plays a major role in the action segmentation task, encoding input frames and extracting salient information for the subsequent action recognition step. Ideally, the encoding step should ensure that only relevant information is retained to help avoid confusion within the action recognition. As such, the choice of feature extractor plays a major role in the overall action segmentation performance. Recently, a popular approach has been to use pre-trained CNNs such as VGG~\cite{simonyan2014very}, GoogleNet~\cite{szegedy2015going}, Residual Neural Network (ResNet)  \cite{he2016deep}, EfficientNet-B0 \cite{tan2019efficientnet}, MobileNet-V2 \cite{sandler2018mobilenetv2} and Inflated 3D ConvNet (I3D) \cite{carreira2017quo} as feature extraction methods. We refer interested readers to Appendix A of supplementary material where we discuss the architectures of ResNet, EfficientNet-B0, MobileNet-V2 and I3D models in detail.

\subsection{Networks and Learning Strategies}
\label{sec:network_learning}

In this section, we provide a brief theoretical overview regarding recent advancements in network architectures and learning strategies used in state-of-the-art action segmentation methods. An in depth discussion of the state-of-the-art action segmentation models is subsequently provided in Sec. \ref{sec:selected_models}.

\subsubsection{Temporal Convolutional Networks (TCN)}
As explained in Sec. \ref{sec:intro}, the action segmentation task involves analysing a sequence of frames to recognise actions as they evolve over time. Therefore, an action segmentation model must be able to capture temporal patterns. In many previous approaches on action segmentation and detection, Recurrent Neural Networks (RNN) such as Long Short-Term Memory (LSTM)~\cite{greff2016lstm} and Gated Recurrent Units (GRU)~\cite{cho2014learning} are widely utilised. While these can capture temporal patterns, they are unable to attend to the entire sequence as they maintain a fixed size memory. In contrast, Temporal Convolutional Networks (TCN) have the ability to analyse the entire sequence at once, and have a lower computational cost compared to recurrent models. Therefore, TCNs have been widely used in action segmentation research \cite{lea2017temporal,farha2019ms,gammulle2021tmmf}.

A TCN uses a 1D (1 dimensional) fully-convolutional network architecture. Each hidden layer outputs a representation the same length as the input layer, with zero padding of length (kernal\_size - 1) used to maintain the equal length of each layers. As such, the network can take an input sequence of any length and after processing obtain an output with the same length, aligning well with the action segmentation task.

\begin{figure*}[!t]
    \centering
    \subfigure[][TCN operation on an input with a single channel.]{\includegraphics[width=.45
    \linewidth]{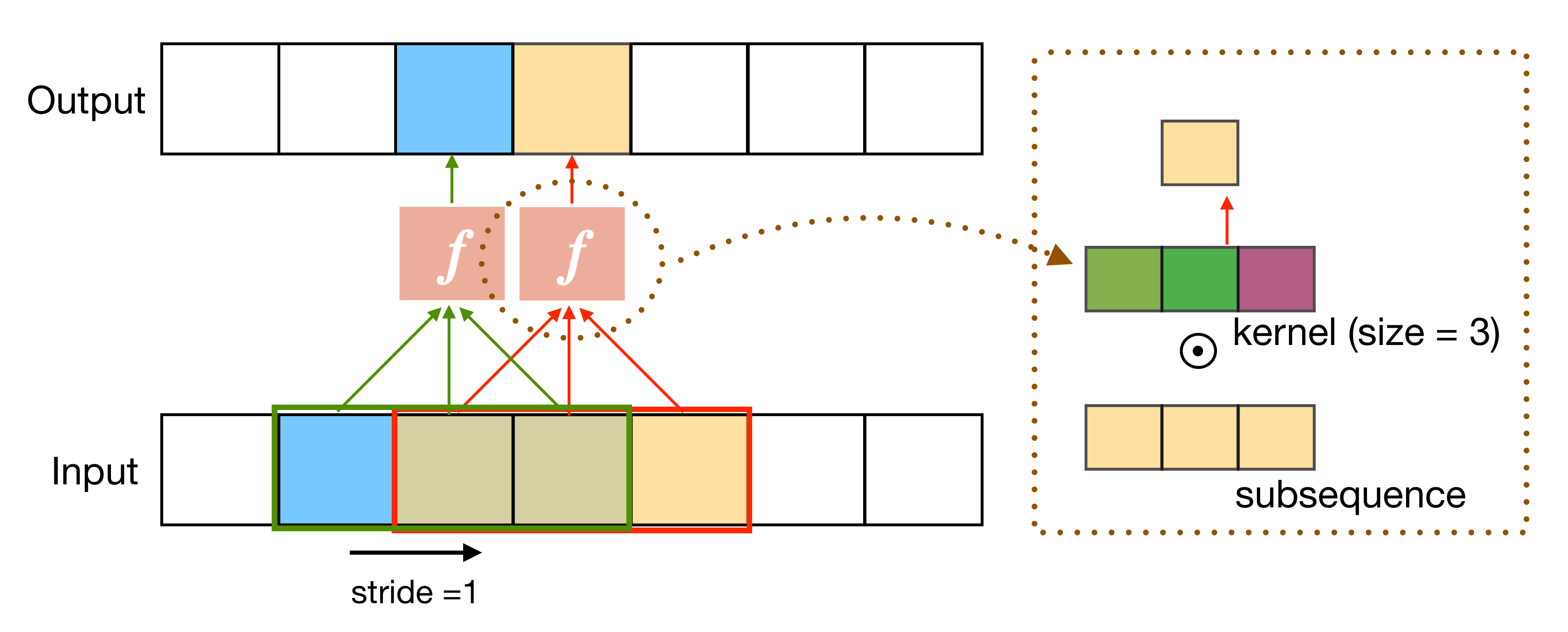}} 
    \subfigure[][TCN operation on an input with multiple channels.]{\includegraphics[width=.43\linewidth]{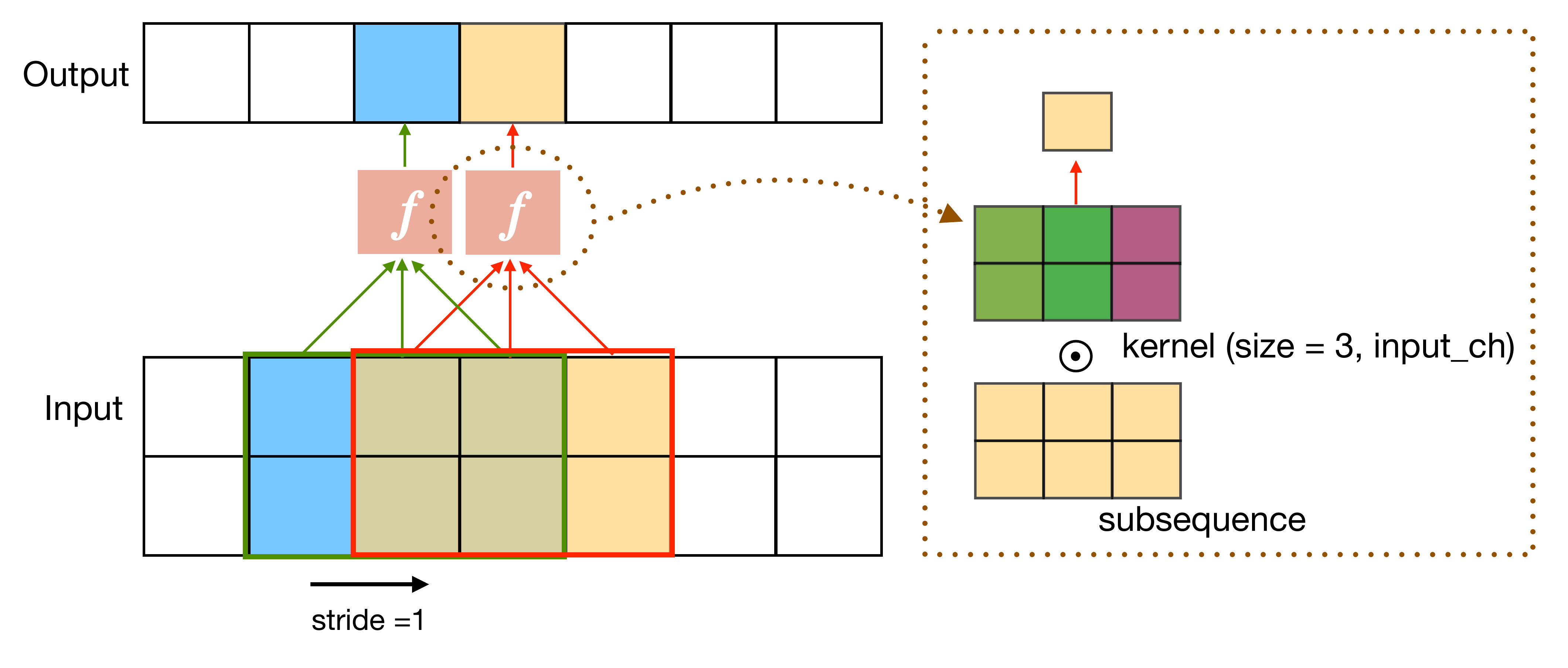}}
    \vspace{-4pt}
    \caption{Temporal Convolutional Networks (TCN).}
    \label{fig:TCN_1}
\vspace{-8pt}
\end{figure*}

As illustrated in Fig. \ref{fig:TCN_1}, in a TCN the 1D convolutions are applied across the temporal axis. The network receives a 3-dimensional tensor of shape [batch\_size, sequence\_length, input\_channels], and outputs a tensor of shape [batch\_size, sequence\_length, output\_channels], obtained through 2-dimensional kernels. Fig. \ref{fig:TCN_1} (a) visualises a scenario where the input\_channels=1, while Fig. \ref{fig:TCN_1} (b) illustrates a scenario where the input tensor contains more than channel (input\_channels = 2). At each step, the output is calculated by obtaining the dot product between the sub-sequence that lies within the sliding window and the learned weights of the kernel vector. In order to obtain the next output element, the window is shifted to the right by a one input element (\textit{i.e.} stride = 1). This process is repeated across the entire sequence, generating an output sequence that has the same length as the input. Therefore, the prediction at each frame is a function of a fixed-length period of time (i.e. the receptive field), such that a subset of input elements impact a specific output element. This receptive field varied according to requirements of the problem. 

One effective way to increase the receptive filed while maintaining a relatively small number of layers is to utilise dilation. As defined in \cite{oord2016wavenet}, dilated convolution (also known as convolution with holes) is a convolution operation where the filter is applied over an area greater than the filter length by skipping input elements with a desired step, based on the dilation factor. 
The dilatation rate allows the network to maintain the temporal order of the samples, yet capture long-term dependencies without an explosion in model complexity.
%
A layer with a dilation factor of $d$ and a kernel with a size of $k$ has a receptive field size spreading over a length of $l$, where $l$ can be defined as, 
\begin{equation}
    l = 1 + d \times (k-1).
\end{equation}

\begin{figure*}[t!]
    \centering
    \includegraphics[width=0.45\linewidth]{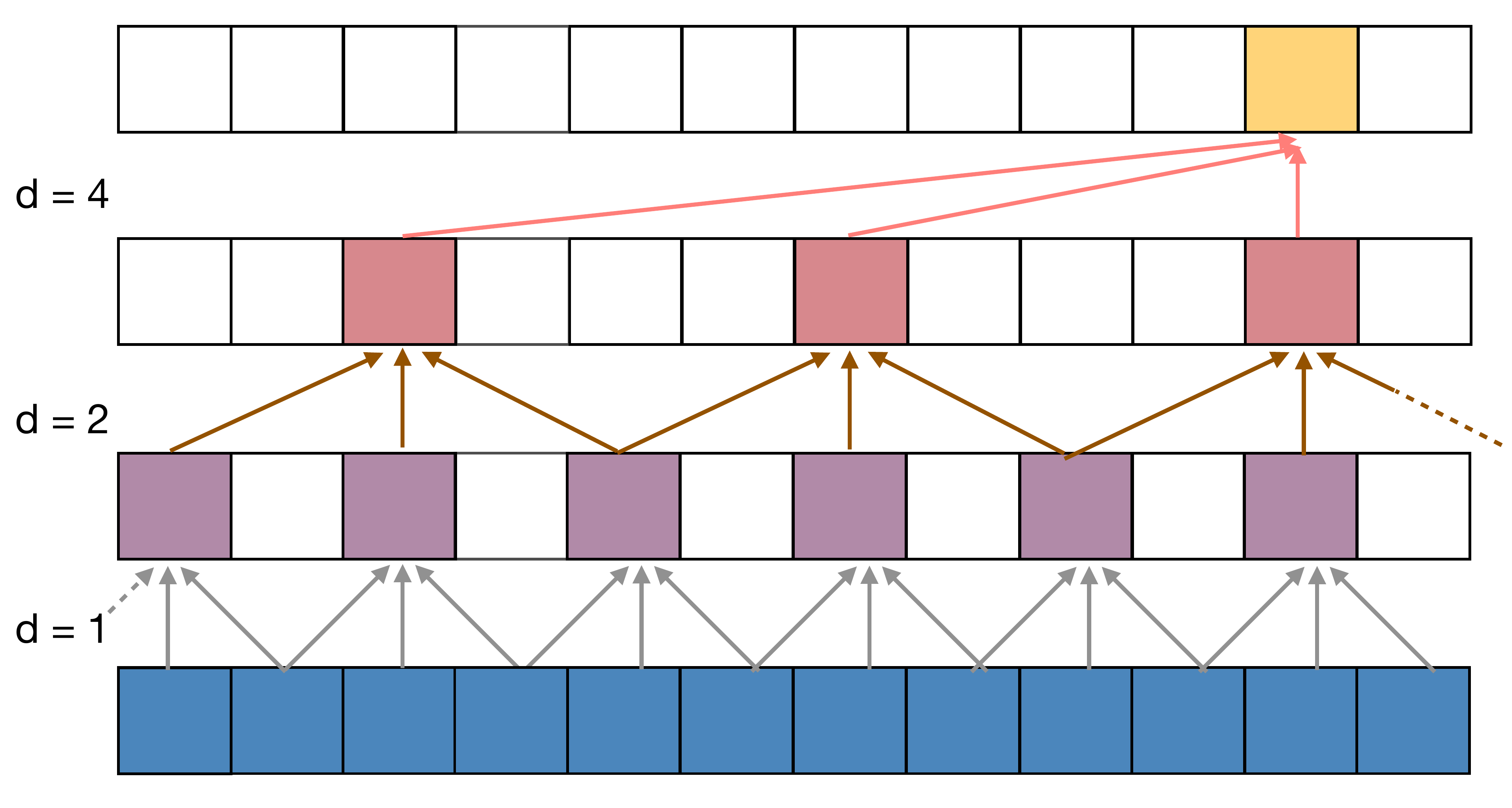}
    \vspace{-3pt}
	\caption{Dilation operations with different dilation factors ($d$) with $kernel_size = 3$. In the case where $d=1$, 3 adjacent input elements are chosen to compute a particular output element by setting the receptive field size to 3. When $d=2$ the receptive field is increased to length 5. When $d=4$, the receptive field expands to a length of 9.}
	\label{fig:dilation}
    \vspace{-10pt}
\end{figure*}

As shown in Fig. \ref{fig:dilation}, when $d=1$ and $kernel\_size=3$, the operation takes the form of a standard convolutional layer where the input elements chosen to calculate an output element are adjacent and the receptive field is 3 elements. In the case where $d=2$, the receptive field expands to spread across 5 elements, and when $d=4$ the receptive field spreads across 9 elements. In practice the dilation factor is increased exponentially resulting in an increase of the receptive field at each layer. For example, in \cite{farha2019ms}, when the $kernel\_size = 3$, the receptive field ($R\_field^l$) at layer l ($l \epsilon [1,l]$) is decided based on the formula,
\begin{equation}
    R\_filed^l = 2^{l+1} - 1.
\end{equation}

As a result of this very large receptive field however is that networks are limited to having only a few layers to avoid over-fitting.

Recent works \cite{bai2018empirical,farha2019ms} have shown that by adding residual connections, TCN models can be further improved. In particular, this has benefited deeper networks by facilitating gradient flow. A residual block is composed of a series of transformations and its outputs are added to the input of the block. This allows the layer to learn suitable modifications to the representation, rather than needing to learn a complete transform. 

\subsubsection{Generative Adversarial Networks (GAN)}

A standard Generative Adversarial Network (GANs) \cite{goodfellow2014generative} is able to learn a mapping from a random noise vector $z$ to an output vector $y$. This is achieved through two networks, the ``Generator'' (G) and the ``Discriminator'' (D), which compete in a two player min-max game. G seeks to learn the input data distribution, while D estimates the authenticity of an input (real/generated). Both models, G and D, are trained simultaneously. 

\begin{figure*}[ht!]
    \centering
    \includegraphics[width=0.5\linewidth]{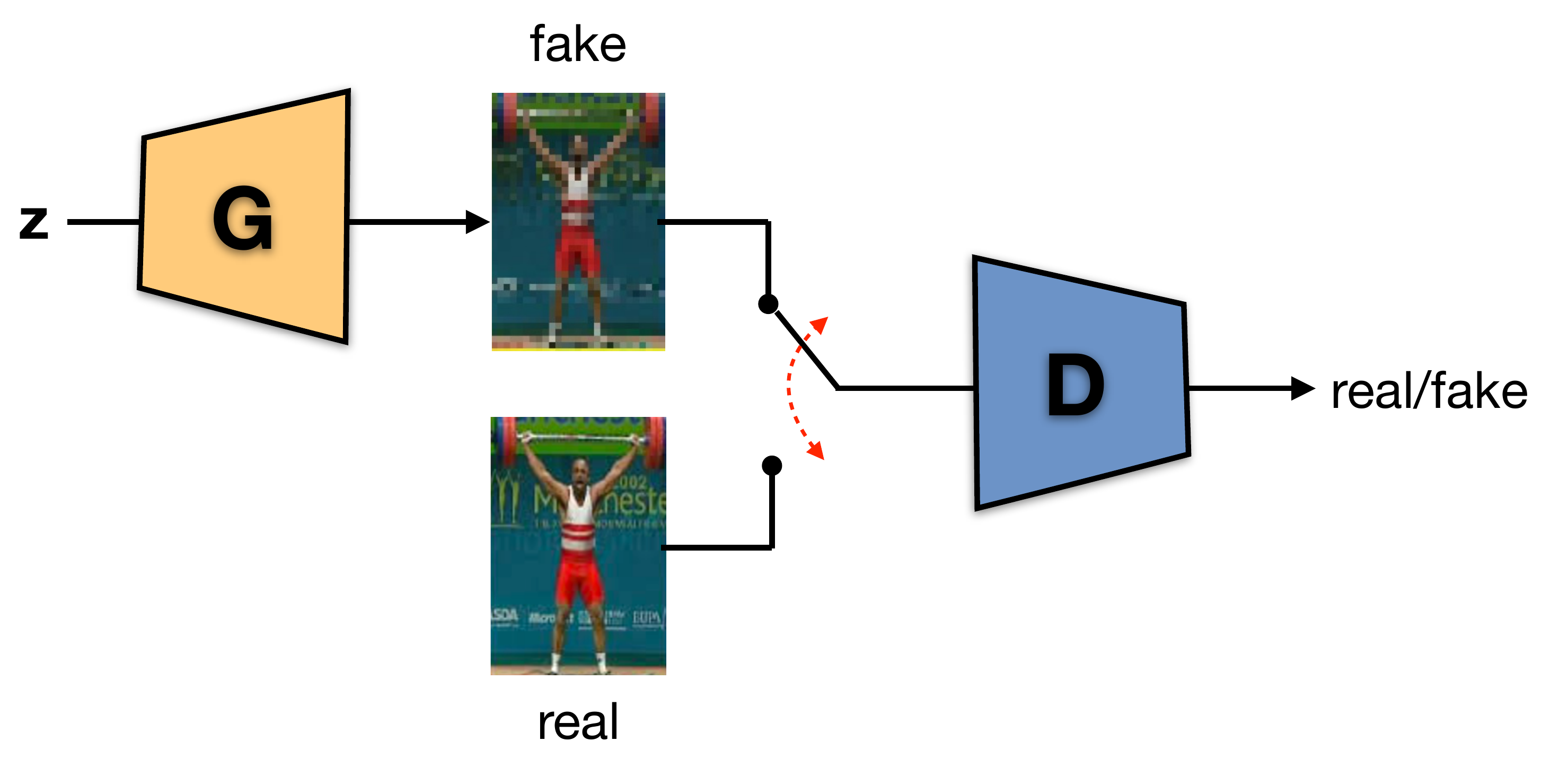}
    \vspace{-8pt}
	\caption{Generative Adversarial Networks (GAN)}
	\label{fig:GAN}
    \vspace{-10pt}
\end{figure*}

Fig. \ref{fig:GAN} illustrates the GAN training procedure. Noise sampled from $P_z(z)$ is fed to $G$. G aims to learn the data distribution of the target data, $P_{data}(x)$, by mapping from the noise space to the target data space. $D$ seeks to output a scalar variable when given an input sample which may be generated by G (fake), or be a real example. The discriminator is trained to perform the real/fake classification. The GAN objective can be written as,
\begin{equation}
\resizebox{0.48\textwidth}{!}{$\min_{G}\max_{D}V(D, G) = \mathbb{E}_{x \sim P_{data}(x)}[log D(x)] + \mathbb{E}_{z \sim P_{z}(z)}[log 1 - D(G(z))] $}. 
\end{equation}

The GAN learning process is considered to be unsupervised. During learning, $D$ is provided with a supervision signal (real/fake ground truth labels), though $G$ is not provided with any labelled data. However, the primary task that the GANs seek to address is to learn the data distribution (i.e. the objective of G), therefore, the overall process is considered to be unsupervised. 

In \cite{mirza2014conditional}, the authors extend the GAN framework and introduced the conditional GAN (cGAN). The generator and discriminator outputs of the cGAN are conditioned on additional data, $c$. The cGAN objective is defined as,
\begin{equation}
\resizebox{0.48\textwidth}{!}{$\min_{G}\max_{D}V(D, G) = \mathbb{E}_{x \sim P_{data}(x)}[log D(x | c)] + \mathbb{E}_{z \sim P_{z}(z)}[log 1 - D(G(z | c) | c)] $}. 
\end{equation}

Following \cite{isola2017image}, the cGAN can be considered a general-purpose image-to-image translation method. Through utilising a cGAN framework, a network can learn a mapping from an input image to an output image while learning a loss formulation to do this mapping. Prior works using image-based conditional models have widely investigated image generation problems including future frame prediction \cite{mathieu2015deep}, product photo generation \cite{yoo2016pixel}, and photographic text-to-image generation \cite{hu2021novel,qi2021pccm}. Furthermore, multiple methods have adapted the cGAN architecture for action recognition \cite{gedamu2021arbitrary,gammulle2019coupled,gammulle2020fine} and prediction \cite{gammulle2019predicting} tasks.    

\subsubsection{Domain Adaptation}

Domain adaptation is used to address scenarios in which a model is employed on a target distribution which is different (but related) to a source data distribution on which the model was trained. In general, domain adaptation strategies can be divided into: (i) discrepancy-based domain adaptation; (ii) adversarial-based domain adaptation; and (iii) reconstruction-based domain adaptation \cite{wang2018deep}. 

In Discrepancy-based methods, a divergence criterion between the source and target domain is minimised such that a domain-invariant feature representation can be obtained. 
Popular domain discrepancy measures include Maximum Mean Discrepancy (MMD) \cite{gretton2012kernel}, Correlation Alignment (CORAL) \cite{sun2017correlation}, Contrastive Domain Discrepancy (CDD) \cite{kang2019contrastive} and the Wasserstein metric \cite{shen2018wasserstein}. When comparing two inputs using MMD, we first transform the feature maps to a latent space where the similarity between the two feature maps is measured based on the mean embedding of the features. In contrast, CORAL utilises second-order statistics (correlation) between the source and target domains \cite{rahman2020correlation}. In CDD, class level domain discrepancies are considered where the intraclass discrepancies are minimised and the domain discrepancy between different classes is maximised \cite{kang2019contrastive}. For the Wasserstein metric, the Wasserstein distance measure is used to evaluate domain discrepancies.

Adversarial-based approaches use generative models to minimise domain confusion. In the CoGAN architecture \cite{liu2016coupled}, two generator/discriminator pairs are used and some weights are shared such that a domain-invariant feature space can be learned. 

Following a similar line of work, in \cite{ajakan2014domain} a domain confusion loss is introduced in addition to the model's primary loss (i.e. classification loss), and they seek to make samples from both domains mutually indistinguishable for the classifier. 

In contrast, in the reconstruction-based domain adaptation approach an auxiliary reconstruction task is created such that the shared representation between the source and target domain can be learned by the network through jointly training both primary and auxiliary (reconstruction) tasks. An example of this paradigm is \cite{ghifary2016deep}, where the authors propose the two tasks of classifying the source data and reconstructing the unlabelled target data. Another popular reconstruction-based approach is to employ Cycle GANs \cite{zhu2017unpaired}, where data is translated between domains. 

\subsubsection{Graph Convolution Networks}

In a conventional convolution layer, the input is multiplied by a filter or kernel, which is defined by a set of weights. This filter slides across the input representation, generating an activation map. In Graph Convolution Networks (GCNs), the same concept is applied to irregular or non-structured data, rather than the regular Euclidean grids over which CNNs operate. Hence, GCNs can be seen as a generalised form of CNNs. 

Traditional CNNs analyse local areas based on fixed connectivity (determined by the convolutional kernel), potentially limiting performance and leading to difficulty in interpreting the structures being modelled, particularly  when the relationships being modelled cannot be easily fitted to a regular grid. Graphs, on the other hand, offer more flexibility to analyse unordered data by preserving neighbouring relations.
Similar to CNNs, GCNs learn  abstract  feature  representations for each  feature at a node via message passing, in which nodes successively  aggregate feature vectors from their neighbourhood to compute a new feature vector at the next hidden layer in the network.

GCNs can be broadly categorised into two classes: spectral GCNs and spatial GCNs. In Spectral GCN approaches the graph convolution operation is formulated based on graph signal processing, while in spatial GCNs it is formulated as aggregating information from neighbours~\cite{wu2020comprehensive}. The popularity of spatial GCNs has rapidly grown over the years compared to its spectral counterpart, despite the solid mathematical foundation of spectral GCNs. 
One of the major drawbacks of spectral GCNs is the requirement that the entire graph be processed simultaneously, which is impractical for large graphs with millions of connections. In contrast, in spatial GCNs operations can be performed locally. Furthermore, the assumption of a fixed graph in spectral GCNs leads to poor generalisation \cite{wu2020comprehensive}.

\begin{table*}[!t]
\centering
\caption{Performance of SOTA action segmentation models. The results for the  EPIC-Kitchens are obtained from \cite{huang2020improving}.}
\resizebox{0.95\textwidth}{!}{
\begin{tabular}{ccccccccccc}
\hline
\multirow{2}{*}{Method} & \multicolumn{5}{c}{50 Salads ~\cite{stein2013combining}}                                     & \multicolumn{5}{c}{EPIC-Kitchens ~\cite{Damen2021RESCALING}}                                                                         \\
                                 & \textbf{Acc} & \textbf{Edit} & \textbf{F1@0.1} & \textbf{F1@0.25} & \textbf{F1@0.5} & \textbf{Acc} & \textbf{Edit} & \textbf{F1@0.1} & \multicolumn{1}{l}{\textbf{F1@0.25}} & \multicolumn{1}{l}{\textbf{F1@0.5}} \\ \hline
Bi-LSTM                          & 55.7         & 55.6          & 62.6            & 58.3    & 47.0   & 43.3         & 29.1          & 19.0            & 11.7                        & 5.0                        \\
ED-TCN~\cite{lea2017temporal}                           & 64.7         & 59.8          & 68.0            & 63.9    & 52.6   & 42.9         & 23.7          & 21.8            & 13.8                        & 6.5                        \\
SS-AGAN~\cite{gammulle2020fine}                          & 73.3         & 69.8          & 74.9            & 71.7    & 67.0   & -            & -             & -               & -                           & -                          \\
Coupled-AGAN~\cite{gammulle2019coupled}                     & 74.5         & 76.9          & 80.1            & 78.7    & 71.1   & -            & -             & -               & -                           & -                          \\
MS-TCN ~\cite{farha2019ms}                           & 80.7         & 67.9          & 76.3            & 74.0    & 64.5   & 43.6         & 25.3          & 19.4            & 12.3                        & 5.7                        \\
GTRM~\cite{huang2020improving}                             & -            & -             & -               & -       & -      & 43.4         & 42.1          & 31.9            & 22.8                        & 10.7                       \\
DTGRM~\cite{wang2021temporal}                            & 80.0         & 72.0          & 79.1            & 75.9    & 66.1   & -            & -             & -               & -                           & -                          \\
Global2Local~\cite{gao2021global2local}                     & 82.2         & 73.4          & 80.3            & 78.0    & 69.8   & -            & -             & -               & -                           & -                          \\
SSTDA~\cite{chen2020action}                            & 83.2         & 75.8          & 83.0            & 81.5    & 73.8   & -            & -             & -               & -                           & -                          \\
MSTCN + HASR~\cite{ahn2021refining}                     & 81.7         & 77.4          & 83.4            & 81.8    & 71.9   & -            & -             & -               & -                           & -                          \\
SSTDA + HASR~\cite{ahn2021refining}                     & 83.5         & 82.1          & 74.1            & 77.3    & 82.7   & -            & -             & -               & -                           & -                          \\
MS-TCN++~\cite{li2020ms}                         & 83.7         & 74.3          & 80.7            & 78.5    & 70.1   & -            & -             & -               & -                           & -                          \\
BACN~\cite{wang2020boundary}                             & 84.4         & 74.3          & 82.3            & 81.3    & 74.0   & -            & -             & -               & -                           & -                          \\ \hline
\end{tabular}}
\vspace{-4pt}
\label{table:SOTA-models}
\end{table*}

With the network structure and node information as inputs, GNNs can be formulated to perform various graph analytic tasks such as:
\begin{itemize}
    \item \textit{Node-level prediction}: A GNN operating at the node-level computes values for each node in the graph and is thus useful for node classification and regression.
    In node classification, the task is to predict the node label for every node in the graph. To compute node-level predictions, the node embedding is fed to a Multi-Layer Perceptron (MLP). 

    \item \textit{Graph-level prediction}: GNNs that predict a single value for an entire graph perform a graph-level prediction. This is commonly used to classify entire graphs, or compute similarities between graphs.
    To compute graph-level predictions, the same node embedding used in node-level prediction is input to a pooling process followed by a separate MLP. 
\end{itemize}

One of the key reasons to apply GCNs to action segmentation is their ability to simultaneously handle spatial and temporal information. In spatio-temporal GCNs, the node and/or edge structure changes over time, allowing them to seamlessly model both spatial and temporal relations in the data. For example, location nodes could simply represent the spatial dependencies of the observations, such as distances between pairs of sensors, while by using recurrent neural networks to model the nodes one can represent the temporal evolution of the observations, generating a spatio-temporal graph.

\subsection{Action Segmentation Models}
\label{sec:selected_models}

In the following subsections we discuss recent action segmentation models in detail. In this section we focus on end-to-end models, that receive the raw data as input and perform segmentation. A growing number of methods seek to optimise either the results or even the structure of these networks, and these are discussed in Sec. \ref{sec:selected_models_2}.

\subsubsection{Temporal Convolutional Network for Action Segmentation and Detection \cite{lea2017temporal}}

This work can be considered one of the first to utilise TCNs for human action segmentation. The authors proposed two TCN based architectures: Encoder-Decoder TCN (ED-TCN) and Dilated TCN. The ED-TCN model has temporal convolution, pooling and upsampling operations. This network is relatively shallow (only 3 layers in the encoder) compared to prior networks that were proposed to segment human actions. However, due to the length of the 1D convolution filters used the network was able to capture long-term temporal dependencies in the input feature sequences. In the 2nd architecture, Dilated TCN, the authors have removed the pooling and upsampling operations and have used dilated convolution filters. This allowed the Dilated TCN network to capture long-term temporal patterns as well as pairwise transitions between the action segments.
The authors have utilised off the shelf categorical cross entropy to train the model, and evaluations were conducted on 3 datasets (50 Salds~\cite{fathi2011learning}, MERL shopping~\cite{singh2016multi}, and GTEA~\cite{fathi2011learning}). The ED-TCN model showed promising performance on all 3 datasets. Specifically, it achieved 64.7\% frame wise action classification accuracy and 59.8\% Edit score and 52.6 F1@50 on the 50 Salads dataset. In their empirical comparisons, the authors demonstrate that this architecture is not only capable of outperforming recurrent architectures such as Bi-LSTM \cite{singh2016multi}, but it is also faster to train. For instance, a typical Bidirectional LSTM architecture only achieves 55.7\% frame wise action classification accuracy and 55.6\% Edit score and 47.0 F1@50 on the 50 Salads dataset, despite its greater computational burden.

\subsubsection{Multi-Stage Temporal Convolutional Network (MS-TCN) \cite{farha2019ms} }
\label{sec:mstcn_info}

\begin{figure*}[!t]
    \centering
    \includegraphics[width=0.5\linewidth]{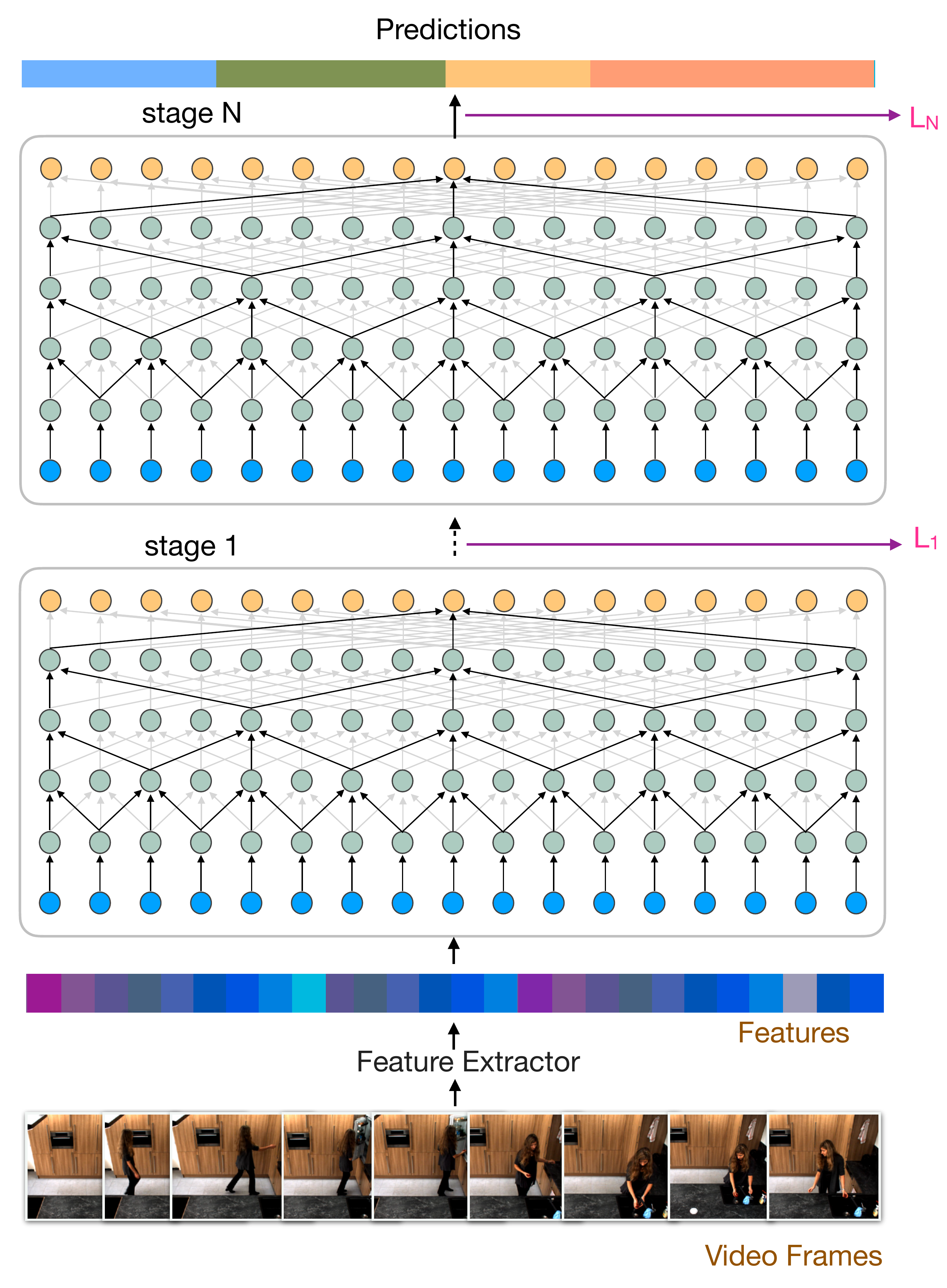}
    \vspace{-6pt}
	\caption{Multi-Stage Temporal Convolutional Network (MS-TCN) architecture, recreated from \cite{farha2019ms}.}
	\label{fig:mstcn_arc}
    \vspace{-12pt}
\end{figure*}

Motivated by the tremendous success of the TCN architecture of \cite{lea2017temporal}, Farha et al. proposed a multi-stage TCN model which stacks multiple single-stage TCN blocks. This was proposed to address the limited temporal resolution offer by the TCN convolution operation. MS-TCN addresses this limitation by aggregating temporal information across multiple-stages. Fig. \ref{fig:mstcn_arc} illustrates the multi-stage architecture of MS-TCN.

\begin{figure*}[t!]
    \centering
    \subfigure[][Dilated Residual Layer.]
    {\includegraphics[width=.20 \linewidth]{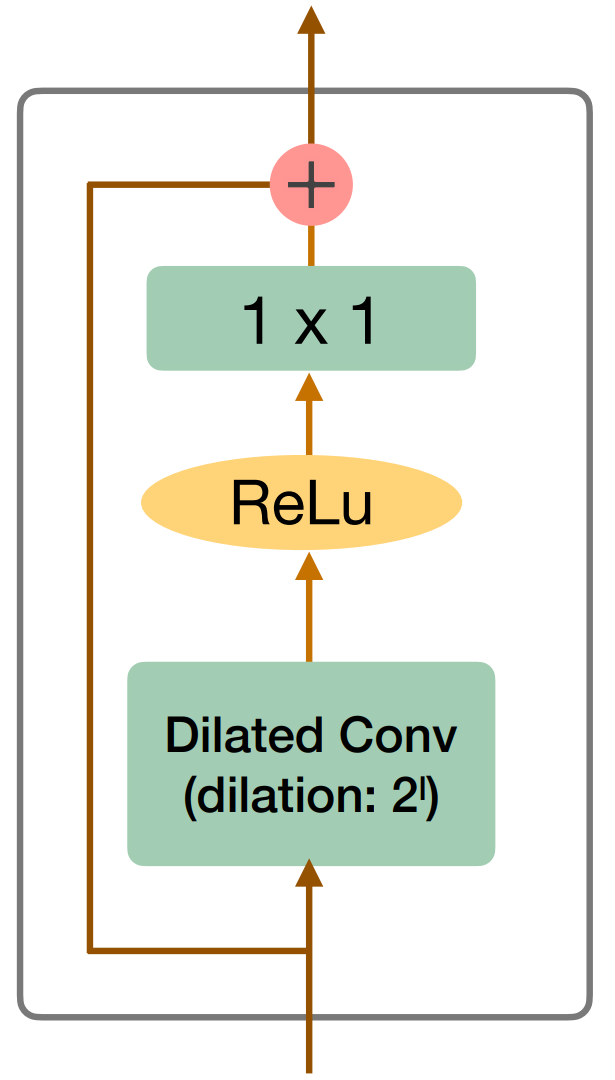}} 
    \subfigure[][Dual Dilated Residual Layer.]
    {\includegraphics[width=.38\linewidth]{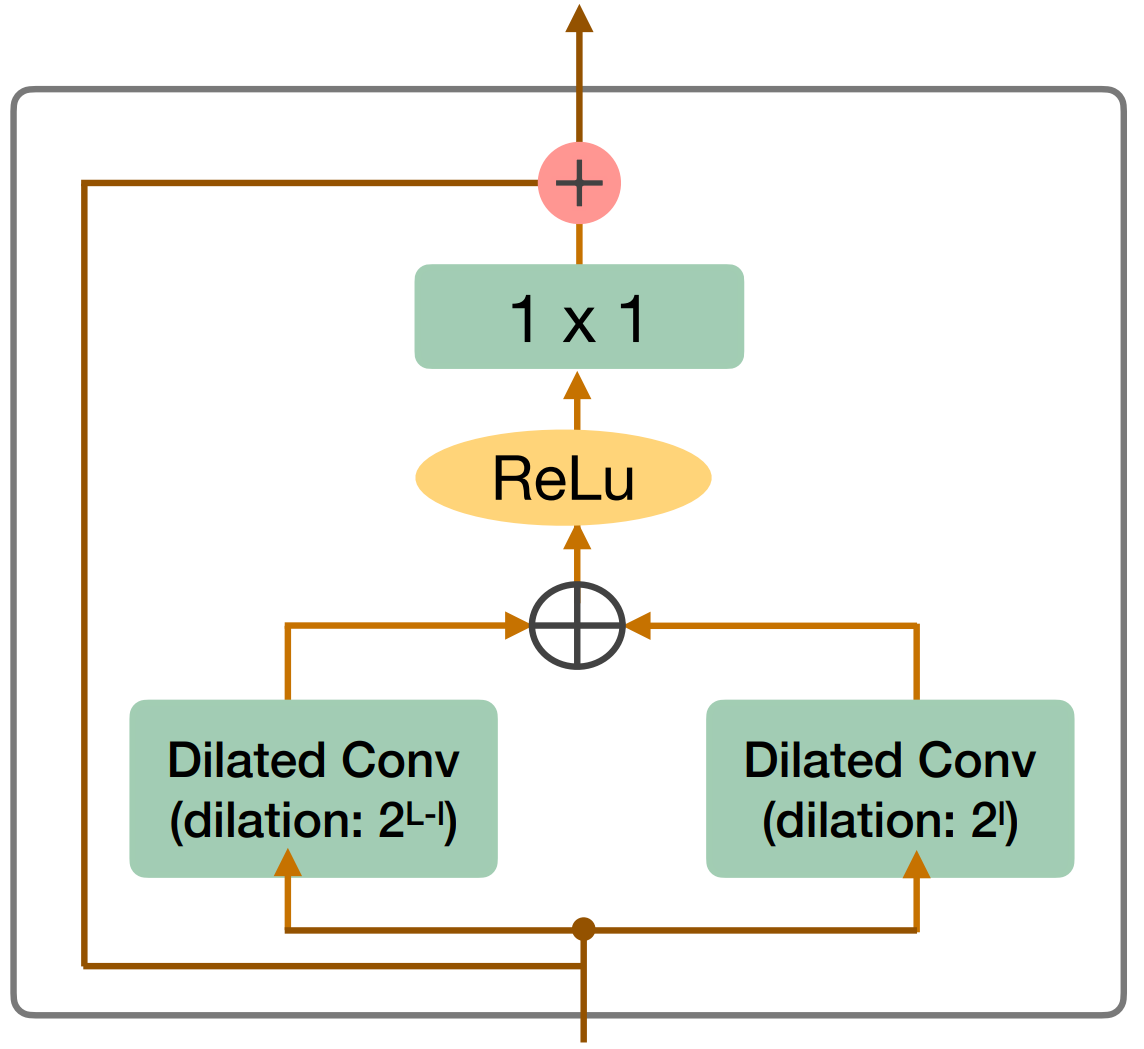}}
    \vspace{-3pt}
    \caption{Dilated Residual Layers of MS-TCN (a) and MS-TCN++ (b).}
    \label{fig:dilated_residuals}
    \vspace{-10pt}
\end{figure*}

The first layer of the first-stage TCN model of \cite{farha2019ms} is a $1 \times 1$ convolution layer which is then followed by several dilated 1-dimensional convolutional layers. Inspired by \cite{oord2016wavenet}, the dilation factor is doubled at each layer (\textit{i.e.} 1, 2, 4,$\dots$, 512), while using a constant number of convolutional filters at each layer. In order to support gradient flow, residual connections are applied. The MS-TCN model is based only on TCNs and no pooling or fully-connected layers are used. \cite{farha2019ms} reports that the pooling layers reduce the temporal resolution, while the addition of fully-connected layers greatly increases the number of trainable parameters and limits the model to operating on fixed-size inputs. Through this fully-convolutional approach, the model is efficient during both training and testing phases. 

In the subsequent stages of the network, Farha et al. \cite{farha2019ms} take the predictions from the previous stage and refine them through subsequent TCN models which have an architecture similar to that outlined above. This allows the network to capture dependencies between different action classes and action segments, and refines predictions when appropriate, helping to reduce segmentation errors. 

The authors also proposed to augment the model learning process with a combination of categorical cross entropy loss and a custom smoothing loss which reduces over-segmentation. Specifically, this loss is based on truncated mean squared error over the log probabilities of frame-wise predictions, and penalises the model for having predictions oscillate across consecutive frames. The authors have tested the proposed algorithm on the 50 Salads, GTEA and Breakfast~\cite{Kuehne12} datasets where it has shown significant performance gain compared to baselines methods. For instance, this model achieves 80.7\% frame wise classification accuracy. 67.9\% Edit score, and 64.5 \% F1@50 on the 50 Salads dataset, which is an approximately 15\% gain in the fame-wise classification accuracy over the ED-TCN model of \cite{lea2017temporal}.

\subsubsection{Multi-Stage Temporal Convolutional Network - Extended (MS-TCN++) \cite{li2020ms}}
\label{sec:mstcnplus_info}

This model extends the MS-TCN model of \cite{farha2019ms}. In particular, the authors show that the dilation factor and it's use in deep networks may lead to information being lost, as some input samples may receive less consideration due to the skipping of inputs through the dilated convolutions. While the deeper layers have larger temporal receptive fields, the shallow layers have very small receptive fields which could restrict them from learning informative temporal relationships. 
To address this, the authors propose expanding the network to use dual dilated convolution layers. Fig. \ref{fig:dilated_residuals} provides an overview of this dual dilated convolution layer. In the first convolution the dilation factor is exponentially increased to, for instance, $2^l$ where $l$ is the layer number. In the 2nd dilation layer, the lower layers start with a higher dilation factor, $2^{L-l}$ where $L$ is the total number of layers in the network, and it is exponentially decreased with the network depth. 

The authors show that the dual dilation strategy in MS-TCN++ has enabled them to capture both local and global features in all layers, irrespective of the network depth. Therefore, the predictions in all stages of the multi-stage architecture are more accurate and fewer refinements are needed. As such, they were able to use a relatively smaller number of stages compared to the original MS-TCN formulation, making it comparatively more efficient. 

In their empirical evaluations the authors have utilised 50 Salads, GTEA and their Breakfast datasets, where approximately a 3\% increase in frame-wise classification accuracy is observed when comparing MS-TCN++ to the MS-TCN architecture on the 50 Salads dataset. Furthermore, MS-TCN++ is capable of outperforming MS-TCN with an approximately 6\% gain in both edit distance and F1@50 score values. 

\subsubsection{Boundary-Aware Cascade Networks \cite{wang2020boundary}} 
Another extension to the TCN based human action segmentation pipeline is presented in \cite{wang2020boundary} where the authors tackle the problem of inaccurate action segmentation boundaries and misclassified short action segments. The authors show that this misclassification occurs due to deficiencies in the temporal modelling strategies employed by existing state-of-the-art models. Specifically, they identified that ambiguities in action boundaries and ambiguities in frames within long action segments cause these problems, and propose a cascade learning strategy to address this. They learn a series of cascade modules which progressively learn weak-to-strong frame level classifiers. In the earlier stages, the classifier has weak capacity and learns to recognise actions from less ambiguous and more informative frames. In later stages, more capacity is added to the classifiers and they pay more attention to ambiguous frames and try to make the predictions for those frames more accurate. 

To further augment the learning process, the authors propose a temporal smoothing loss function which further refines the predictions of action boundaries. Unlike the MS-TCN framework where the loss function is hand-engineered using prior knowledge, the authors propose to exploit information from the action boundaries and use it to maintain the semantic consistency of the frames within the predicted boundary. The proposed Local Barrier Pooling (LBP) module uses a binary classifier which predicts the boundaries of the current segment. A set of video specific weights are used to aggregate frame level predictions within identified boundaries. Hence, the prediction smoothing weights used are video and boundary specific. 

This system is evaluated using 50 Salads, GTEA and the Breakfast datasets. As shown in Tab.~\ref{table:SOTA-models}, on 50 Salads, the model achieves 84.4\% frame-wise classification accuracy, 74.3 \% edit score and 74.0\% F1@50 score, achieving a notable improvement in the temporal segmentation metrics over MS-TCN. 

\subsubsection{Coupled Action GAN ~\cite{gammulle2019coupled}}

A GAN based human action recognition framework is proposed in ~\cite{gammulle2019coupled}, where the generator network of the GAN is trained to generate what is termed an ``action code'', a unique representation of the action. The discriminator is trained to discriminate between the ground truth action codes and the synthesised action codes. Through this adversarial learning process, the generator learns an embedding space which segregates different action classes. 

To aid the temporal learning in this architecture, a context extractor is proposed which receives the previously generated action codes and the features from the current RGB frame. Using this information the context extractor generates an embedding to represent the evolution of the current action. This is leveraged by the generator in the action code synthesis process. To further augment this architecture an auxiliary branch is proposed, which receives auxiliary information such as depth or optical flow data. The generator network in the auxiliary branch also synthesises the action code using the auxiliary input and both generator networks try to synthesise action codes which are indistinguishable to real codes by the discriminator. In the coupled GAN setting, the context extractor also receives the past action codes and auxiliary features from the auxiliary branch.

This framework is tested using the 50 Salads, MERL Shopping and the GTEA datasets. It achieves 74.5\% 76.9\% and 71.1\% for frame-wise classification accuracy, Edit score and F1@50, respectively on the 50 Salads dataset. Tab. \ref{table:SOTA-models} further compares the results against the other state-of-the-art methods.  

\subsubsection{Semi-Supervised Action GAN ~\cite{gammulle2020fine}} 

An extension to this architecture is proposed in  \cite{gammulle2020fine}, where the discriminator was  extended to jointly classify the action class of the input frame together with the real/fake classification. This allows the framework to seamlessly utilise both synthesised unlabelled data and real labelled data, rendering a semi-supervised learning solution. 

To further aid the learning the context extractor is extended to a Gated Context Extractor in \cite{gammulle2019coupled}. This maintains a fixed length context queue where we sequentially push features of the observed frames to the queue. Utilising a series of gated operations, which individually evaluate the informativeness of each embedding in the context queue relative to the current frame, we dynamically determine how to combine context information together with the current observation. From this augmented feature vector, the generator network generates the action code. 

Similar to \cite{gammulle2019coupled} we conducted evaluations on 50 Salads, MERL Shopping and the GTEA datasets, and Semi-Supervised Action GAN achieves 73.3\% 69.8\% and 67.0\% for frame-wise classification accuracy, Edit score and F1@50, respectively on the 50 Salads dataset. The comparatively lower performance compared to our Coupled Action GAN architecture (see Tab. \ref{table:SOTA-models}) arises from the the lack of auxiliary feature stream, which offers additional valuable information. 

\subsection{Action Segmentation Augmentation Methods}
\label{sec:selected_models_2}

In this section we briefly discuss methods that augment those presented in Section \ref{sec:selected_models}. The models discussed here either build upon and refine the results produced by an earlier segmentation approach, or seek to optimise the segmentation architecture.


\subsubsection{Self-Supervised Temporal Domain Adaptation (SSTDA) \cite{chen2020action}}
\label{sec:sstda_info}

SSTDA \cite{chen2020action} is also an extension of the MS-TCN model, however, it tackles the domain discrepancy challenge within the action segmentation task and proposes an augmentation to the MS-TCN model to address this challenge. Specifically, the authors of \cite{chen2020action} consider the fact that there are significant spatio-temporal variations among the ways that humans perform the same action due to personalised preferences, habits and provided instructions. 

To overcome this issue the authors utilise two domain predictors based on two self-supervised auxiliary tasks, binary domain prediction (BDP) and sequential domain prediction (SDP). The BDP predicts the domain of each frame while the SDP predicts the domain of each video segment. Through these auxiliary tasks the framework can learn the similarities and differences between the way that a certain action is performed by different participants, allowing a more comprehensive modelling and understanding of context. The authors combine the auxiliary domain prediction task losses together with the frame-wise classification loss which is generated using the MS-TCN model, and train the complete framework end-to-end. 

Similar to both MS-TCN and MS-TCN++ the evaluations were conducted on 50 Salads, GTEA and the Breakfast dataset. In the 50 Salads dataset, a 1.1\% frame-wise action classification accuracy increase is observed compared to MS-TCN. Furthermore, a substantial 6.9\% and 8.6\% increase is observed for Edit and F1@50 score values, respectively. Note that Edit and F1@50 measure the temporal segmentation accuracy and the critical performance gain of SSTDA in these metrics clearly illustrate the superior temporal learning capabilities of SSTDA, where it has been able to overcome the small-scale differences between different subjects performing the same action.

\subsubsection{Graph-based Temporal Reasoning Module (GTRM)~\cite{huang2020improving}}
\label{sec:GTRM_info}
The GTRM proposed in ~\cite{huang2020improving} is composed of two GCNs, where each node is a representation of an action segment. This module is proposed to build on top of a backbone model, which itself is an existing action segmentation model. When given an output prediction from the chosen backbone, each segment is mapped to a graph node (where a node represents an action segment of an arbitrary length), and then passed through the two graph models to refine the classification and temporal boundaries of the nodes. To encourage the temporal relation mapping, the backbone and the proposed GTRM model are jointly optimised.

The model effectiveness is evaluated on two public datasets, Extended GTEA (EGTEA) \cite{li2018eye} and EPIC-Kitchens \cite{Damen2021RESCALING}. Through the experimental results, the authors have shown that by utilising GTRM on top of backbone models such as Bi-LSTM \cite{singh2016multi}, ED-TCN \cite{lea2017temporal} and MS-TCN, the backbone's initial results can be improved. For example, the original MS-TCN model achieves an edit score of $25.3\%$ on Epic-Kitchens dataset while it improves by $7.2\%$ when the MS-TCN is integrated with the GTRM module and achieves an edit score of $32.5\%$.    

\subsubsection{Dilated Temporal Graph Reasoning Module (DTGRM)~\cite{wang2021temporal}}
\label{sec:DTGRM_info}

The DTGRM is an action segmentation module that is designed based on relational reasoning and GCNs. The authors proposed to overcome the difficulties of GCNs when applied to long video sequences, as a large number of nodes (video frames) makes it hard for a GCN to effectively map temporal relations within the video. The proposed approach models temporal relations and their dependencies between different time spans, and is composed of two main components: multi-level dilated temporal graphs that map the temporal relations; and an auxiliary self-supervised task that encourages the dilated graph reasoning module perform the temporal relational reasoning task while reducing model over-fitting.

The authors reuse the dilated TCN in the MS-TCN framework (Sec. \ref{sec:mstcn_info}) as the backbone model. Similar to the original MS-TCN, the backbone model is fed pre-trained I3D features and the action class likelihood predictions are obtained through a final softmax function. These predictions are then sent through the proposed DTGRM framework. Inspired by the multi-stage refinement in the MS-TCN architecture, the authors of DTGRM also iteratively refine the predictions through DTGRM $S$ times before obtaining the final prediction results.

The proposed model is evaluated on 50 Salads, GTEA and Breakfast datasets. On 50 Salads, this model achieved an 80\% frame-wise classification accuracy, 72\% edit score and 66.1\% F1@.50 score. Compared to the MS-TCN model, the DTGRM achieved better segmentation performance with 4.1\% and 1.6\% improvements for edit score and F1@.50 respectively.  

\begin{figure*}[!t]
    \centering
    \includegraphics[width=0.7\linewidth]{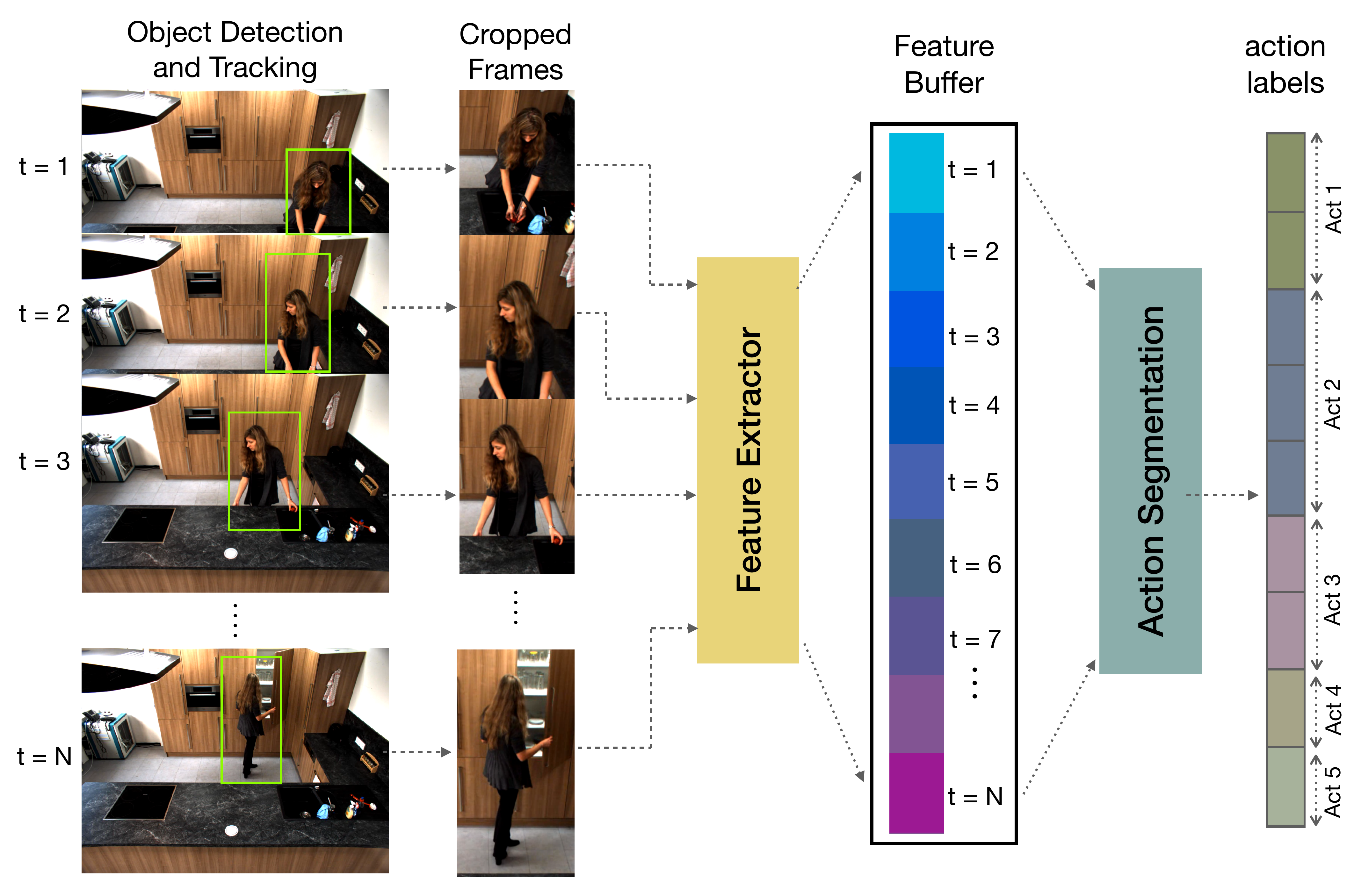}
	\caption{Action Segmentation with Object Detection.}
	\label{fig:framework_cropped}
    \vspace{-8pt}	
\end{figure*}

\subsubsection{Hierarchical Action Segmentation Refiner (HASR) \cite{ahn2021refining}}

A model agnostic add-on to refine action segmentation predictions is presented in \cite{ahn2021refining}, which encodes the context of the observed video in a hierarchical manner and uses context to refine fine-grained predictions. 

Specifically, in this architecture frame-level predictions from the backbone action segmentation model are encoded as a vector which is subsequently used as a query to generate attention weights for the corresponding frame-level features. Then, a video-level representation is generated by an encoder which consists of 2 residual blocks interleaved by a 1D max-pooling operation. Instead of directly passing the generated segment level embeddings to this encoder module, the authors sample multiple sub-sequences from the segment level embeddings and pass them multiple times. The authors demonstrate that this mechanism can compensate for erroneous predictions generated by the backbone action segmentation model when modelling the overall context of the video. The refiner network is a GRU network which accepts the video context and the initial predictions from the action segmentation model and corrects the erroneous predictions. 

The authors have tested the proposed refiner on the 50 Salads, GTEA and Breakfast datasets, and it shows a significant improvement in accuracy over the backbone network. For instance, in 50 Salads dataset for the MS-TCN model, for both edit and F1@50 metrics an approximate 7\% performance gain is observed. This gain is approximately 2\% and 3\% for edit and F1@50 scores for the SSTDA architecture. 

\subsubsection{Global2Local: efficient structure search for video action segmentation \cite{gao2021global2local}}

Different from the above research, Gao et al.~\cite{gao2021global2local} propose a mechanism to locate optimal receptive field combinations for an action segmentation model through a coarse-to-fine search mechanism which is termed a global-to-local search. 

In particular, the authors illustrate the deficiencies with existing state-of-the-art architectures where the receptive field parameters such as the dilation rate and pooling size of each layer are hand defined by evaluating a small number of combinations. However, these parameters are crucial factors which determine the capacity of the network to recover short-term and long-term temporal dependencies.

To address this limitation, the authors propose a search algorithm which automatically identifies the best parameters to configure a given network. However, exhaustively evaluating all possible parameters is computationally infeasible. For instance, possible combinations are in the range of $1024^{40}$ for the MS-TCN architecture. A solution is proposed via a global-to-local search algorithm which narrows the search space using a low-cost genetic algorithm based global search operation, and a more exhaustive fine-grained search is subsequently conducted in the narrowed search space using an expectation guided algorithm. 

This receptive field refinement strategy has been evaluated on the 50 Salads and Breakfast datasets (see Tab. \ref{table:SOTA-models}). One of the noteworthy achievements of this architecture is the significant ~5\% increase in both Edit and F1@50 segmental scores in the 50 salads dataset compared to the MSTCN model, only by refining the receptive field sizes, without modifying the overall network structure. 

\section{Object Detection for Action Segmentation}
\label{sec:dettrack_actionrec}

Features relating to detected humans can be incorporated into the action segmentation workflow to guide the localisation of salient regions (\textit{i.e.} regions that contain humans) in the video stream, and promote the selection of discriminative features from regions-of-interest (ROIs). As common applications of human action segmentation (including target videos within this study) include human-object interaction, it is often of value to track the subject through time as well.

In a real-world setting, there is typically a large amount of background clutter which contains objects which are unrelated to the action being performed. Furthermore, in some situations multiple people may be observed, yet actions are performed by a single subject. In such situations, object detection and tracking can help identify the person of interest. This pipeline is illustrated in Fig.~\ref{fig:framework_cropped}. 
By incorporating this pre-processing phase, the person of interest is identified throughout the video which allows the action segmentation model to focus on information that is relevant to the action performed. In particular, the detected person of interest is cropped and features are extracted from a cropped region around the subject rather than the full-frame, as shown in Fig.~\ref{fig:gen_framework}


\subsection{Object Detection}
\label{sec:object_det}

Object detection, an important yet challenging task in computer vision, aims to discover object instances in an image given a set of predefined object categories. Detecting objects is a difficult problem that requires the solution to two main tasks. First, the detector must handle the recognition problem, distinguishing between foreground and background objects, and assigning them the correct object class labels. Second, the detector must solve the localisation problem, assigning precise bounding boxes to objects. Object detectors have achieved exceptional performance in recent years, thanks to advances in deep convolutional neural networks. 

Object detectors can be categorised as \textit{anchor-based} or \textit{anchor-free} methods \cite{jiao2019survey}. The core idea of anchor-based models is to introduce a constant set of bounding boxes, referred to as \textit{anchors}, which can be viewed as a set of pre-defined proposals for bounding box regression. Such anchors are defined by the user before the model is trained. Models typically refine these anchors to produce the final set of bounding boxes that contain the detected objects. Nevertheless, using anchors requires several hyperparameters (\textit{e.g.} the number of boxes, sizes and the aspect ratios). Even slight changes in these hyperparameters impact the end-result, thus selecting an optimal set of anchors is, to an extent, dependent on the experience and skill of the researcher.

Overcoming the limitations imposed by hand-crafted anchors, anchor-free methods offer significant promise to cope with extreme variations in object scales and aspect ratios~\cite{ke2020multiple}. Such approaches, for example, can perform object bounding box regression based on anchor points instead of boxes (\textit{i.e.} the object detection is reformulated as a keypoint localisation problem).

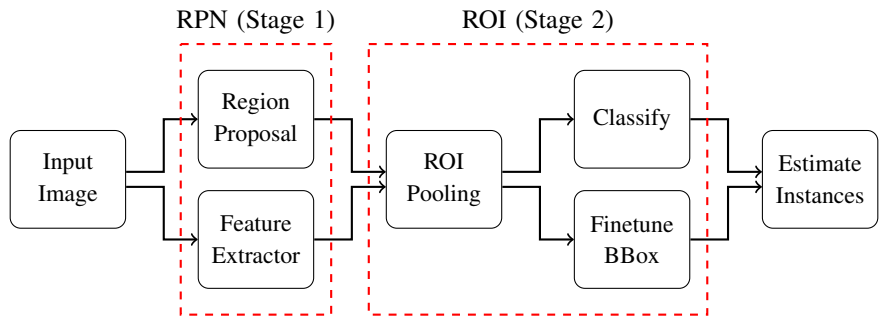
\begin{figure*}[tb]
\centering
\begin{tikzpicture}[yscale=2,xscale=2.5]
\tikzstyle{block} = [rectangle, rounded corners, text centered, draw=black, fill=white, minimum size=1.3cm, text width=1.3cm,  anchor=center]
\node (image) [block] at (0,0) {\small Input Image};
\node (rpn) [block, fill=white] at (1,0.4) {\small Region Proposal };
\node (dss) [block, fill=white] at (2,0) {\small ROI Pooling};
\node (cls) [block, fill=white] at (3,0.4) {\small Classify};
\node (bbox) [block, fill=white] at (3,-0.4) {\small Finetune BBox};
\node (feat) [block,  fill=white] at (1,-0.4) {\small Feature Extractor};
\node (inst) [block,  fill=white] at (4,0.0) {\small Estimate Instances};
\draw [->,thick] ($(image.east)+(0.0,0.05)$) -- ++(0.2,0) |- (rpn.west);
\draw [->,thick] ($(image.east)-(0.0,0.05)$) -- ++(0.2,0) |- (feat.west);
\draw [->,thick] (rpn.east) -- ++(0.2,0) |- ($(dss.west) + (0,0.05)$);
\draw [->,thick] (feat.east) -- ++(0.2,0) |- ($(dss.west)- (0,0.05)$);
\draw [->,thick] ($(dss.east)+(0,0.05)$) -- ++(0.2,0) |- (cls.west);
\draw [->,thick] ($(dss.east)-(0,0.05)$) -- ++(0.2,0) |- (bbox.west);
\draw [->,thick] (cls.east) -- ++(0.2,0) |- ($(inst.west)+(0,0.05)$);
\draw [->,thick] (bbox.east) -- ++(0.2,0) |- ($(inst.west)-(0,0.05)$);
\draw [dashed,thick,red] (0.6,0.9) -- node[above,black,pos=.5, align=center, text width=3cm] {RPN (Stage 1) } ++ (0.8,0) -- (1.4,-0.9) -- (0.6,-0.9) -- (0.6,0.9);
\draw [dashed,thick,red] (1.6,0.9) -- node[above,black,pos=.5, align=center, text width=3cm] {ROI (Stage 2)} ++ (1.8,0) -- (3.4,-0.9) -- (1.6,-0.9) -- (1.6,0.9);
\end{tikzpicture}
\caption{Typical high level algorithm applied by a two-stage detector. Most modern methods implement stage 1 and 2 using neural networks, and estimate instances via a non-max suppression (NMS) clustering algorithm.}
\label{fig:overview_twostage}
\vspace{-8pt}
\end{figure*}

The design of object detectors can also be broadly divided into two types: two-stage and one-stage detectors \cite{jiao2019survey}. 
\begin{itemize}
    \item \textit{Two-stage detection frameworks} use a region proposal network (RPN) to identify regions of interest (ROIs). A second network is then applied to the ROIs to identify the object class and to regress to an improved bounding box. %
    Two-stage detectors are often more flexible than their one-stage counterparts, since other \textit{per instance} operations can be easily added to the second network to enhance their capability \textit{(e.g.} instance segmentation and instance tracking). Fig. \ref{fig:overview_twostage} provides a high level algorithmic overview of the two-stage method. 

    \item \textit{One-stage detection frameworks} use a single network to perform classification and bounding box regression. 
    Single-stage detectors are often faster to evaluate, however most designs cannot match two-stage detectors for bounding box localisation accuracy.
\end{itemize}

\begin{figure*}[!t]
\begin{center}
\includegraphics[width=0.9\linewidth]{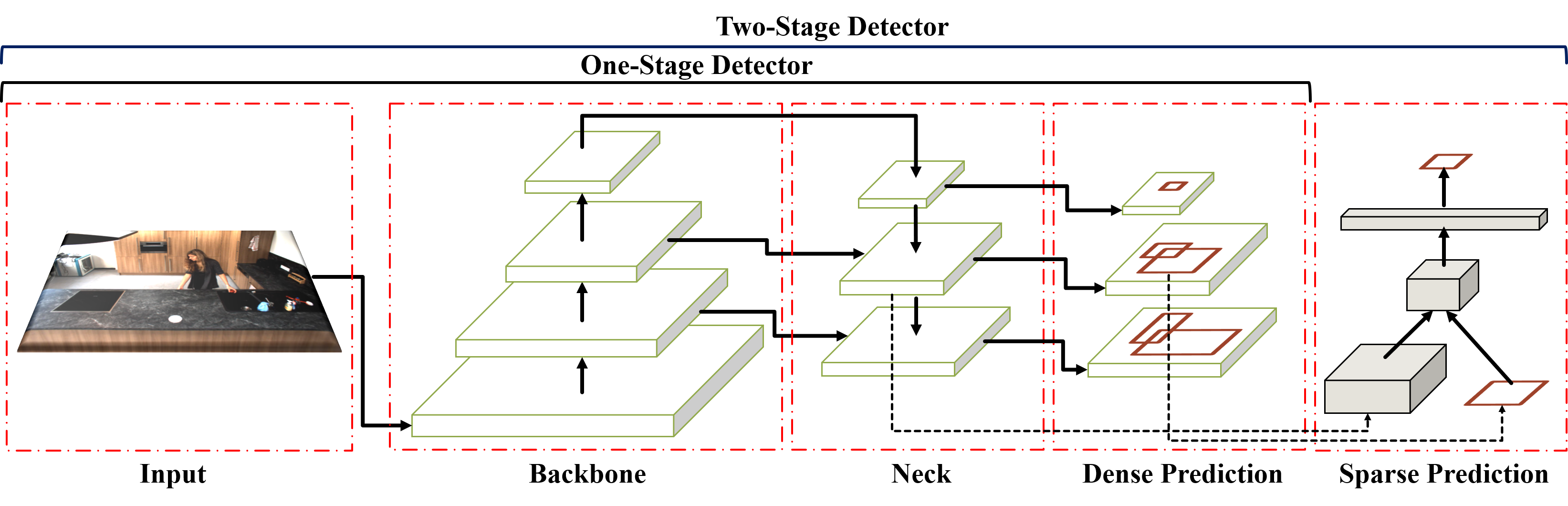}
\end{center}
\vspace{-9pt}
   \caption{Structure of object detectors, highlighting the main components of the Backbone, Neck and Head, and how information flows between these components. Image recreated from~\cite{bochkovskiy2020yolov4}.
   }
\vspace{-10pt}
\label{fig:overview_objectdetectors}
\end{figure*}

Variants of models are based on changes in the main components of the standard architecture of an object detector. The structure for two-stage and one-stage object detectors is shown in Fig.~\ref{fig:overview_objectdetectors}. Common components that may be changed are detailed as follows:

\begin{itemize}
    \item \textbf{Backbone:}
    Backbones are used as feature extractors. As discussed above in Sec. ~\ref{sec:feature_extraction}, they are mainly commonly feed-forward CNNs or residual networks. These networks are pre-trained on image classification datasets (\textit{e.g.} ImageNet~\cite{krizhevsky2012imagenet}), and then fined-tuned on the detection dataset.
    In addition to the networks already introduced, other popular backbones include 
    Inception-v3~\cite{szegedy2016rethinking} and Inception-v4~\cite{szegedy2017inception},
    ResNext~\cite{xie2017aggregated} and ResNet-vd~\cite{he2019bag}, SqueezeNet~\cite{iandola2016squeezenet}, ShuffleNet~\cite{zhang2018shufflenet}, Darknet-53~\cite{redmon2018yolov3}, and CSPNet~\cite{wang2019cspnet}.
    \item \textbf{Neck:} 
    These are extra layers that sit between the backbone and the head, and are used to extract neighbouring feature maps from different stages of the backbone. Such features are summed element-wise, or concatenated prior to being fed to the head.
    Usually, a neck consists of several bottom-up and top-down paths, such that enriched information is fed to the head. In this scheme, the top-down network propagates high-level large scale semantic information down to shallow network layers, while the bottom-up network encodes the smaller scale visual details via deep network layers. Therefore, the head’s input will contain spatially rich information from the bottom-up path, and semantically rich information from the top-down path.    
    The neck can, for example, be a feature pyramid network (FPN)~\cite{lin2017feature}, a bi-directional FPN (BiFPN)~\cite{tan2020efficientdet}, a spatial pyramid pooling (SPP)~\cite{he2015spatial} network, and a path aggregation network (PANet)~\cite{liu2018path}.
    \item \textbf{Head:}
    This is the network component in charge of the detection (classification and regression) of bounding boxes. 
    The head can be a dense prediction (one-stage) or a sparse prediction (two-stage) network. Object detectors that are anchor-based apply the head network to each anchor box.

\end{itemize}

Given the basic object detector design, there are many methods to improve classification accuracy, bounding box estimation and evaluation speed. For the purpose of this review, we categorise these improvements as either bag of freebies (BoF) or bag of specials (BoS). These methods are:
\begin{itemize}
    \item \textbf{Bag of freebies (BoF):}
    Methods that can improve object detector accuracy without increasing the inference cost. These methods only change the training strategy, or only increase the training cost.
    Examples are data augmentation and regularisation techniques used to avoid over-fitting such as DropOut, DropConnect and DropBlock.
    \item \textbf{Bag of specials (BoS):} 
    Modules and post-processing methods that only increase the inference cost by a small amount, but can significantly improve the accuracy of object detection. 
    These modules/methods usually introduce attention mechanisms including Squeeze-and-Excitation and Spatial Attention Modules, to enlarge the receptive field of the model and enhance feature integration capabilities.
\end{itemize}

We refer the readers to Appendix B-A of the supplementary material for a thorough review of the most relevant object detector architectures from the literature. We organise the object detectors as follows:

\begin{enumerate}
    \item Anchor-based (Two-stage Frameworks): 
    \textit{Faster-RCNN}~\cite{ren2015faster}, \textit{R-FCN})~\cite{dai2016r}, \textit{Libra R-CNN}~\cite{pang2019libra}, \textit{Mask R-CNN}~\cite{he2017mask}, \textit{Chained cascade network and Cascade R-CNN}~\cite{ouyang2017chained,cai18cascadercnn}, and \textit{TridentNet}~\cite{li2019scale}.

    \item Anchor-based (One-stage Frameworks): 
    \textit{Single shot multibox detector (SSD)~\cite{liu2016ssd}}, \textit{(DSSD)}~\cite{fu2017dssd}, \textit{RetinaNet}~\cite{lin2017focal},  \textit{M2det}~\cite{zhao2019m2det}, \textit{EfficientDet}~\cite{tan2020efficientdet}, and YOLO family detectors (\textit{YOLOv3}~\cite{redmon2018yolov3}, \textit{YOLOv4}~\cite{bochkovskiy2020yolov4}, and \textit{Scaled-YOLOv4}~\cite{wang2021scaled}).
    
    \item Anchor-free Frameworks: 
    \textit{DeNet}~\cite{tychsen2017denet}, \textit{CornerNet}~\cite{law2018cornernet}, \textit{CornetNet-lite}~\cite{law2019cornernet}, \textit{CenterNet (objects as points)}~\cite{zhou2019objects}, \textit{CenterNet (keypoint triplets)}~\cite{duan2019centernet},    \textit{FCOS}~\cite{tian2019fcos}, and \textit{YOLOX}~\cite{ge2021yolox}.
\end{enumerate}

Among well-known single stage object detectors, the \textit{You only look once (YOLO)} family of detectors (\textit{YOLO}~\cite{redmon2016you}, \textit{YOLOv2}~\cite{redmon2017yolo9000}, \textit{YOLOv3}~\cite{redmon2018yolov3}, \textit{YOLOv3-tiny}, \textit{YOLOv3-SPP}) have demonstrated impressive speed and accuracy. This detector can run well on low powered hardware, thanks to the intelligent and conservative model design. 
In particular, their recent variants \textit{YOLOv4}~\cite{bochkovskiy2020yolov4} and \textit{Scaled-YOLOv4}~\cite{wang2021scaled} achieve one of the best trade-offs between speed and accuracy.

\textit{YOLOv4}~\cite{bochkovskiy2020yolov4} is composed of CSPDarknet53 as a backbone, a SPP additional module, a PANet as the neck, and a YOLOv3 as the head. 
CSPDarknet53 is a novel backbone that can enhance the learning capability of the CNN by integrating feature maps from the beginning and the end of a network stage. The BoF for YOLOv4 backbones includes CutMix and Mosaic data augmentation~\cite{yun2019cutmix}, DropBlock regularisation~\cite{ghiasi2018dropblock}, and class label smoothing~\cite{szegedy2016rethinking}. The BoS for the same CSPDarknet53 are Mish activation~\cite{misra2019mish}, cross-stage-partial-connections (CSP)~\cite{wang2019cspnet}, and multi-input weighted residual connections (MiWRC).
Finally, the scaling cross stage partial network \textit{(Scaled-YOLOv4)}~\cite{wang2021scaled} achieves one of the best trade-offs between speed and accuracy. In this approach, YOLOv4 is redesigned to form YOLOv4-CSP with a network scaling approach that modifies not only the network depth, width, and resolution; but also the structure of the network. Thus, the backbone is optimised and the neck (PANet) uses CSP and Mish activations.

\subsection{Multi-object Tracking}
\label{sec:multi-track}

Given the location of an arbitrary target of interest in the first frame of a video, the aim of visual object tracking is to estimate its position in all the subsequent frames. The ability to perform reliable and effective object tracking depends on how a tracker can deal with challenges such as occlusion, scale variations, low resolution targets, fast motion and the presence of noise. Visual object tracking algorithms can be categorised into single-object and multiple-object trackers (MOT), where the latter is the scope of this manuscript.

Due to recent progress in object detection, \textit{tracking-by-detection} has become the leading paradigm for multiple object tracking. This method is composed of two discrete components: object detection and data association. Detection aims to locate potential targets-of-interest from video frames, which are then used to guide the tracking process. The \textit{association} component uses geometric and visual information to allocate these detections to new or existing object trajectories (also known as \textit{tracklets}), \textit{e.g}, the re-identification (ReID) task~\cite{zheng2015scalable}. We refer to a \textit{tracklet} as a set of linked regions defined over consecutive frames~\cite{peng2020tpm}.

The core algorithm in multi-object tracking is this data association method. In many implementations, this method computes the similarity between detections and existing tracklets, identifies which detections should be matched with existing tracklets, and finally creates new tracklets where appropriate. 

The similarity function computes a score between two object instances, $I_0$ and $I_1$, observed at differing times, and indicates the likelihood that they are the same object instance. The similarity function is typically implemented via a combination of geometric methods (\textit{e.g.} motion prediction or bounding box overlap) and visual appearance methods (embeddings).

%
Despite the wide range of methodologies described in the literature, the great majority of MOT algorithms include some or all of the following steps:
1) Detection stage; 2) Feature extraction/motion prediction stage (appearance, motion or interaction features); 3) Affinity stage (similarity/distance score calculation); and 4) Association stage (associate detections and tracklets).
MOTs approaches can be divided with respect to their complexity into separate detection and embedding (SDE) methods, and joint detection and embedding (JDE) algorithms.
\begin{itemize}
    \item \textit{SDE methods} completely separate stages of detection and embedding extraction. Such a design allows the system to adapt to various detectors with fewer changes, and the two components can be tuned separately 
    (\textit{e.g.} \textit{Simple online and realtime tracking (SORT)}~\cite{Bewley2016_sort}, \textit{DeepSORT}~\cite{Wojke2017simple}, \textit{ByteTrack}~\cite{zhang2021bytetrack}).

    \item \textit{JDE methods} learn to detect objects and extract embeddings at the same time via a shared neural network, and use multi-task learning to train the network. This design takes into account both accuracy and speed, and can achieve high-precision real-time multi-target tracking
    (\textit{e.g.} \textit{Tracktor}~\cite{bergmann2019tracking}, \textit{CenterTrack}~\cite{zhou2020tracking}, \textit{FairMOT}~\cite{zhang2021fairmot},  \textit{SiamMOT}~\cite{shuai2021siammot}).
\end{itemize}

We refer the readers to Appendix B-B of the supplementary material, where we introduce these SDE and JDE frameworks in detail.
From these methods, \textit{ByteTrack}~\cite{zhang2021bytetrack} is a simple and effective tracking by association method for real-time applications, and makes the best use of detection results to enhance multi-object tracking.
ByteTrack keeps all detection boxes (detected by YOLOX) and makes associations across all boxes instead of only considering high scoring boxes, to reduce missed detections. In the matching process, an algorithm called BYTE first predicts the tracklets using a Kalman filter, which are then matched with high-scoring detected bounding boxes using motion similarity. Next, the algorithm performs a second matching between the detected bounding boxes with lower confidence values and the objects in the tracklets that could not be matched.

\section{Experiments}
\label{sec:experiments}

\subsection{Dataset}
Widely used datasets in the current literature include
Breakfast \cite{Kuehne12}, 50Salads \cite{stein2013combining},
MPII cooking activities dataset \cite{rohrbach2012database}, 
MPII cooking 2 dataset \cite{rohrbach15ijcv},
EPIC-KITCHENS-100 \cite{Damen2021RESCALING},
GTEA \cite{fathi2011learning},
ActivityNet \cite{Heilbron_2015_CVPR},
THUMOS15 \cite{idrees2017thumos},
Toyota Smart-home Untrimmed dataset \cite{Das_2019_ICCV}, and
FineGym \cite{shao2020finegym}. 
We refer the readers to Appendix C of the supplementary material for more details on these publicly available datasets.

For our experiments, we use the \textit{MPII Cooking 2 fine-grained action dataset} \cite{rohrbach15ijcv}, considering its popularity and challenging nature due to the unstructured manner in which the actions evolve over time.
Note that the egocentric view of the \textit{EPIC-KITCHENS-100} \cite{Damen2021RESCALING} and \textit{GTEA} \cite{fathi2011learning} datasets means that the human subject cannot be detected and segmented. As our evaluation explicitly considers the role that object detection and tracking can play in supporting the action segmentation task, we cannot utilise these datasets for our evaluation, despite their recent popularity. 
Furthermore, due to the lack of available public multi-person action segmentation datasets, we can not directly evaluate the state-of-the-art methods in a multi-person setting. However, we believe that our evaluations illustrate a general purpose pipeline for the readers to use in a real-world setting where there are multiple people concurrently conducting temporally evolving actions.

\begin{figure*}[!t]
\begin{center}
\includegraphics[width=0.8\linewidth]{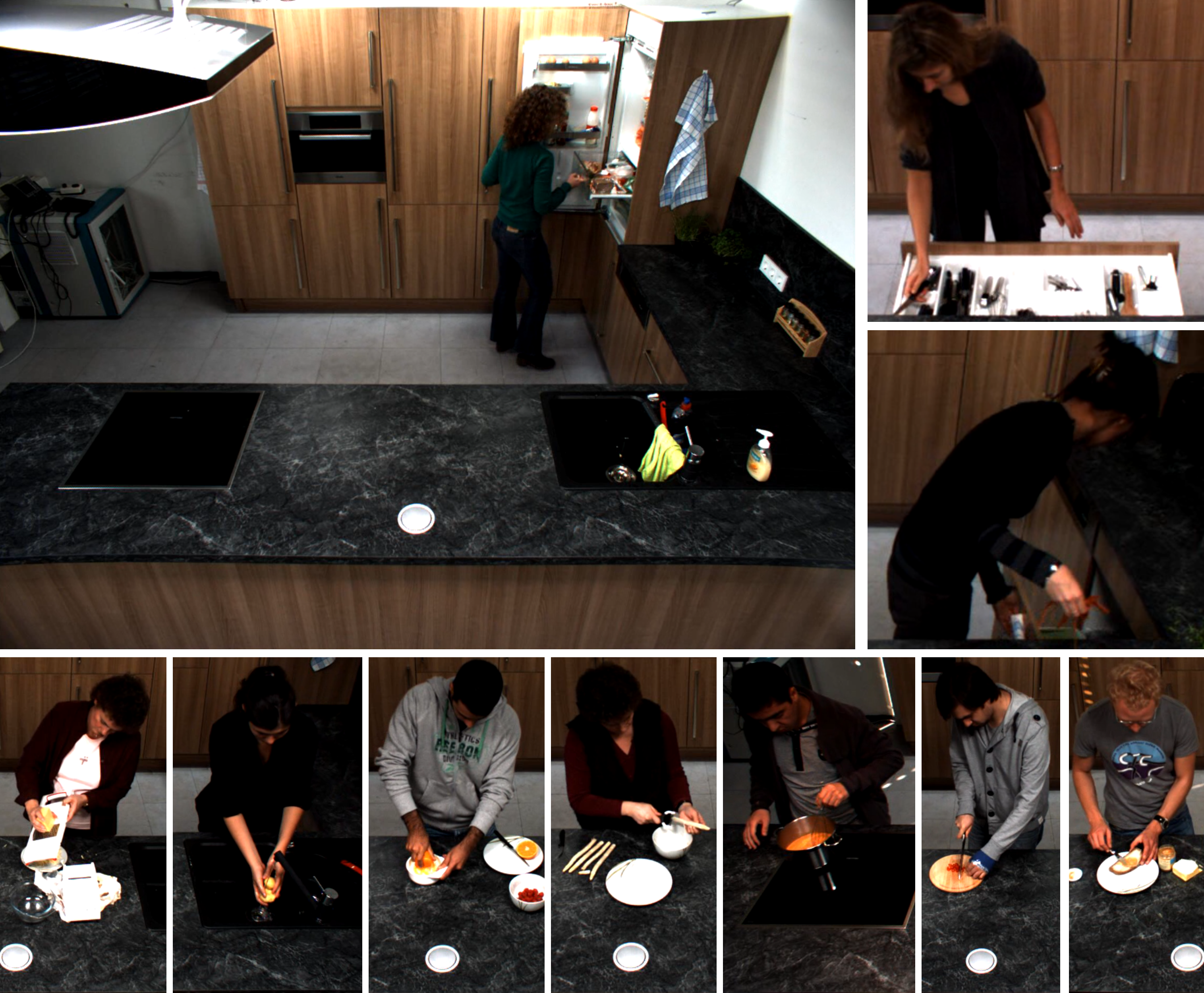}
\end{center}
\vspace{-9pt}
   \caption{Single frames from the MPII Cooking 2 fine-grained action \cite{rohrbach15ijcv} dataset that illustrate the full scene of the environment and several fine-grained cooking activities with different participants. Samples of those activities include take out, dicing, peel, cut, squeeze, spread, wash, and throw in garbage.
   }
\vspace{-4pt}
\label{fig:mpii-samples}
\end{figure*}

The MPII Cooking 2 dataset contains 273 videos with a total length of more than 27 hours, captured from 30 subjects preparing a certain dish, and comprising 67 fine-grained actions. Similar to the MPII cooking activities dataset, this dataset is captured from a  single camera at a 1624 $\times$ 1224 pixel resolution, with a frame rate of 30 fps. The camera is mounted on the ceiling and captures the front view of a person working at the counter. The duration of the videos range from 1 to 41 minutes. As per the MPII cooking activities dataset, this dataset offers different subject specific patterns and behaviours for the same dish, as no instructions regarding how to prepare a certain dish are provided to the participants. In addition to video data, the dataset authors provide human pose annotations, their trajectories, and text-based video script descriptions. 
Example frames of the dataset are shown in Fig.~\ref{fig:mpii-samples}. 

\subsection{Evaluation Metrics:}
Our evaluations utilise both frame-wise and segmentation metrics. As the frame-wise metric we utilise the frame-wise accuracy, which is a widely used metric for action recognition and segmentation tasks \cite{gammulle2017two,simonyan2014two,farha2019ms}. However, as stated in \cite{farha2019ms} such frame-wise metrics are unable to properly detect over-segmentation errors, as such errors have a low impact on the frame-wise accuracy. Furthermore, longer duration action classes have a higher impact on the frame-wise metric. Therefore, it is essential to utilise segmentation metrics to fully evaluate action segmentation/detection performance. In our experiments, we use the following segmentation metrics addition to frame-wise accuracy.

\begin{itemize}

    \item \textbf{Segmental edit score (edit)} is calculated through the normalised edit distance for each ground truth (G) and predicted (P) label using the Wagner-Fischer algorithm. Then this is calculated as follows,
    %
    \begin{equation}
        (1-S_{e}(G,P)) \times{100},
    \end{equation}
    
    where the best score is 100 while the worst is 0.
    
    \item \textbf{Segmental F1 score (F1@k)} is calculated by first computing whether or not each predicted action segment is a true positive (TP) or false positive (FP), by comparing its IoU with respect to the corresponding ground truth with threshold, $k$. Then, the segmental F1 score is obtained through precision (P) and recall (R) calculations as follows,
    \begin{equation}
        F1= 2 \frac {P\times{R}}{P+R},
    \end{equation}
    
    where,
    
    \begin{equation}
        P = \frac{TP}{TP+FP} ,\, R = \frac{TP}{TP+FN}.
    \end{equation}

\end{itemize}

\subsection{Implementation Details}
\label{sec:imple_details}

For the selected action segmentation models we follow settings outlined in the original works, except we use different feature extraction backbones. More information on the selection of the feature extraction models is provided in Sec. \ref{sec:init_eval}. More details on the feature extraction layers and the feature dimensions are provided in Table~\ref{tab:backbone_info}.    


\subsection{Initial Evaluations}
\label{sec:init_eval}

For our initial experiment, we evaluate the selected action segmentation and detection models on the MPII Cooking 2 dataset. We select the Multi-Stage Temporal Convolutional Network (MS-TCN) \cite{farha2019ms}, Multi-Stage Temporal Convolutional Network - Extended (MS-TCN++) \cite{li2020ms}, Self-Supervised Temporal Domain Adaptation (SSTDA) \cite{chen2020action} and Dilated Temporal Graph Reasoning Module (DTGRM)~\cite{wang2021temporal} models for evaluation. Models are selected considering both their performance and efficiency. In practical applications, such as in mobile-robotics, computing resources are limited and as such we have to consider both performance and efficiency when selecting the models. 

We use three different feature extractors ResNet-50, EfficientNet-B0 and MobileNet-v2. More details regarding these backbone models are provided in Sec. \ref{sec:feature_extraction}. Here, we select the ResNet-50 backbone model instead of the I3D model (see Sec. \ref{sec:feature_extraction}) that is widely used in action segmentation methods \cite{farha2019ms,chen2020action} as the I3D model requires additional calculations of the optical flow inputs. Keeping the ResNet-50 as our first backbone, we consider two light-weight models EfficientNet-B0 and MobileNet-v2 as these are of more relevance given the final goal on adapting the models to a real-world application. We only make use of the action labels during training, and additional information such as object annotations that are available with the dataset are not utilised. 
In Table~\ref{tab:cooking_results} we report the experimental results when using the original full-frames as the model input. In addition, as described in Sec. \ref{sec:dettrack_actionrec}, selection of individual persons in the scene and recognising their individual actions has to be carried out when there are multiple people in the scene. For instance, consider a human-robot interaction scenario as our target application. In this scenario, the robot must identify the actions performed by the human in order to react appropriately. In a real-world environment where there are multiple humans, prior to action recognition the system must detect the human, generating a single bounding box for each person detected. Then, an object tracking technique is integrated to estimate human position in subsequent frames, and to maintain the identity of the subject. Therefore, in our initial evaluations presented in Table~\ref{tab:cooking_results} we also evaluate how the state-of-the-art models perform with image cropped using a bounding box. 

\begin{table}[!t]
\centering
\caption{Feature extraction backbones used in the evaluations.}
\resizebox{0.47\textwidth}{!}{%
\begin{tabular}{|c|c|c|c|}
\hline
Backbone        & Implementation   & Layer                             & Feat\_dim \\ \hline
ResNet50        & Keras-Tensorflow & avg\_pool (GlobalAveragePooling2) & 2048      \\ \hline
EfficientNet-B0 & PyTorch          & AdaptiveAvgPool2d-277             & 1280      \\ \hline
MobileNet-V2    & PyTorch          & AdaptiveAvgPool2d-157             & 1280      \\ \hline
\end{tabular}}
\label{tab:backbone_info}
\vspace{-6pt}
\end{table}

Among the object detectors, we consider single-stage methods as they are better suited to practical applications due to the faster inference. We seek to obtain an optimal balance between inference time and accuracy, noting that there is a trade-off between these, by evaluating different components such as the backbone, neck and head of the object detection architecture.
From the proposed models discussed earlier, we adopt the \textit{Scaled-YOLOv4} model available in~\cite{darknet-ab}, which achieves competitive accuracy while maintaining a high processing frame rate (inference speed greater than 30 FPS).
To alleviate the high computational demand and achieve real-time object tracking, we use \textit{ByteTrack}~\cite{ByteTrack}, which is a comparatively simple tracking method that can handle occlusion and view-point changes.
Both these approaches also allow use for commercial purposes.

\begin{table*}[]
\centering
\caption{Evaluation Results on the MPII Cooking 2 Dataset: The evaluations are carried out based on different feature extraction backbones (ResNet50, EfficientNet-B0 and MobileNet-V2). Performance using full-frame features (Full) and cropped frame features based on object detection bounding boxes (Bbox) are given.}
\begin{tabular}{|c|c|c|c|c|c|c|c|}
\hline
Method                   & Backbone                         & Full/Bbox & Acc            & Edit           & F1@0.10        & F1@0.25        & F1@0.50        \\ \hline
\multirow{6}{*}{MS-TCN}   & \multirow{2}{*}{ResNet50}        & Full      & 39.98          & 25.78          & 24.70          & 22.18          & 15.49          \\ \cline{3-8} 
                         &                                  & Bbox      & 40.48          & 24.71          & 22.34          & 20.14          & 13.72          \\ \cline{2-8} 
                         & \multirow{2}{*}{EfficientNet-B0} & Full      & 44.43          & 31.14          & 27.29          & 25.47          & 18.88          \\ \cline{3-8} 
                         &                                  & Bbox      & 40.37          & 30.06          & 25.86          & 24.38          & 16.78          \\ \cline{2-8} 
                         & \multirow{2}{*}{MobileNet-V2}    & Full      & 40.55          & 29.51          & 27.82          & 25.14          & 18.40          \\ \cline{3-8} 
                         &                                  & Bbox      & 37.13          & 24.06          & 19.77          & 18.11          & 11.28          \\ \hline
\multirow{6}{*}{MS-TCN++} & \multirow{2}{*}{ResNet50}        & Full      & 41.47          & 25.45          & 23.92          & 21.32          & 13.97          \\ \cline{3-8} 
                         &                                  & Bbox      & 39.77          & 27.90          & 23.80          & 21.16          & 14.61          \\ \cline{2-8} 
                         & \multirow{2}{*}{EfficientNet-B0} & Full      & 44.40          & 29.64          & 27.42          & 25.49          & 19.42          \\ \cline{3-8} 
                         &                                  & Bbox      & 41.32          & 28.60          & 25.11          & 23.25          & 17.59          \\ \cline{2-8} 
                         & \multirow{2}{*}{MobileNet-V2}    & Full      & 41.61          & 27.91          & 24.96          & 22.45          & 16.07          \\ \cline{3-8} 
                         &                                  & Bbox      & 37.71          & 23.25          & 19.56          & 17.14          & 10.56          \\ \hline
\multirow{6}{*}{SSTDA}   & \multirow{2}{*}{ResNet50}        & Full      & 32.77          & 17.80          & 14.37          & 12.66          & 10.65          \\ \cline{3-8} 
                         &                                  & Bbox      & 31.78          & 19.50          & 12.26          & 10.88          & 8.55           \\ \cline{2-8} 
                         & \multirow{2}{*}{EfficientNet-B0} & Full      & 36.09          & 25.71          & 14.98          & 13.90          & 10.99          \\ \cline{3-8} 
                         &                                  & Bbox      & 32.76          & 19.57          & 14.10          & 10.07          & 9.84           \\ \cline{2-8} 
                         & \multirow{2}{*}{MobileNet-V2}    & Full      & 42.50          & \textbf{32.18} & 26.81          & 24.52          & 17.62          \\ \cline{3-8} 
                         &                                  & Bbox      & 35.50          & 23.89          & 20.33          & 17.84          & 10.04          \\ \hline
\multirow{6}{*}{DTGRM}   & \multirow{2}{*}{ResNet50}        & Full      & 40.71          & 28.60          & 24.70          & 22.09          & 15.46          \\ \cline{3-8} 
                         &                                  & Bbox      & 40.17          & 25.96          & 24.05          & 21.88          & 15.22          \\ \cline{2-8} 
                         & \multirow{2}{*}{EfficientNet-B0} & Full      & \textbf{45.43} & 30.27          & \textbf{28.04} & \textbf{26.17} & \textbf{19.97} \\ \cline{3-8} 
                         &                                  & Bbox      & 42.84          & 27.09          & 25.49          & 23.18          & 17.73          \\ \cline{2-8} 
                         & \multirow{2}{*}{MobileNet-V2}    & Full      & 41.96          & 29.36          & 25.80          & 24.00          & 18.02          \\ \cline{3-8} 
                         &                                  & Bbox      & 37.57          & 24.58          & 20.42          & 18.97          & 11.64          \\ \hline
\end{tabular}
\label{tab:cooking_results}
\vspace{-6pt}
\end{table*}

The next step is to select a backbone that is light-weight and offers good performance for the action segmentation task. As stated in the previous section (Sec. \ref{sec:imple_details}), we consider \textit{EfficientNet-B0}, \textit{MobileNet-V2} and \textit{ResNet50} as backbones. Action segmentation results for the MPII cooking dataset with these backbones applied to the four models, MS-TCN (Sec. \ref{sec:mstcn_info}), MS-TCN++ (Sec. \ref{sec:mstcnplus_info}), SSTDA (Sec. \ref{sec:sstda_info}) and DTGRM (Sec. \ref{sec:DTGRM_info}) are reported in Table~\ref{tab:cooking_results}.

Overall, the DTGRM model with EfficientNet features shows a high level of performance, achieving the best accuracy and F1 scores, although the edit score is 1.91\% lower than the results obtained through SSTDA with a mobileNet backbone. Considering the performance across feature extraction backbones, no single backbone is the most effective across all action segmentation models. However, we observe that MS-TCN, MS-TCN++ and DTGRM all obtained better results when using the EfficientNet backbone, while SSTDA achieves the best results with MobileNet features. We also note that in some instances, the cropped features (or Bbox) have a positive impact on the results. For example, MS-TCN using ResNet50 features has improved by 3.27\% by utilising cropped frame features. However, for the most part, the cropped-frame features tend to slightly degrade the overall results.

Considering this degradation of results, we carried out further experiments by first fine-tuning the feature extraction model with randomly selected cooking data from the Cooking 2 dataset. The fine-tuning for full-frame and cropped frames has been performed separately to align with the overall experiment. Furthermore, we also experimented with expanding the detected bounding box by 10\% before cropping the frames, to capture relevant context information in the images. We report the results in Table~\ref{tab:cooking_results_2}. As shown in Table~\ref{tab:cooking_results_2}, the fine-tuning has significantly improved the results for all the experiments. In some cases, applying 10\% bounding box padding has slightly improved the accuracy. For example, through 10\% padding, the models MS-TCN and DTGRM have achieved a 0.44\% and 0.12\% of accuracy gain respectively. However, when considering the edit score and F1 scores, applying 10\% bounding box padding has resulted in improvements for all the action segmentation models.

\begin{table*}[ht!]
\centering
\caption{Evaluation Results on MPII Cooking 2 Dataset using the fine-tuned MobileNet-V2 feature extraction model. Fine-tuning is performed considering full-frame (Full) and cropped frame features based on object detections (Bbox), and also by applying 10\% bounding box padding. We compare the results with the initial results based on the pre-trained MobileNet-V2 feature extractor (trained on ImageNet data, with no fine-tuning performed).  }
\begin{tabular}{|c|c|l|l|l|l|l|l|}
\hline
Method                    & Full/Bbox             & \multicolumn{1}{c|}{FE} & \multicolumn{1}{c|}{Acc} & \multicolumn{1}{c|}{Edit} & \multicolumn{1}{c|}{F1@0.10} & \multicolumn{1}{c|}{F1@0.25} & \multicolumn{1}{c|}{F1@0.50} \\ \hline
\multirow{5}{*}{MS-TCN}    & \multirow{2}{*}{Full} & pre-trained             & 40.55                    & 29.51                     & 27.82                        & 25.14                        & 18.40                        \\ \cline{3-8} 
                          &                       & fine-tuned              & \textbf{43.82}           & 27.65                     & 25.13                        & 23.28                        & 17.37                        \\ \cline{2-8} 
                          & \multirow{3}{*}{Bbox} & pre-trained             & 37.13                    & 24.06                     & 19.77                        & 18.11                        & 11.28                        \\ \cline{3-8} 
                          &                       & fine-tuned              & 41.23                    & 25.69                     & 23.07                        & 20.66                        & 14.26                        \\ \cline{3-8} 
                          &                       & fine-tuned (10\%) & 41.67                    & 27.38                     & 24.67                        & 22.73                        & 16.77                        \\ \hline
\multirow{5}{*}{MS-TCN ++} & \multirow{2}{*}{Full} & pre-trained             & 41.61                    & 27.91                     & 24.96                        & 22.45                        & 16.07                        \\ \cline{3-8} 
                          &                       & fine-tuned              & 43.11                    & 29.94                     & 27.39                        & 25.46                        & 18.67                        \\ \cline{2-8} 
                          & \multirow{3}{*}{Bbox} & pre-trained             & 37.71                    & 23.25                     & 19.56                        & 17.14                        & 10.56                        \\ \cline{3-8} 
                          &                       & fine-tuned              & 41.08                    & 26.12                     & 22.82                        & 20.10                        & 13.00                        \\ \cline{3-8} 
                          &                       & fine-tuned (10\%) & 41.08                    & 28.24                     & 25.01                        & 23.17                        & 16.78                        \\ \hline
\multirow{5}{*}{SSTDA}    & \multirow{2}{*}{Full} & pre-trained             & 42.50                    & 32.18                     & 26.81                        & 24.52                        & 17.62                        \\ \cline{3-8} 
                          &                       & fine-tuned              & 43.22                    & \textbf{33.31}            & \textbf{29.44}               & \textbf{26.73}               & \textbf{19.20}               \\ \cline{2-8} 
                          & \multirow{3}{*}{Bbox} & pre-trained             & 35.50                    & 23.89                     & 20.33                        & 17.84                        & 10.04                        \\ \cline{3-8} 
                          &                       & fine-tuned              & 38.78                    & 24.26                     & 21.36                        & 18.66                        & 10.89                        \\ \cline{3-8} 
                          &                       & fine-tuned (10\%) & 38.08                    & 24.83                     & 21.76                        & 18.93                        & 10.12                        \\ \hline
\multirow{5}{*}{DTGRM}    & \multirow{2}{*}{Full} & pre-trained             & 41.96                    & 29.36                     & 25.80                        & 24.00                        & 18.02                        \\ \cline{3-8} 
                          &                       & fine-tuned              & 41.98                    & 28.16                     & 26.09                        & 24.11                        & 18.29                        \\ \cline{2-8} 
                          & \multirow{3}{*}{Bbox} & pre-trained             & 37.57                    & 24.58                     & 20.42                        & 18.97                        & 11.64                        \\ \cline{3-8} 
                          &                       & fine-tuned              & 42.62                    & 23.57                     & 17.39                        & 15.78                        & 11.40                        \\ \cline{3-8} 
                          &                       & fine-tuned (10\%) & 42.74                    & 23.79                     & 18.69                        & 17.18                        & 12.23                        \\ \hline
\end{tabular}
\label{tab:cooking_results_2}
\vspace{-6pt}
\end{table*}

\subsection{Adapting to Real-World Applications}
\label{sec:adapting_real}

The action segmentation methods presented in Table \ref{tab:cooking_results} are challenging to directly adapt to real-world applications. One primary challenge is due to the temporal modelling structure that these models employ which requires us to provide a sequence of frames instead of a single frame. Specifically, these methods operate over frame buffers which are sequentially filled with the observed frames, and thus depending on how the buffering is performed and the size of the buffer, there may be delays between the first observation of actions being performed, and when predictions are made by the model are reported. In this section, we investigate and propose a pipeline to apply an action segmentation method to a real-world task, where the current action should be detected with low latency.

\begin{figure*}[ht!]
    \centering
    \includegraphics[width=0.55\linewidth]{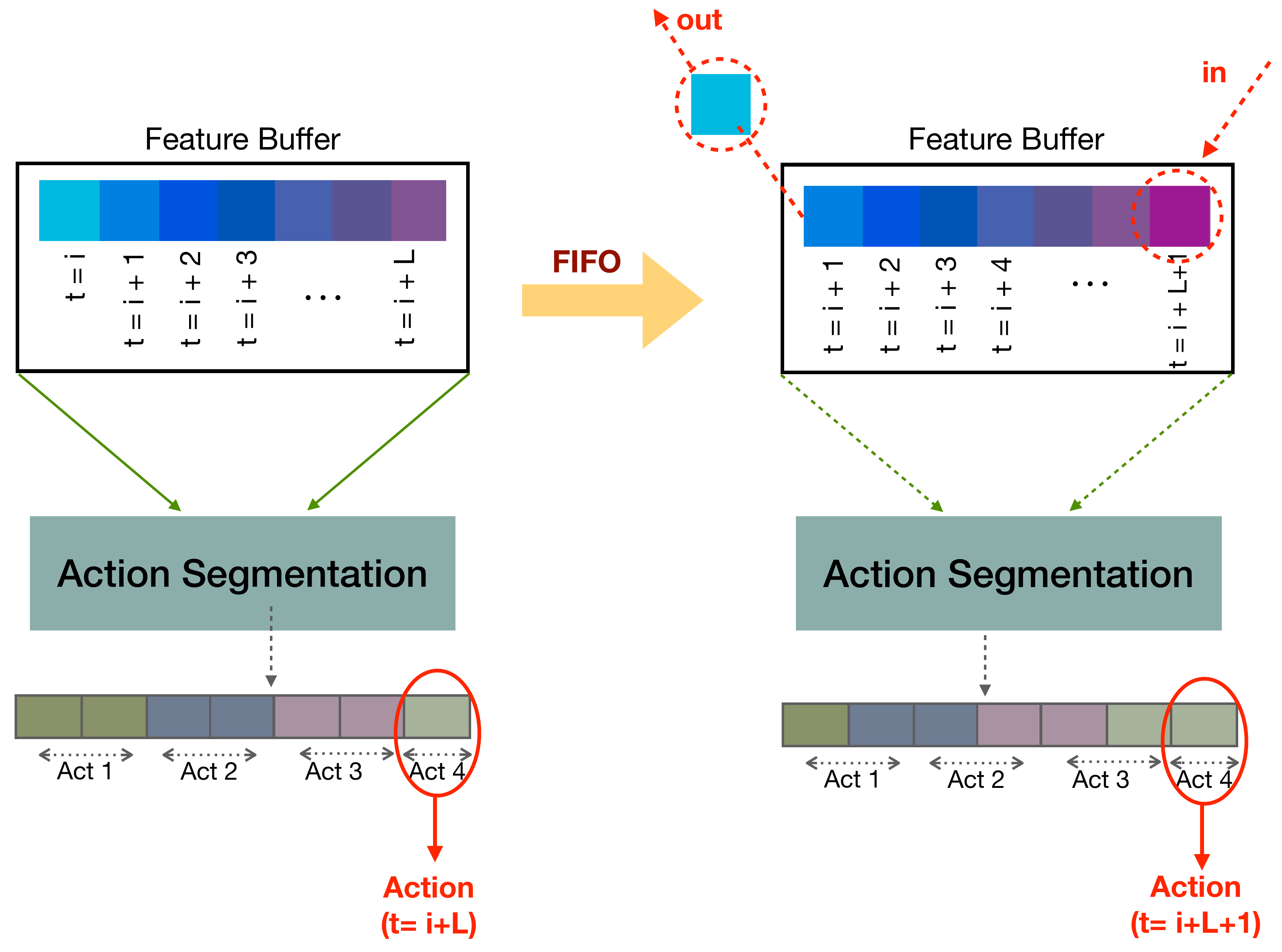}
	\caption{Adapting the feature buffer to a real world setting: the initial buffer (left) and the updated buffer (right) are shown. The buffer is updated to function in a first in, first out (FIFO) manner by removing the oldest feature vector while the feature vector of the latest frame is pushed into the end of the buffer. Once the buffer is updated, the features within the buffer are passed through the Action segmentation model and the prediction corresponding to the latest frame is considered to be the recognised action.}
	\label{fig:framework_real}
\end{figure*} 

In a real-world setting when the action is temporally evolving, it is important to be able to accurately estimate the current action. Action segmentation pipelines use a buffer of frames to provide spatio-temporal context and improve segmentation accuracy. To enable low-latency action segmentation, we use a first-in-first-out (FIFO) buffer, which allows the action segmentation models to observe the temporal progression of the action, and recognise the new actions/behaviours as they occur. The FIFO buffer is updated by removing the oldest feature vector while the features vector of the latest frame is pushed into the end of the buffer. This is illustrated in Fig. \ref{fig:framework_real}. When the feature buffer is updated, the buffer is passed through the action segmentation model to generate the action predictions. 

The buffer length has a crucial impact on the accuracy and the throughput of the action segmentation model. If the buffer length is large, the action segmentation model has a larger temporal receptive field, however, when processing the buffer regularly it leads to redundant predictions as it is repetitively predicting a large number of frames. Furthermore, when the buffer length is $b$ frames, there exists a $b$ frame delay between the observation of the first frame and the first prediction made by the action segmentation model, as the buffer needs to be filled in order to invoke the action segmentation model. 

In a real-world system our interest is the current action (\textit{i.e}. action observed in the most recent frames). For example, in Fig. \ref{fig:framework_real} the predictions at $t=i+L$ and $t=i+L+1$ are considered to be the current action prediction. 

\begin{table*}[ht!]
\centering
\caption{Evaluation results on the MPII Cooking 2 dataset based on varying buffer lengths: The results are compared against the results obtained with original sequence lengths (by maintaining a variable length buffer) where the feature extraction backbone is fine-tuned with 10\% bounding box padding applied.}
\begin{tabular}{|l|c|c|c|c|c|c|}
\hline
\multicolumn{1}{|c|}{\textbf{Method}} & \multicolumn{1}{l|}{\textbf{Length}} & \multicolumn{1}{l|}{\textbf{Acc}} & \multicolumn{1}{l|}{\textbf{Edit}} & \multicolumn{1}{l|}{\textbf{F1@0.10}} & \multicolumn{1}{l|}{\textbf{F1@0.25}} & \multicolumn{1}{l|}{\textbf{F1@0.50}} \\ \hline
\multirow{6}{*}{MS-TCN}                & 100                                  & 37.87                             & 19.72                              & 15.39                                 & 14.33                                 & 11.46                                 \\ \cline{2-7} 
                                      & 250                                  & 39.13                             & 19.64                              & 18.63                                 & 16.83                                 & 12.31                                 \\ \cline{2-7} 
                                      & 500                                  & 40.01                             & 19.42                              & 20.18                                 & 17.94                                 & 13.22                                 \\ \cline{2-7} 
                                      & 750                                  & 40.00                             & 17.85                              & 22.35                                 & 19.73                                 & 13.60                                 \\ \cline{2-7} 
                                      & 1000                                 & 40.17                             & 20.44                              & 23.23                                 & 21.00                                 & 15.19                                 \\ \cline{2-7} 
                                      & \multicolumn{1}{l|}{Original}        & \textbf{41.67}                    & \textbf{27.38}                     & \textbf{24.67}                        & \textbf{22.73}                        & \textbf{16.77}                        \\ \hline
\multirow{6}{*}{MS-TCN++}              & 100                                  & 39.59                             & 19.49                              & 14.88                                 & 13.52                                 & 10.92                                 \\ \cline{2-7} 
                                      & 250                                  & 40.48                             & 19.38                              & 17.41                                 & 15.52                                 & 11.85                                 \\ \cline{2-7} 
                                      & 500                                  & 40.97                             & 20.07                              & 21.15                                 & 18.87                                 & 14.08                                 \\ \cline{2-7} 
                                      & 750                                  & 38.24                             & 18.39                              & 22.49                                 & 20.33                                 & 13.44                                 \\ \cline{2-7} 
                                      & 1000                                 & 38.82                             & 21.32                              & 23.95                                 & 21.10                                 & 13.62                                 \\ \cline{2-7} 
                                      & Original                             & \textbf{41.08}                    & \textbf{28.24}                     & \textbf{25.01}                        & \textbf{23.17}                        & \textbf{16.78}                        \\ \hline
\multirow{6}{*}{SSTDA}                & 100                                  & 31.76                             & 19.14                              & 18.40                                 & 16.21                                 & 7.45                                  \\ \cline{2-7} 
                                      & 250                                  & 36.56                             & 23.20                              & 20.00                                 & 17.01                                 & 8.31                                  \\ \cline{2-7} 
                                      & 500                                  & 36.70                             & 23.22                              & 20.67                                 & 17.03                                 & 9.63                                  \\ \cline{2-7} 
                                      & 750                                  & 37.75                             & 23.97                              & 20.73                                 & 17.33                                 & 9.97                                  \\ \cline{2-7} 
                                      & 1000                                 & 37.84                             & 24.09                              & 20.97                                 & 18.09                                 & 9.98                                  \\ \cline{2-7} 
                                      & Original                             & \textbf{38.08}                    & \textbf{24.83}                     & \textbf{21.76}                        & \textbf{18.93}                        & \textbf{10.12}                        \\ \hline
\multirow{6}{*}{DTGRM}                & 100                                  & 35.29                             & 20.75                              & 16.63                                 & 15.23                                 & 12.36                                 \\ \cline{2-7} 
                                      & 250                                  & 37.83                             & 17.29                              & 14.79                                 & 13.09                                 & 9.67                                  \\ \cline{2-7} 
                                      & 500                                  & 38.27                             & 17.31                              & 16.35                                 & 14.38                                 & 10.35                                 \\ \cline{2-7} 
                                      & 750                                  & 41.28                             & 17.71                              & 16.81                                 & 14.90                                 & 10.60                                 \\ \cline{2-7} 
                                      & 1000                                 & 40.57                             & 17.67                              & \textbf{19.77}                        & \textbf{18.05}                        & \textbf{13.64}                        \\ \cline{2-7} 
                                      & Original                             & \textbf{42.74}                    & \textbf{23.79}                     & 18.69                                 & 17.18                                 & 12.23                                 \\ \hline
\end{tabular}
\label{tab:buffer_len}
\end{table*}

As illustrated above, buffer size plays a crucial role in this real-world setup.  We conduct experiments with different buffer lengths to investigate the optimal attributes that are required to achieve accurate action recognition. We report the corresponding results in Table~\ref{tab:buffer_len} by evaluating the models with buffer lengths of 100, 250, 500, 750 and 1000, and we compare the results with those obtained on the original video sequences (by maintaining a variable length buffer). We use the fine-tuned feature extractor backbone with 10\% bounding box padding applied. We observe a slight drop in results, especially in the accuracies and the edit scores, when the buffer length is reduced from the original sequence length to a buffer length of 1000. However, DTGRM achieves a higher F1 segmental score when the buffer length is 1000, while for the other models the highest results are obtained by maintaining the original sequence lengths. Overall, when the buffer lengths are reduced the action segmentation results drop. However, in some instances, the models achieved slightly higher results on shorter buffer lengths than longer ones. For example, MS-TCN++ achieves better accuracy and edit score values when the buffer lengths are 500 and 250 than the corresponding results on 750 and 1000 length buffers. Similarly, for DTGRM maintaining a 750 length buffer achieved higher accuracy and edit score values compared to maintaining a 1000 length buffer. Even though these scenarios with longer buffer lengths are still able to produce higher F1 segmental scores, compared to the results where buffer lengths are 1000, we noticed a slight deduction on F1 scores on DTGRM results when the original lengths are maintained. 

\begin{table*}[!t]
\caption{Evaluation times in seconds for different buffer sizes. Note that runtime is calculated only for the evaluation of the action segmentation model. Feature extraction time is not taken into consideration.}
\resizebox{\textwidth}{!}{%
\begin{tabular}{|c|ccc|ccc|ccc|ccc|}
\hline
\textbf{Model}         & \multicolumn{3}{c|}{\textbf{MSTCN}}                                                   & \multicolumn{3}{c|}{\textbf{MSTCN++}}                                                 & \multicolumn{3}{c|}{\textbf{SSTDA}}                                                   & \multicolumn{3}{c|}{\textbf{DTGRM}}                                                   \\ \hline
\textbf{Buffer Size}   & \multicolumn{1}{c|}{\textbf{100}} & \multicolumn{1}{c|}{\textbf{500}} & \textbf{1000} & \multicolumn{1}{c|}{\textbf{100}} & \multicolumn{1}{c|}{\textbf{500}} & \textbf{1000} & \multicolumn{1}{c|}{\textbf{100}} & \multicolumn{1}{c|}{\textbf{500}} & \textbf{1000} & \multicolumn{1}{c|}{\textbf{100}} & \multicolumn{1}{c|}{\textbf{500}} & \textbf{1000} \\ \hline
\textbf{Runtimes (ms)} & \multicolumn{1}{c|}{112.9}        & \multicolumn{1}{c|}{729.1}        & 1410.5        & \multicolumn{1}{c|}{169.8}        & \multicolumn{1}{c|}{719.6}        & 1470.6        & \multicolumn{1}{c|}{387.7}        & \multicolumn{1}{c|}{1675.4}       & 3225.8        & \multicolumn{1}{c|}{1601.4}       & \multicolumn{1}{c|}{8830.6}       & 24819.8       \\ \hline
\end{tabular}}
\label{tab:runtimes}
\end{table*}

In Tab. \ref{tab:runtimes} we provide evaluation times in seconds for models with different buffer sizes, which were introduced in Tab. \ref{tab:buffer_len}. Note that runtimes are calculated only for the evaluation of the action segmentation models. Feature extraction time is not taken into consideration, as all the models use MobileNet-V2 feature extractor and thus have a constant per-frame feature extraction cost. We would like to highlight the trade off between accuracy and shorter runtimes. Even though models with shorter buffer sizes are less robust, they have higher throughput and depending on the application requirements they may be more appropriate.

\section{Limitations and Future directions}
\label{sec:limitations}


In this section, we outline the limitations of existing state-of-the-art human action segmentation techniques, and discuss various open research questions and highlight future research directions.

\subsection{Interpretation of the Action Segmentation Models}

Models that result from deep learning are hard to interpret, as most of the decisions are made in an end to end manner and models are highly parameterised. As such, model interpretation plays a crucial role when making black-box deep learning models transparent. However, to the best of our knowledge, no existing state-of-the-art deep learning action segmentation methods have utilised model interpretation techniques, or tried to explain why certain model decisions are made. As such, in some instances it is hard to evaluate whether a given model is making informed predictions, or whether it is simply memorising the data via the high model capacity.  

We believe one of the major hindrances in generating such interpretation outputs is the lack of a model agnostic spatio-temporal model interpretation pipeline. For instance, off-the-shelf deep model interpretation frameworks such as GradCAM \cite{selvaraju2017grad}, Local Interpretable Model-agnostic Explanations (LIME) \cite{ribeiro2016should}, and Guided backpropagation (GBP) \cite{springenberg2014striving} are widely used to interpret deep spatial models. However, spatio-temporal deep model interpretation developments are not as advanced. Therefore, more research is invited towards the design of model agnostic spatio-temporal model interpretation strategies. Furthermore, attention should be paid to the end-users when designing such model interpretation mechanisms. For instance, a given interpretation pipeline can be informative for machine learning practitioners, however may not be useful to the end-users of the action segmentation system due to their lower technical knowledge, and hence may not serve to build his or her trust regarding the system. 

\subsection{Model Generalisation}

One of the major obstacles that is regularly faced during the deployment of an action segmentation model in a real-world setting is a mismatch between the training and deployment environment, which leads the deployed model to generate erroneous decisions. This mismatch can be due to changes in the data capture settings, operating conditions, or changes in the sensor types and modalities. With particular respect to action segmentation models, differences between the humans observed in the training and testing data and the way they perform actions is also a major source of mismatch. Therefore, the design of generalisable models plays a pivotal role within the real-world deployment of human action segmentation models. Nevertheless, to the best of our knowledge, SSTDA \cite{chen2020action} is the only model among recent state-of-the-art methods which tries to compensate for this mismatch. In particular, the SSTDA model tries to overcome subject level differences observed when performing the same action, and learn a unique feature representation that is invariant to such differences. We note that none of the recent literature has investigated how to overcome other domain shifts such as changes in operating conditions, or sensor modalities. 

We propose meta-learning \cite{coskun2021domain} and domain generalisation \cite{zhu2018towards} as two fruitful pathways for future research towards the design of real-world deployment-ready action segmentation frameworks. Meta-learning is a subfield of transfer learning and focuses on how a model trained for a particular task can be quickly adapted to a novel situation, without completely re-training on the data. On the other hand, domain generalisation offers methods that may allow a model trained for the same task to be adapted to a particular sub-domain (\textit{i.e.} data with domain shift). For instance, in the action segmentation setting this could involve adapting a model trained using the videos from an overhead camera to work with videos captured from a head mounted camera with a comparatively narrow field of view. Via domain generalisation such a model could be rapidly adapted to work with first-person view videos without significant tuning to the new domain.  

\subsection{Deployment on Embedded Systems}

In the past few years, the focal point of human action segmentation research has been on attaining higher accuracies which have led to deeper, more complex, and computationally expensive architectures. To date, direct deployment of human action segmentation pipelines on onboard embedded hardware for applications such as robotics has been deemed infeasible. In such a setting, human detection, tracking, a feature extraction backbone, and action segmentation models are all necessary to be of use as this entire pipeline should run in real-time to allow for timely decisions. Therefore, mechanisms to reduce model complexity and/or to improve the computational power and throughput of embedded systems are of benefit. 

We observe that this challenge can be solved using both hardware and software augmentations. For instance, Vision Processing Units such as an Intel Movidius Myriad-X \footnote{https://www.intel.com.au/content/www/au/en/products/details/processors/ movidius-vpu/movidius-myriad-x.html} or a Pixel Visual Core \footnote{https://blog.google/products/pixel/pixel-visual-core-image-processing-and-machine-learning-pixel-2/} can be coupled with the onboard evaluation hardware which will enable fast inference of deep neural networks. Furthermore, deep learning acceleration libraries such as NVIDIA TensorRT \footnote{https://developer.nvidia.com/tensorrt} can be used to augment this throughput and optimise large networks for embedded systems. Most importantly, mechanisms to reduce the complexity of deep models without impacting their performance is an important avenue for research. Model pruning strategies can be used to remove nodes that do not have a significant impact on the final decision layer of the model. Another interesting paradigm we suggest is to reuse the feature extraction backbones for subsequent evaluations. For instance, an action segmentation model can share the same features that the human detector or tracker uses, which would avoid multiple feature extraction evaluation steps within a pipeline. Further research can be conducted to evaluate the viability of such feature sharing. 

\subsection{Optimising Repeated Predictions from a Feature Buffer}

As illustrated in Sec. \ref{sec:adapting_real}, a FIFO buffer is used when executing a temporal model (such as an action segmentation model) in real-world environments. We observe that it is redundant to make repetitive predictions across the entire sequence of frames within the frame buffer, yet it is also critical that a model has access to a sufficiently large temporal window to understand how behaviours are evolving which aids prediction. 

We conducted a preliminary evaluation of the effect of the buffer size on model performance (See Table~\ref{tab:buffer_len}), and these results suggest that there is a considerable fluctuation in performance of state-of-the-art models with different buffer sizes. It is unreasonable to assume a fixed buffer size across applications, as the buffer size is an application-specific parameter, therefore, further investigation should be conducted to minimise the dependency of those models on the buffer size.

Another interesting avenue for future research is how to adapt the buffer-model evaluation pipeline to facilitate real-time response. In such a setting, the action segmentation model has to be invoked at every observed frame or batch of frames. However, as the model has already made predictions about the majority of earlier frames in the buffer, a mechanism that reuses previously obtained knowledge and directs computational resources to the more recent frames is desirable. The use of custom loss functions that give more emphasis to recent frames in the observed sequence to improve their classification accuracy is one such possible approach. Such mechanisms can be investigated in future research, and would be valuable for adapting state-of-the-art methods for real-world applications.  

\subsection{Handling Unlabelled and Weakly Labelled Data}

Despite the increase in publicly available datasets for action segmentation tasks, they remain substantially smaller in size than large-scale datasets such as ImageNet. Therefore, it is vital that action segmentation models have the ability to leverage unlabelled or weakly labelled data. This has added benefits with respect to adapted models to different environments where it is costly and infeasible to hand annotate large training corpora.  Some preliminary research in this direction is presented in \cite{gammulle2020fine, singh2021semi, terao2020semi}. For instance in \cite{gammulle2020fine} GANs are used to perform learning in a semi-supervised setting, exploiting both real-training data and synthesised training data obtained from the generator of the GAN, where as in \cite{singh2021semi, terao2020semi} the authors apply pseudo-labelling strategy to generate one-hot labels for unlabelled examples. 

Furthermore, weakly supervised action segmentation is a newly emerging domain and has gained a lot of attention within the action segmentation community. Specifically, in a weakly supervised action segmentation system, lower levels of supervision are provided to the model. For instance, the supervision signal may simply be a list of actions without any information regarding their order or how many repetitions of each action occurs, or annotations may be presented as a list of actions without individual action start end times. One popular approach to solve this task has been to generate pseudo ground truth labels and iteratively refine them~\cite{richard2017weakly, ding2018weakly}. A new line of work called time-stamp supervision \cite{li2021temporal} is also emerging within the weakly-supervised action segmentation domain. Here ground truth action classes for different segments of a video are provided and the goal is to determine action transition boundaries. Despite these recent efforts, the domain is less explored and the accuracy gap between the semi-supervised, weakly-supervised models and fully supervised models is large. Hence, further research in this direction to reduce this accuracy gap is encouraged. 

\subsection{Incorporating Background Context Together with Human Detections}

In a multi-human action segmentation setting, one potential challenge that we observe is the loss of background context when the features from cropped human detections are fed to an action segmentation model. Depending on the predicted bounding boxes, the action segmentation model may fail to capture information from the objects in the background, or objects that the human is interacting with. This information is crucial to properly recognise the ongoing action. Therefore, mechanisms to incorporate this information into the action segmentation pipeline should be investigated. 

A naive approach would be to expand the size of the detected bounding box such that information from the surroundings is also captured. However, this could lead to the inclusion of misleading information, such as information from the other humans in the surroundings if the environment is cluttered. We suggest following more informative pathways for further investigations: i) explicitly detecting the interacting objects and feeding these features as additional feature vectors to the model; ii) capturing the features from a scene parsing network, such as an image segmentation model, and propagating these features to the action segmentation model such that it can see the global context of the scene. 

A preliminary investigation towards the first pathway is presented in prior works \cite{gammulle2018multi, dasgupta2021context, kazakos2021little}. For instance, in the group activity recognition setting person-level features and the scene-level features are incorporated into the action recognition model in \cite{gammulle2018multi} where as in \cite{dasgupta2021context} the authors extend this idea to utilise pose context as well. However, these works try to recognise actions that are already segmented videos. 
Towards this end a mechanism to extract temporal context is purposed in \cite{kazakos2021little} for  egocentric action segmentation, however, more sophisticated modelling schemes are welcomed in order to capture the complete background context across the entire unsegmented video.  

%

\bibliographystyle{IEEEtran}
\bibliography{sample-base_main}

\appendices

\subfile{supp.tex}

%% file: supp.tex
\title{APPENDIX}

\maketitle

\clearpage
\begin{center}
\textbf{Continuous Human Action Recognition for Human-Machine Interaction: A Review \\ Supplementary Material}
\end{center}

\section{Feature Extraction}

In practice, for feature extraction a pre-trained CNN is commonly utilised. Early works~\cite{bilen2016dynamic,busto2018open,gao2018im2flow} used pre-trained networks such as ALexNet~\cite{krizhevsky2012imagenet}, which are comparatively shallow networks with only 5 convolutional layers. More recent methods \cite{liu2018global,gammulle2017two,zhang2016video} have used deeper architectures such as VGG~\cite{simonyan2014very} and GoogleNet~\cite{szegedy2015going}. Even though deeper models can learn more discriminative features than shallow networks, deeper network can struggle during training due to problems such as vanishing gradients caused by the increased network depth.

The following subsections will discuss recent backbones that are commonly used within the action segmentation domain.

\subsubsection{Residual Neural Network (ResNet)}

\begin{figure}[!t]
    \centering
    \includegraphics[width = 0.5\linewidth]{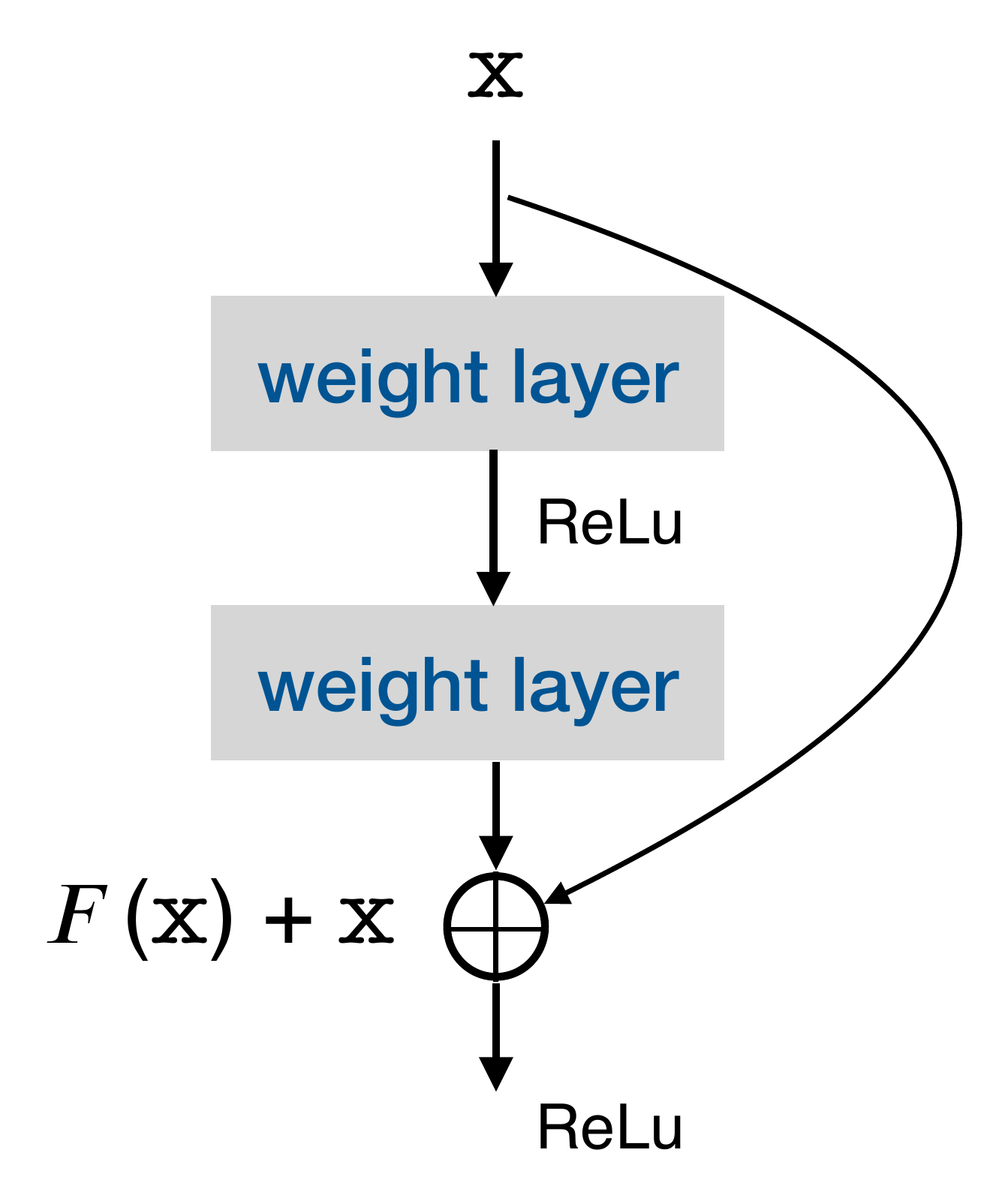}
    \vspace{-8pt}
    \caption{Residual Block Formulation. Adapted from \cite{he2016deep}}
    \label{fig:id_map}
    \vspace{-10pt}
\end{figure}

\begin{figure}[!t]
    \centering
    \includegraphics[width = 1\linewidth]{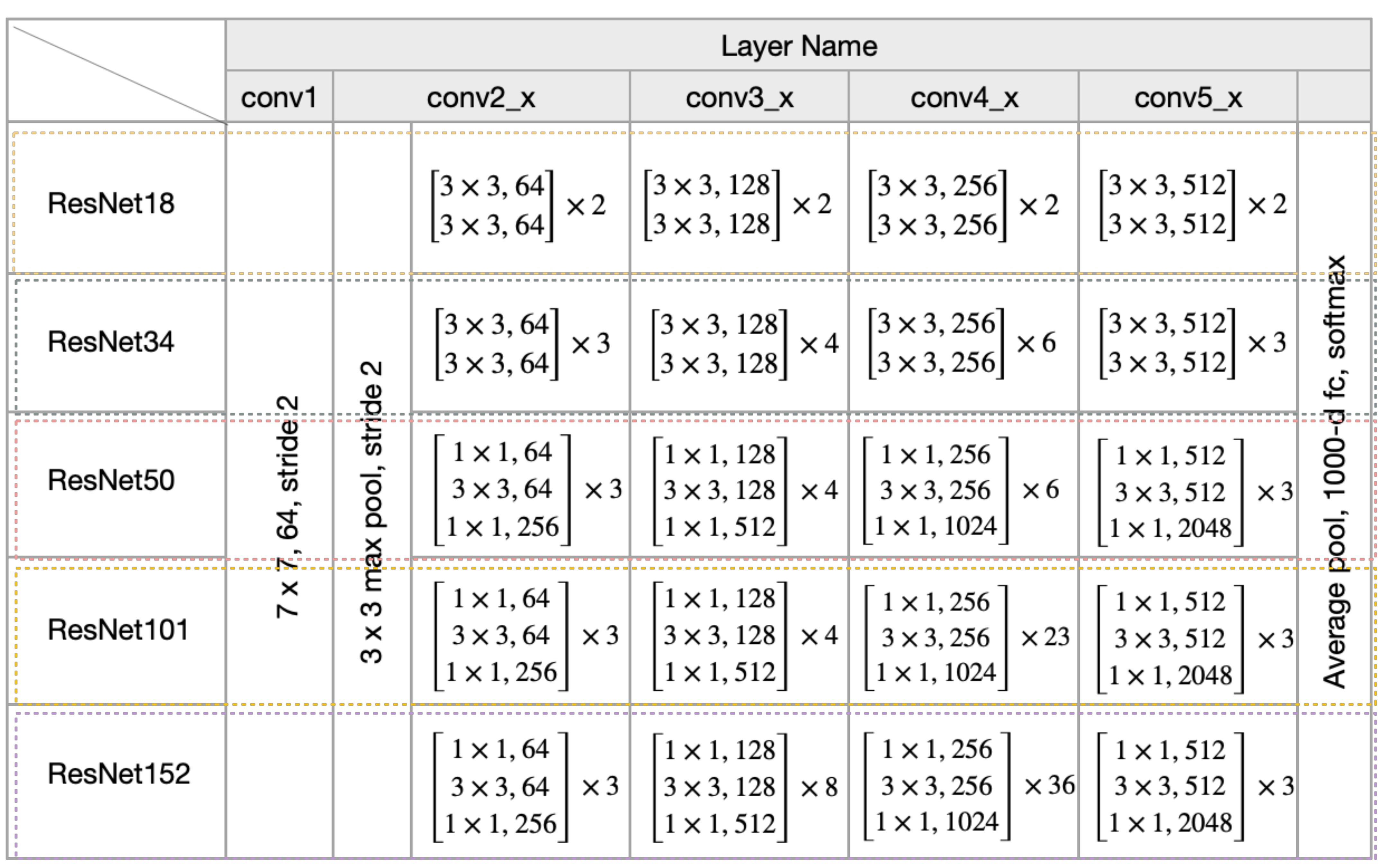}
    \vspace{-8pt}
    \caption{Residual Network Architectures: Networks with a depth of 18, 34, 50, 101, and 152 layers. Adapted from \cite{he2016deep}}
    \label{fig:resnets}
    \vspace{-8pt}
\end{figure}

In \cite{he2016deep}, the proposed residual block with identity mapping has shown to overcome limitations of previous networks and make the training of deeper networks far easier. Fig. \ref{fig:id_map} shows the residual block formulation where the identity mapping is achieved through the skip connection, which adds the output of an earlier layer to a later layer without needing any additional parameters. By utilising this identity mapping, they have shown that residual networks are easier to optimise and can achieve better accuracies in networks with increased depth. The authors introduced multiple residual nets with depths of 18, 34, 50, 101, and 152 layers, and the architectural details are included in Fig. \ref{fig:resnets}. These ResNet models are less complex (in terms of parameters) compared to previous pre-trained networks such as VGG networks, where even the deeper ResNet network (\textit{i.e.} ResNet152) is less complex (11.3 billion FLOPs\footnote{Floating Point Operations Per Second}) than both 16 and 19 layer VGG variants (\textit{i.e.} VGG16 and VGG19) which have  15.3 and 19.6 billion FLOPS respectively. 

Due to the aforementioned advantages, pre-trained ResNet networks have been widely used as a feature extraction backbone within both the  action recognition domain \cite{vats2020event,hussein2019timeception,gammulle2020hierarchical,gammulle2019predicting, gammulle2019forecasting} and related  problem domains \cite{liu2020adaptive,bhunia2021joint,gammulle2020two}. 

\subsubsection{EfficientNet-B0}

In \cite{tan2019efficientnet}, the authors proposed a CNN architecture which could both improve accuracy and efficiency through a reduction in the number of parameters and FLOPS factor compared to previous models such as GPipe \cite{huang2019gpipe}. In particular, they proposed an effective compound scaling mechanism to increase the model capacity, facilitating improved accuracy. Among their models, EfficientNet-B0 is the simplest and most efficient model, and achieves 77.3\% accuracy on ImageNet while having only 5.3M parameters and 0.39B FLOPS. In comparison, ResNet50 achieves only 76\% accuracy \cite{tan2019efficientnet} despite a substantially larger number of trainable parameters (\textit{i.e.} 26M trainable parameters and 4.1B FLOPS). 

The main building block of the EfficientNet-B0 architecture is the mobile inverted bottleneck MBConv, to which a squeeze-and-excitation optimisation is added. MBConv has similarities to the inverted residual block used in MobileNet-V2 (details in Sec.\ref{mob_net_info}), and contains skip connections between the start and the end of a convolutional block. Table~\ref{tab:effNet_arc} shows the network architecture of EfficientNet-B0.

\begin{table}[!t]
\caption{The Network Architecture of EfficientNet-B0.}
\resizebox{0.46\textwidth}{!}{%
\begin{tabular}{|c|c|c|c|c|}
\hline
\textbf{Stage} & \multicolumn{1}{l|}{\textbf{Operator}} & \multicolumn{1}{l|}{\textbf{Resolution}} & \multicolumn{1}{l|}{\textbf{\#Channels}} & \multicolumn{1}{l|}{\textbf{\#Layers}} \\ \hline
1         & Conv3$\times$3                          & 224 $\times$ 224                       & 32                              & 1                             \\ \hline
2         & MBConv1, k3$\times$3                    & 112 $\times$ 112                       & 16                              & 1                             \\ \hline
3         & MBConv6, k3$\times$3                    & 112 $\times$ 112                       & 24                              & 2                             \\ \hline
4         & MBConv6, k5$\times$5                    & 56 $\times$ 56                         & 40                              & 2                             \\ \hline
5         & MBConv6, k3$\times$3                    & 28 $\times$ 28                         & 80                              & 3                             \\ \hline
6         & MBConv6, k5$\times$5                    & 14 $\times$ 14                         & 112                             & 3                             \\ \hline
7         & MBConv6, k5$\times$5                    & 14 $\times$ 14                         & 192                             & 4                             \\ \hline
8         & MBConv6, k3$\times$3                    & 7 $\times$ 7                           & 320                             & 1                             \\ \hline
9         & Conv1$\times$1 \& Pooling \& FC         & 7 $\times$ 7                           & 1280                            & 1                             \\ \hline
\end{tabular}}
\vspace{3pt}
\label{tab:effNet_arc}
\vspace{-8pt}
\end{table}

\begin{figure}[!t]
    \centering
    \includegraphics[width = 1\linewidth]{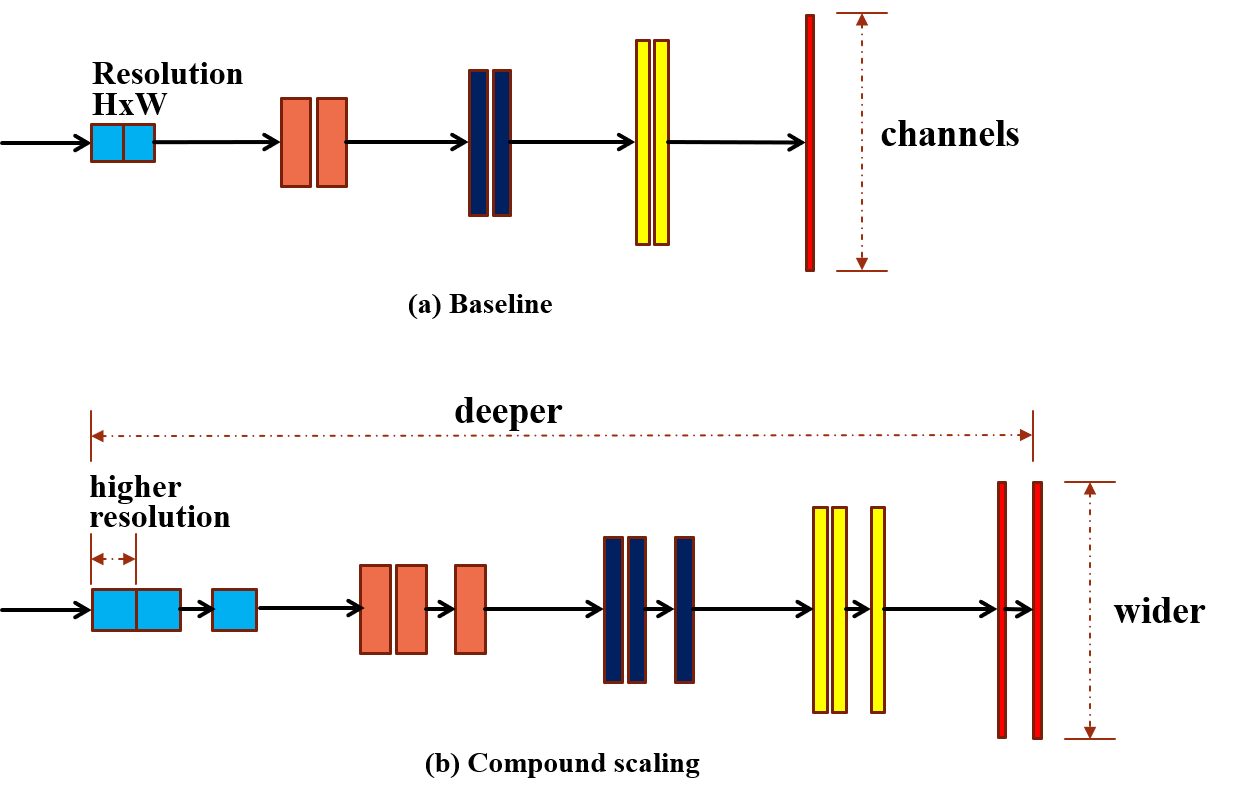}
    \vspace{-8pt}
    \caption{Compound Scaling of EfficientNet: Uniformly scale all three dimensions, including width, depth and resolution with a fixed ratio. Recreated from \cite{tan2019efficientnet}}
    \label{fig:c_scaling}
    \vspace{-8pt}
\end{figure}

An important component of EfficientNet is the compound scaling method which focuses on uniformly scaling all dimensions (\textit{i.e.} depth, width and resolution) using a compound coefficient. When the depth of the network is individually increased, the networks gain the capacity to learn more complex features while the dimension of width increases the network's ability to learn fine-grained features. Similarly, when the resolution is increased the models are able to learn finer details from the input image (i.e. detect smaller objects and finer patterns). Although each scaling operation improve network performance, care should be taken as accuracy gain and performance can diminish as model size increases. For example, as the depth of the network is continually increased the network can suffer from vanishing gradients and become difficult to train. Therefore, in EfficientNet, instead of performing arbitrary scaling, a method called compound scaling is applied by uniformly scaling the network depth, width and resolution with fixed scaling coefficients. This scaling method is illustrated in Fig. \ref{fig:c_scaling}. As described in \cite{tan2019efficientnet}, if $2^N$ times more computational resources are available, it is only a matter of increasing the network depth by $\alpha^N$, the width by $\beta^N$, and the resolution by $\gamma^N$, where $\alpha$, $\beta$ and $\gamma$ are constant coefficients that are determined by a small grid search operating over the original small model. In EfficientNet, the compound coefficient $\phi$ is used to uniformly scale network width, depth, and resolution where $\phi$ is a user-specified coefficient determining the number of resources available for model scaling; while $\alpha$, $\beta$, $\gamma$ determine how these extra resources will be deployed to increase to network depth, width and resolution, respectively.

The advancements and flexibility offered by the EfficientNet architecture has seen it widely adopted across computer vision as a feature extractor \cite{yin2020using,huo2020lightweight,kim2021efficient}. 

\subsubsection{MobileNet-V2}
\label{mob_net_info}

\begin{figure}[!t]
    \centering
    \includegraphics[width = 0.9\linewidth]{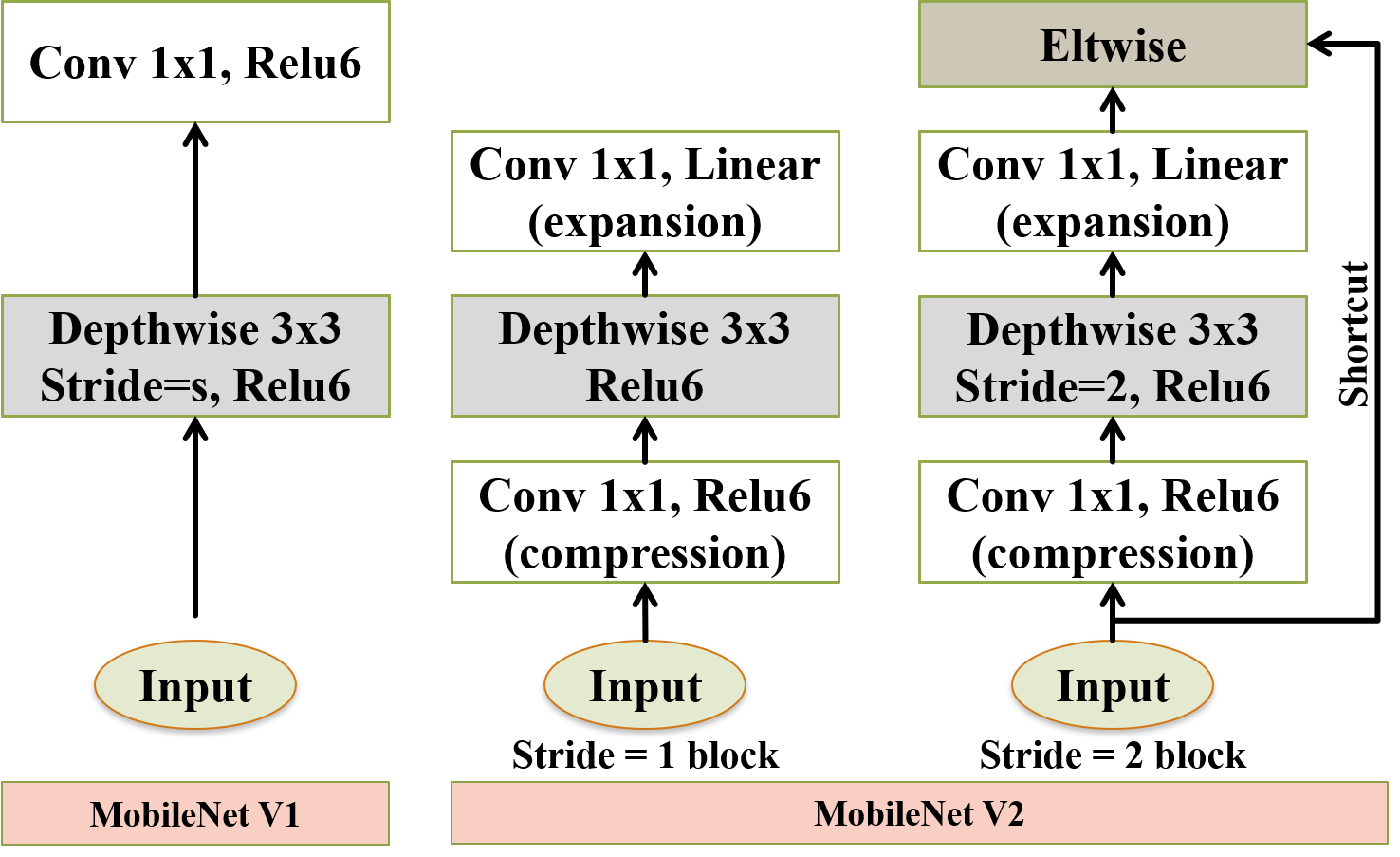}
    \vspace{-6pt}
    \caption{MobileNet V1 and V2 convolutional blocks. Recreated from \cite{howard2017mobilenets}}
    \label{fig:mobnets}
    \vspace{-2pt}
\end{figure}

The initial version of MobileNet, MobileNet-V1~\cite{howard2017mobilenets}, dramatically reduced the computational cost and model complexity in comparison to other networks to enable the use of DCNNs in mobile or low computational power environments. In order to achieve this, MobileNetV1 introduced with Depth-wise Separable Convolutions. Essentially, the model contains 2 layers: depthwise convolutions which perform lightweight filtering by applying a single convolutional filter to each input channel; and a point-wise convolution which is a $1 \times 1$ convolution that learns new features through a linear combinations of the input channels. MobileNetV2 \cite{sandler2018mobilenetv2} builds upon the findings of MobileNetV1 by utilising the Depth-wise Separable Convolutions as the building block, while adding two other techniques to further improve performance: having linear bottlenecks between the layers and shortcut connections between the bottlenecks. Fig. \ref{fig:mobnets} compares the convolutional blocks used in MobileNetV1 and MobileNetV2. Fig. \ref{tab:mobnetv2} describes the MobileNetV2 architecture where t, c, n, s refer to the expansion factor, number of output channels, repeating number and stride respectively. For spatial convolutions, $3 \times 3$ kernels are used. Due to the light-weight model architecture MobileNet is widely used as a feature extraction backbone in applications  in a live or real-time setting \cite{gowda2021smart,zhang2018temporal}.

\begin{table}[t!]
\centering
\caption{MobileNetV2 Architecture. Each layer is repeated \textbf{n} times and \textbf{t}, \textbf{c}, \textbf{s} denote the expansion factor, output channels and strides respectively.}
\begin{tabular}{|c|c|c|c|c|c|}
\hline
\textbf{Input} & \textbf{Operator} & \textbf{t} & \textbf{c} & \textbf{n} & \textbf{s} \\ \hline
$224^2 \times 3$        & conv2d            & -          & 32         & 1          & 2          \\ \hline
$112^2 \times 32$       & bottleneck        & 1          & 16         & 1          & 1          \\ \hline
$112^2 \times 16$       & bottleneck        & 6          & 24         & 2          & 2          \\ \hline
$56^2 \times 24$        & bottleneck        & 6          & 32         & 3          & 2          \\ \hline
$28^2 \times 32$        & bottleneck        & 6          & 64         & 4          & 2          \\ \hline
$14^2 \times 64$        & bottleneck        & 6          & 96         & 3          & 1          \\ \hline
$14^2 \times 96$        & bottleneck        & 6          & 160        & 3          & 2          \\ \hline
$7^2 \times 160$        & bottleneck        & 6          & 320        & 1          & 1          \\ \hline
$7^2 \times 320$        & conv2d $1 \times 1$      & -          & 1280       & 1          & 1          \\ \hline
$7^2 \times 1280$       & avgpool $7 \times 7$     & -          & -          & 1          & -          \\ \hline
$1 \times 1 \times 1280$   & conv2d $1 \times 1$      & -          & k          & -          &            \\ \hline
\end{tabular}
\vspace{5pt}
\label{tab:mobnetv2}
\vspace{-8pt}
\end{table}

\subsubsection{Inflated 3D ConvNet (I3D)}

The previously discussed feature extraction backbones are image-based networks, capturing only appearance information from an input image. However, in video-based action recognition capturing temporal information alongside spatial features is vital. One way of achieving this is by utilising 3D convolutional neural networks. 3D convolutional networks are similar to standard convolutional networks, however 3D convolutional networks are capable of creating a hierarchical structure representing spatio-temporal data through the use of spatio-temporal filters. The limitation of such 3D convolutional networks is that the training process becomes more difficult due to having an additional kernel dimension compared to 2D convolutional networks. Therefore, most previous state-of-the art 3D networks are relatively shallow. For example, the C3D network \cite{tran2015learning} is composed of only 8 convolutional layers.    
In \cite{carreira2017quo}, the authors introduced a novel model, two-stream Inflated 3D ConvNet (I3D), which proposed inflating a 2D CNN (\textit{i.e.} Inception-V1~\cite{szegedy2015going}), such that filters and pooling kernels within an existing deep CNN model are expanded, adding a third dimension such that the network can learn spatio-temporal features from video inputs. A $N \times N$ 2D filter is inflated to an $N \times N \times N$ filter with the addition of a temporal dimension. As outlined in \cite{carreira2017quo}, compared to the C3D network, the I3D architecture is much deeper, yet contains fewer parameters. 

\begin{figure}[t!]
    \centering
    \includegraphics[width = 1\linewidth]{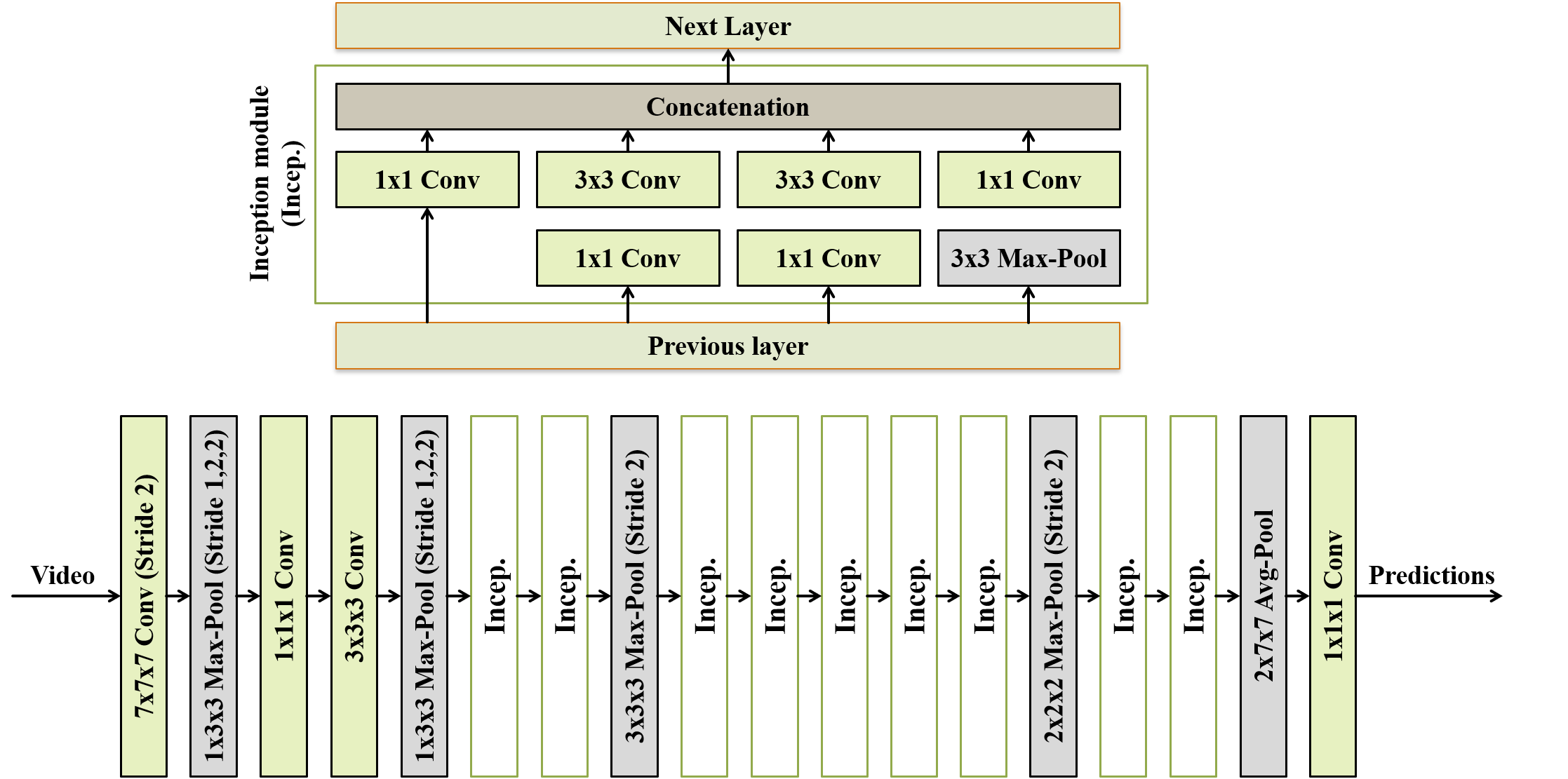}
    \caption{Inflated Inception-V1 Architecture. Recreated from \cite{szegedy2015going}}
    \label{fig:i3d}
    \vspace{-10pt}
\end{figure}

Fig. \ref{fig:i3d} shows the network architecture of Inflated Inception-V1. In the network, the first two pooling layers do not perform temporal pooling where $1 \times 3 \times 3$ kernels with stride 1 are used, while the remaining max-pooling layers have symmetric kernels and strides. However, as stated in \cite{carreira2017quo}, the final model is trained on 64 frame snippets, which can be extracted across a large temporal receptive field. As such, the model captures long-term temporal information through the fusion of both RGB and flow streams. The authors further improved this architecture by individually training two networks on RGB and optical flow inputs respectively and performing average pooling afterwards. This 2-stream architecture has been widely used in action recognition research as a feature extractor. For example, in \cite{farha2019ms,li2020ms,chen2020action} 2048 dimensional feature vectors are extracted by the two networks (RGB and optical flow), where 1024 dimensional features extracted through individual networks are concatenated.

\section{Object Detection and Tracking}

Table~\ref{table:detection_actionrec} lists selected samples from the literature that have demonstrated the benefits of incorporating detection and tracking techniques to aid action recognition.

\begin{table*}[ht!]
\centering
\caption{Example methods which have used object detection to support action recognition or segmentation.}
\resizebox{1\textwidth}{!}{%
\begin{tabular}{
l
>{\raggedright\arraybackslash}p{3.5cm}
>{\raggedright\arraybackslash}p{5cm}
>{\raggedright\arraybackslash}p{2.5cm}
>{\raggedright\arraybackslash}p{4.5cm}}
\toprule
\textbf{Authors} &
\textbf{Detector + Tracker} & 
\textbf{Action rec model} &
\textbf{Dataset} &
\textbf{Additional remarks} \\
\midrule
Wang et al. (2016)~\cite{wang2016two} & Faster R-CNN &
SR-CNN & UCF101 & No Tracking \\ 
Wang et al. (2016)~\cite{zhang2018action} & Faster R-CNN + Distance-based tracker &
Two-stream CNN & J-HMBD, UCF & \\ 
Tsai et al. (2020)~\cite{tsai2020deep} & YOLOv3 + DeepSORT &
I3D~\cite{carreira2017quo} & NTU RGB+D &  \\ 
Ishioka et al. (2020)~\cite{ishioka2020single} & Faster R-CNN + Kalman filter &
I3D~\cite{carreira2017quo} & Construction site &  \\ 
Ali et al. (2022)~\cite{ali2022video} & YOLOv3 + DeepSORT &
I3D~\cite{carreira2017quo} & Activis & Diagnosis of children with autism \\ 
Ahmed et al. (2021)~\cite{ahmed2021automated} & Faster R-CNN + SORT &
MS-TCN++ & SARAH tasks & Stroke rehabilitation assessment \\ 
\midrule
Ahmedt-Aristizabal et al. (2019)~\cite{ahmedt2019understanding} & Mask R-CNN &
Skeleton-based (LSTM Pose Machine) + LSTM; CNN+LSTM & Mater Hospital & Diagnosis of Epilepsy \\ 
Chen et al. (2020)~\cite{chen2020repetitive} & YOLOv3 &
Skeleton-based (CPM) & Assembly action &  \\ 
Ghosh et al. (2020)~\cite{ghosh2020stacked} & Faster R-CNN &
Skeleton-based (STGCN~\cite{yan2018spatial}) & CAD120 & Object detector as feature extractor; Dataset provides skeletal data. \\ 
Li et al. (2021)~\cite{li2021pose} & OpenPose~\cite{cao2019openpose} &
Skeleton-based (PR-GCN~\cite{li2021pose}) & CAD120 &  \\ 
\bottomrule
\end{tabular}}
\label{table:detection_actionrec}
\vspace{-6pt}
\end{table*}

\subsection{Object Detection}
In this Subsection, we introduce some relevant object detector architectures from the literature.

\subsubsection{Anchor-based: Two-stage Frameworks} \hfill

\textit{R-CNN series}. \textit{Faster-RCNN}~\cite{ren2015faster} comprises two modules, an RPN which generates a set of object proposals and an object detection network, based on the \textit{Fast R-CNN} detector~\cite{girshick2015fast}, which refines the proposal locations. 
One extension to Faster R-CNN is the feature pyramid network (\textit{FPN})~\cite{lin2017feature} which provides a robust way to deal with images of different scales, while maintaining real-time performance. 
Another extension is the region fully convolutional network (\textit{R-FCN})~\cite{dai2016r} that seek to make the Faster R-CNN network faster by making it fully convolutional, and delaying the cropping step.
To reduce imbalance at sample, feature, and objective level during training, \textit{Libra R-CNN}~\cite{pang2019libra} is introduced by integrating three components: intersection over union (IoU)-balanced sampling, a balanced feature pyramid, and a balanced L1 loss.
Finally, \textit{Mask R-CNN}~\cite{he2017mask} is an extension of the Faster R-CNN architecture which adds a branch for predicting segmentation masks for each region of interest in parallel with the existing branch for classification and bounding box regression. Mask R-CNN can be seen as a more accurate object detector by including a feature pyramid network.

\textit{Chained cascade network and Cascade R-CNN}~\cite{ouyang2017chained,cai18cascadercnn} are multi-stage extensions of R-CNN frameworks. They consist of a sequence of detectors trained end-to-end with increasing IoU thresholds, to be sequentially more selective and thus reduce false positives.

\textit{TridentNet}~\cite{li2019scale} is an object detector that generates scale-specific feature maps using a parallel multi-branch architecture (trident blocks), in which each branch shares the same transformation parameters, but has differing receptive fields through dilated convolutions.

\subsubsection{Anchor-based: One-stage Frameworks} \hfill

\textit{Single shot multibox detector (SSD)}~\cite{liu2016ssd} is a single-shot detector. As such, it predicts the bounding boxes and classes directly from feature maps in a single pass. To improve accuracy, SSD introduces small convolutional filters to predict object classes, and offsets to a default set of boundary boxes. 
Deconvolutional single shot detector \textit{(DSSD)}~\cite{fu2017dssd} is a modified version of SSD which adds a prediction and deconvolution module. DSSD uses an integration structure similar to the FPN in SSD to generate an integrated feature map.

\textit{RetinaNet}~\cite{lin2017focal} is a detector composed of a backbone network and two task-specific subnetworks. The backbone is responsible for computing a convolutional feature map over an entire input image and is an off-the-self convolution network. The first subnetwork performs classification on the backbones output; the second subnetwork performs convolution bounding box regression. RetinaNet uses an FPN to replace the multi-CNN layers in SSD which integrate features from higher and lower layers in the backbone network.

\textit{M2det}~\cite{zhao2019m2det} uses a multi-level feature pyramid network (MLFPN) to generate more effective feature pyramids. Features with multiple scales and from multiple levels are processed as per the SSD architecture to obtain bounding box locations and classification results in an end-to-end manner.

\textit{EfficientDet}~\cite{tan2020efficientdet} is an object detector that adopts EfficientNet and several optimisation strategies such as the use of a BiFPN to determine the importance of different input features, and a compound scaling method that uniformly scales the depth, width and resolution as well as the box/class network at the same time.

\textit{You only look once (YOLO) family of detectors} 
Among well-known single stage object detectors, \textit{YOLO}~\cite{redmon2016you} and \textit{YOLOv2}~\cite{redmon2017yolo9000}) have demonstrated impressive speed and accuracy. This detector can run well on low powered hardware, thanks to the intelligent and conservative model design.
\textit{YOLOv3}~\cite{redmon2018yolov3} makes an incremental improvement to the YOLO by adopting the backbone network Darknet-53 as the feature extractor.
\textit{YOLOv3-tiny} is a compact version of YOLOv3 that has only nine convolutional layers and six pooling layers, making it much faster but less accurate.
\textit{YOLOv3-SPP} is a revised YOLOv3, which has one SPP module~\cite{he2015spatial} in front of its first detection header. 

Other relevant models on the YOLO series are:

\textit{YOLOv4}~\cite{bochkovskiy2020yolov4}, which  is composed of CSPDarknet53 as a backbone, a SPP additional module, a PANet as the neck, and a YOLOv3 as the head. 
CSPDarknet53 is a novel backbone that can enhance the learning capability of the CNN by integrating feature maps from the beginning and the end of a network stage. The BoF for YOLOv4 backbones include CutMix and Mosaic data augmentation~\cite{yun2019cutmix}, DropBlock regularisation~\cite{ghiasi2018dropblock}, and class label smoothing~\cite{szegedy2016rethinking}. The BoS for the same CSPDarknet53 are Mish activation~\cite{misra2019mish}, cross-stage-partial-connections (CSP)~\cite{wang2019cspnet}, and multi-input weighted residual connections (MiWRC).
Paddle-Paddle YOLO \textit{(PP-YOLO)}~\cite{long2020pp} is based on YOLOv4 with a ResNet50-vd backbone, an SPP for the top feature map, a FPN as the detection Neck, and the detection head of YOLOv3. An optimised version of this model was published as \textit{PP-YOLOv2}~\cite{huang2021pp}, where a PANet is included for the FPN to compose bottom-up paths.
Finally, the scaling cross stage partial network \textit{(Scaled-YOLOv4)}~\cite{wang2021scaled} achieves one of the best trade-offs between speed and accuracy. In this approach, YOLOv4 is redesigned to form YOLOv4-CSP with a network scaling approach that modifies not only the network depth, width, and resolution; but also the structure of the network. Thus, the backbone is optimised and the neck (PANet) uses CSP and Mish activations.

\subsubsection{Anchor-free Frameworks}
As discussed previously, use of an anchor-free mechanism significantly reduces the number of hyperparameters and reduces design choices which need heuristic tuning and expertise, making detector configuration, training and decoding considerably simpler. 
Anchor-free object detectors can be categorised as keypoint based approaches and center-based approaches.
The former predicts predefined key points from the network which are then utilised to generate the bounding box around an object and classify it. The latter employs the center-point or any part-point of an object to define positive and negative samples. From these points it predicts the distance to four coordinates for the generation of a bounding box.

\textit{DeNet}~\cite{tychsen2017denet} is a two-stage detector which first determines how likely it is each location belongs to either the top-left, top-right, bottom-left or bottom-right corner of a bounding box. It then generates ROIs by enumerating all possible corner combinations, and rejects poor ROIs with a sub-detection network. Finally, the model classifies and regresses the ROIs.

\textit{CornerNet}~\cite{law2018cornernet} is a one-stage approach which detects objects represented by paired heatmaps, the top-left corner and bottom-right corner, using a stacked hourglass network~\cite{newell2016stacked}. The network uses associate embeddings and predicts a group of offsets to group corners and produce tighter bounding boxes.
One extension to this model, is \textit{CornetNet-lite}~\cite{law2019cornernet}, which combines two variants (CornerNet-Saccade and CornerNet-Squeeze) to reduce the number of pixels processed and the amount of processing per pixel, respectively. CornerNet-Saccade uses an attention mechanism to eliminate the need for exhaustively processing all pixels of the image, and CornerNet-Squeeze introduces a compact hourglass backbone that leverages ideas from SqueezeNet and MobileNets.

\textit{CenterNet (objects as points)}~\cite{zhou2019objects} considers the center of a box as an object and a key point, and then uses this predicted center to find the coordinates/offsets of the bounding box. The model uses two prediction heads, one to predict the confidence heat map, and the other to predict regression values for box dimensions and offsets. This helps the model remove the NonMaximum Suppression step commonly required during post-processing.
Another variant of this keypoint based approach, \textit{CenterNet (keypoint triplets)}~\cite{duan2019centernet}, detects each object as a triplet using a centre keypoint and a pair of corners. The model introduces two specialised modules, cascade corner pooling and centre pooling, which enrich information acquired by both the top-left and bottom-right corners and extract richer data from the central regions. If a center keypoint is detected in the central region, the bounding box is preserved.

\textit{FCOS}~\cite{tian2019fcos} (fully convolutional one-stage detector) is a center-based approach which computes per-pixel predictions in a fully convolutional manner (\textit{i.e.} perform object detection like segmentation). FCOS is constructed on top of an FPN which acts as a pyramid to aggregate multi-level features from the backbone. FPN predictions are obtained across five feature levels. The outputs are then fed to a detection head (per pixel predictions) consisting of three branches. FCOS introduces a third branch, called the centerness branch, in addition to the two typical branches, classification and regression. The centerness branch provides a measure of how centred the positive sample location is within the regressed bounding box, which improves the performance of anchor-free detectors and brings them on-par with anchor-based detectors.

\textit{YOLOX}~\cite{ge2021yolox} adapts the YOLO series to an anchor-free setting, and incorporates other improvements such as a decoupled head and an advanced label assignment strategy, SimOTA, based on OTA~\cite{ge2021ota}.

Other relevant anchor-free architectures are RepPoints~\cite{yang2019reppoints} and RepPoints v2~\cite{chen2020reppoints}, FSAF~\cite{zhu2019feature}, and ExtremeNet~\cite{zhou2019bottom}.

\subsection{Multi-object Tracking}
In this Subsection, we introduce well-known tracking algorithms under the category of  separate detection and embedding (SDE), and joint detection and embedding (JDE) methods.

\subsubsection{SDE Frameworks} \hfill

\textit{Simple online and realtime tracking (SORT)}~\cite{Bewley2016_sort} is a framework that combines location and motion cues in a very simple way. 
The model uses Kalman filtering~\cite{kalman1960new} to predict the location of the tracklets in the next frame, and then performs the data association using the Hungarian method~\cite{kuhn1955hungarian} with an association metric that measures bounding box overlap.

\textit{DeepSORT}~\cite{Wojke2017simple} augments the overlap-based association cost in SORT by integrating a deep association metric and appearance information. The Hungarian algorithm is used to resolve  associations between the predicted Kalman states and newly-arrived measurements. The tracker is trained in an entirely offline manner. At test time, when tracking novel objects, the network weights are frozen, and no online fine-tuning is required.

\textit{ByteTrack}~\cite{zhang2021bytetrack} keeps all detection boxes (detected by YOLOX) and associates across every box instead of only the high scoring boxes to reduce missed detections. In the matching process, an algorithm called BYTE first predicts the tracklets using a Kalman filter, then they are matched with high-scoring detected bounding boxes using motion similarity. Next, the algorithm performs a second matching between the detected bounding boxes with lower confidence values and the objects in the tracklets that could not be matched.

\subsubsection{JDE Frameworks} \hfill

\textit{Tracktor}~\cite{bergmann2019tracking} uses the Faster-RCNN framework to directly take the tracking results of previous frames as the ROI. The model then removes the box association by directly propagating identities of region proposals using bounding box regression. 

\textit{CenterTrack}~\cite{zhou2020tracking} is based on the CenterNet model~\cite{zhou2019objects}, and predicts the offsets of objects relative to the previous frame while detecting objects in the next frame.

\textit{FairMOT}~\cite{zhang2021fairmot} uses an encoder-decoder network to extract high resolution features from an image, and thus incorporates a Re-ID module within CenterTrack~\cite{zhou2020tracking}. The Re-ID features are an additional target for CenterNet to regress to, in addition to the centre points, the object size, and the offset.

\textit{SiamMOT}~\cite{shuai2021siammot} combines the Faster-RCNN model with two motion models based on siamese-based single-object tracking~\cite{li2019siamrpn++}, an implicit motion model and an explicit motion model. The motion model estimates an instance’s movement between two frames such that detected instances are associated.

Other relevant real-time multi-object trackers that have a JDE framework are QDTrack~\cite{pang2021quasi}, TraDeS~\cite{wu2021track}, CorrTracker~\cite{wang2021multiple},  and transformer-based tracking models (TransTrack~\cite{sun2020transtrack}, MOTR~\cite{zeng2021motr}).

\section{Action Segmentation Datasets}
\label{sec:dataset_info}

In this section, we discuss action segmentation datasets that are widely used in the current literature. 

\vspace{-6pt}
\subsection{Breakfast \cite{Kuehne12}}
This dataset contains 50 fine-grained actions related to breakfast preparation, performed by 52 different subjects in 18 different kitchens. This dataset offers an uncontrolled setting in which to evaluate action segmentation models. This dataset is captured from different camera types, including  webcams, standard industry cameras and stereo cameras. Furthermore, the videos are captured from different viewpoints. The provided dataset has a resolution of 320 $\times$ 240 pixels with a frame rate of 15 fps. 

\vspace{-6pt}
\subsection{50Salads \cite{stein2013combining}}
This dataset offers a multi-modal human action segmentation challenge where the provided data contains RGB video data and Depth maps sampled at 640×480 pixels at 30 fps, as well as 3-axis accelerometer data at 50 fps. This dataset includes 50 sequences of people preparing a mixed salad with two sequences per subject. In total there are 52 fine-grained action classes. Specifically, this dataset offers a setting to train models to recognise manipulative gestures such as hand-object interactions which are important in food preparation, manufacturing, and assembly tasks. 

\vspace{-6pt}
\subsection{MPII cooking activities dataset \cite{rohrbach2012database}}
The dataset consists of 12 subjects performing 65 fine-grained cooking related activities. The dataset is captured from a single camera view at a 1624 $\times$ 1224 pixel resolution with a frame rate of 30 fps. The dataset consists of 44 videos and has a total length of more than 8 hours. The duration of each video ranges from 3 to 41 minutes. To maintain a realistic recording setting, the dataset authors did not provide instructions for individual activities, but informed the participant in advance what dish that they are required to prepare (\textit{e.g}. salad), the ingredients to use (cucumber, tomatoes, cheese), and the utensils (i.e. grater) that they can use. Therefore, this dataset provides a real world action segmentation setting, with subject specific variations across the same activity due to personal preferences (\textit{e.g.} washing a vegetable before or after peeling it). 

\vspace{-6pt}
\subsection{MPII cooking 2 dataset \cite{rohrbach15ijcv}}
This dataset contains 273 videos with a total length of more than 27 hours, captured from 30 subjects preparing a certain dish, and comprising 67 fine-grained actions. Similar to the MPII cooking activities dataset, this dataset is captured from a  single camera at a 1624 $\times$ 1224 pixel resolution, with a frame rate of 30 fps. The camera is mounted on the ceiling and captures the front view of a person working at the counter. The duration of the videos range from 1 to 41 minutes. As per the MPII cooking activities dataset, this dataset offers different subject specific patterns and behaviours for the same dish, as no instructions regarding how to prepare a certain dish are provided to the participants. In addition to video data, the dataset authors provide human pose annotations, their trajectories, and text-based video script descriptions. 

\vspace{-6pt}
\subsection{EPIC-KITCHENS-100 \cite{Damen2021RESCALING}}
EPIC-KITCHENS-100 is a recent and popular dataset within the action segmentation research community. Data is captured from 45 different kitchens using a head mounted GoPro Hero 7. This egocentric perspective of the actions provides a challenging evaluation setting, as the hand-object interactions and critical informative regions are sometimes occluded by the hand, or they occur out of the camera’s field of view.

In total there are 89,977 action segments of fine-grained actions extracted from 700 videos, with a total duration exceeding 100 hours. The dataset has a resolution of 1920 $\times$ 1080 pixels with a frame rate of 50 fps. This dataset follows a verb-noun action labelling format where an action class such as ``put down gloves'' is broken into the verb ``put down'', and the noun ``gloves''. In total this dataset consists of 97 verbs and 300 noun classes. In addition to video data, the dataset authors provide hand-object detections which are extracted from two state-of-the-art object detectors, Mask R-CNN~\cite{he2017mask} and the model of Shan et al.~\cite{shan2020understanding}, and provide this as an additional feature modality. 

\vspace{-6pt}
\subsection{GTEA \cite{fathi2011learning}}
The Georgia Tech Egocentric Activities (GTEA) dataset offers an egocentric action segmentation evaluation setting, where data is captured from a head mounted GoPro camera with a 1280×720 pixel resolution capturing at 15 fps. In total the dataset contains 28 videos where 7 different food preparation activities (such as Hotdog Sandwich, Instant Coffee, Peanut Butter Sandwich) are performed by 4 subjects. This dataset offers 11 different fine-grained action  classes and on average there are 20 action instances per video. In addition to action annotations, this dataset offers hand and object segmentation masks as well as object annotations, which can be used as additional information cues when training the action segmentation models. 

\begin{table}[!t]
\centering
\caption{Evaluation times in seconds for different buffer sizes. Note that runtime is calculated only for the evaluation of the action segmentation model. Feature extraction time is not taken into consideration.}
\begin{tabular}{|c|c|c|}
\hline
Model                    & Buffer Size & Runtime (s) \\ \hline
\multirow{5}{*}{MSTCN}   & 100         & 0.1129      \\ \cline{2-3}
                         & 250         & 0.3502      \\ \cline{2-3}
                         & 500         & 0.7291      \\ \cline{2-3}
                         & 750         & 1.0435      \\ \cline{2-3}
                         & 1000        & 1.4105      \\ \hline
\multirow{5}{*}{MSTCN++} & 100         & 0.1698      \\ \cline{2-3}
                         & 250         & 0.3416      \\ \cline{2-3}
                         & 500         & 0.7196      \\ \cline{2-3}
                         & 750         & 1.1406      \\ \cline{2-3}
                         & 1000        & 1.4706      \\ \hline
\multirow{5}{*}{SSTDA}   & 100         & 0.3877      \\ \cline{2-3}
                         & 250         & 0.7610      \\ \cline{2-3}
                         & 500         & 1.6754      \\ \cline{2-3}
                         & 750         & 2.1942      \\ \cline{2-3}
                         & 1000        & 3.2258      \\ \hline
\multirow{5}{*}{DGTRM}   & 100         & 1.6014      \\ \cline{2-3}
                         & 250         & 3.9543      \\ \cline{2-3}
                         & 500         & 8.8306      \\ \cline{2-3}
                         & 750         & 17.7307     \\ \cline{2-3}
                         & 1000        & 24.8198     \\ \hline
\end{tabular}
\label{tab:runtimes2}
\vspace{-8pt}
\end{table}

\vspace{-6pt}
\subsection{ActivityNet \cite{Heilbron_2015_CVPR}}
ActivityNet consists of videos that are obtained from online video sharing sites, and comprises 648 hours of video recordings. In contrast to previously described datasets where there are multiple actions per video, ActivityNet contains only one or two action instances per video. This dataset has 203 different action classes and an average of 193 video samples per class. The majority of videos have a duration between 5 and 10 minutes, and are captured at 1280 $\times$ 720 pixel resolution at 30 fps. This dataset offers a balanced distribution of actions which include personal care, sports and exercises, socialising and leisure, and household activities.

\vspace{-6pt}
\subsection{THUMOS15 \cite{idrees2017thumos}}
This dataset contains footage drawn from public videos on YouTube. The dataset contains more than 430 hours of video, and manual filtering was conducted to ensure that the videos contain visible actions. Similar to ActivityNet, THUMOS15 also contains one or two action instances per video. Specifically, this dataset has 101 action classes and the authors provide semantic attributes for the videos such as visible body parts, body motion, visible object and the location of the activity, in addition to action class labels. The majority of the videos in this dataset have a short duration, where the average length of an action is 4.6 seconds.  

\vspace{-6pt}
\subsection{Toyota Smart-home Untrimmed dataset \cite{Das_2019_ICCV}}
The dataset consists of videos recorded at 640 x 480 pixel resolution at 20 fps frame rate using Microsoft Kinect sensors. Each action is captured from at least 2 distinct camera angles, and the dataset offers 3 modalities: RGB, depth and the 3D pose of the subject. In total this dataset contains 536 videos with an average duration of 21 minutes. There are 51 unique fine-grained action classes in this dataset, with high intra-class variations within each video. Furthermore, this dataset captures a real-world daily living setting with various spontaneous behaviours, partially captured or occluded frames, as well as elementary activities that do not follow a specific temporal ordering. Therefore, this dataset offers a challenging real-world evaluation setting in which to test action segmentation models. 

\vspace{-6pt}
\subsection{FineGym \cite{shao2020finegym}}
This dataset is composed of videos which capture professional gymnastic performances, across 303 videos for a total duration of 708 hours. These videos were sub-sampled into 32,697 samples which range in length from 8 to 55 seconds. The majority of the videos are captured at either 720P or 1080P resolution. Compared to other publicly available datasets, this dataset offers a rich annotation structure where the action labels are categorised using a three-level semantic hierarchy. For instance, the action ``balance beam'' contains fine-grained actions such as ``leap-jumphop'', ``beam-turns'', and these sub-actions are further divided into finely defined class labels such as ``Handspring forward with leg change'' and ``Handspring forward''.  This 3 stage annotation structure offers 99, 288 and 530 unique fine-grained action classes. Furthermore, this dataset is highly diverse in terms of viewpoints and poses due to the ``in the wild'' nature of the videos used.

\section{Adapting to Real-World Applications : Runtimes}

In Table~\ref{tab:runtimes2} we provide evaluation times in seconds for models with different buffer sizes, which were introduced in Sec IV-E of the main paper.